\documentclass[m,master]{./template/WeSTthesis}
\usepackage[backend=bibtex8,
            natbib=true,
            style=numeric-comp,
					  maxcitenames=2,
						maxbibnames=99,
            defernumbers=true]{biblatex} 
\renewbibmacro*{doi+eprint+url}{%
  \iftoggle{bbx:doi}
    {\iffieldundef{doi}{%
			\printfield{urll}}{}}

 \iftoggle{bbx:url}
	  {\iffieldundef{doi}
			{\usebibmacro{url+urldate}}{}}
			{}%
	\newunit\newblock%
	\iftoggle{bbx:eprint}
		{\usebibmacro{eprint}}
		{}%
	\newunit\newblock%
	\iftoggle{bbx:doi}
		{\printfield{doi}}
		{}
}
\usepackage[section,subsection]{placeinsoverride}
\usepackage{amsfonts}
\usepackage{amssymb}
\usepackage{mathtools}
\usepackage{textcomp}

\usepackage[labelfont=bf]{caption}
\RequirePackage[utf8]{inputenc}
\usepackage{microtype}
\usepackage{floatrow}
\DeclareFloatFont{textsize}{\footnotesize}
\usepackage{tabularx}
\newcolumntype{Y}{>{\centering\arraybackslash}X}

\usepackage[toc,nonumberlist]{glossaries}
\makeglossary%

\newcommand{\I}{\enquote{I}}
\newcommand{\im}{\I{}-message}
\newcommand{\ims}{{\im}s}
\newcommand{\Y}{\enquote{you}}
\newcommand{\ym}{\enquote{You}-message}
\newcommand{\yms}{{\ym}s}
\newcommand{\iyms}{\I{}- and {\yms}}
\newcommand{\afd}{AfD}
\newcommand{\afds}{\afd{} discussions}
\newcommand{\RQ}[1]{\textsf{RQ#1}}
\newcommand{\citeW}[1]{\citep{#1}}

\usepackage{csquotes} 
\usepackage{booktabs} 

\usepackage{hyperref}
\usepackage{cleveref} 

\hypersetup{%
  pdftitle    = {\enquote{Did I Say Something Wrong?}\\A Word-Level Analysis of Wikipedia Articles for Deletion Discussions},
  pdfauthor   = {Michael Ruster},
  pdfkeywords = {wikipedia,I messages,you messages,Gordon,function words,Pennebaker,classification,discussion,arousal},
  colorlinks  = true,
  unicode     = true,
  linkcolor   = black,
  citecolor   = black,
  filecolor   = black,
  urlcolor    = black,
}
\title{\enquote{Did I Say Something Wrong?}\\A Word-Level Analysis of Wikipedia Articles for Deletion Discussions}
\author{Michael Ruster}
\degreecourse{Web Science}
\firstreviewer{Prof.\ Dr.\ Steffen Staab}
\firstreviewerinfo{Institute for Web Science and Technologies}
\secondreviewer{Ren\'{e} Pickhardt}
\secondreviewerinfo{Institute for Web Science and Technologies}
\newglossaryentry{article} {%
  name={article},
  plural={articles},
  description={is a Wikipedia page containing encyclopedic contents.
              Discussion pages like talk pages or \afd{} are not considered articles.
              Neither are user pages or Wikipedia policy pages or behavioural guidelines}
}
\newacronym{afd}{AfD}{Articles for Deletion}
\newglossaryentry{discussion} {%
  name={\afd{} discussion},
  plural={\afd{} discussions},
  description={is a discussion on Wikipedia focused around a Wikipedia article.
    The goal is to determine in a collaborative effort whether the article should be deleted.
    Contributing to the discussion does not require a Wikipedia account}
}
\newglossaryentry{editor} {%
  name={editor},
  plural={editors},
  description={is a user who actively contributes to Wikipedia, e.g.\ by writing articles or participating in discussions}
}
\newglossaryentry{contribution} {%
  name={contribution},
  plural={contributions},
  description={is any active interaction with Wikipedia.
    Visiting and reading articles is not seen as a contribution but e.g.\ writing articles or participating in discussions are}
}
\newglossaryentry{contributor} {%
  name={contributior},
  plural={contributors},
  description={is a Wikipedia user that has at least once made changes to a Wikipedia page}
}
\newglossaryentry{disruptive} {%
  name={disruptive},
  description={describes contributions made, which negatively affect the collaborative process.
    Such contributions were either made in good faith or by editors who usually act in good faith and have no intention of damaging Wikipedia.
    On Wikipedia, posts that attack others or that are detrimental to the process are considered disruptive.
    We assume that disruptive posts will lead to their author being blocked}
}
\newglossaryentry{constructive} {%
  name={constructive},
  description={describes any contribution made, that positively affects the collaborative process.
    On Wikipedia, posts that discuss the topic at hand objectively are considered constructive}
}
\newglossaryentry{post} {%
  name={post},
  plural={posts},
  description={is a special form of contribution in which content is added to a discussion}
}
\newglossaryentry{block log} {%
  name={block log},
  description={is a document which records whenever an administrator blocks a user.
  It contains information about who blocked whom when and why}
}

\begin{document}

\pagenumbering{roman}
\maketitle
\cleardoublepage%
\paragraph*{Abstract}
This thesis focuses on gaining linguistic insights into textual discussions on a word level.
It was of special interest to distinguish messages that constructively contribute to a discussion from those that are detrimental to them.
Thereby, we wanted to determine whether \iyms{} are indicators for either of the two discussion styles.
These messages are nowadays often used in guidelines for successful communication.
Although their effects have been successfully evaluated multiple times, a large-scale analysis has never been conducted.
Thus, we used Wikipedia Articles for Deletion (short: AfD) discussions together with the records of blocked users and developed a fully automated creation of an annotated data set.
In this data set, messages were labelled either constructive or disruptive.
We applied binary classifiers to the data to determine characteristic words for both discussion styles.
Thereby, we also investigated whether function words like pronouns and conjunctions play an important role in distinguishing the two.
We found that \yms{} were a strong indicator for disruptive messages which matches their attributed effects on communication.
However, we found \ims{} to be indicative for disruptive messages as well which is contrary to their attributed effects.
The importance of function words could neither be confirmed nor refuted.
Other characteristic words for either communication style were not found.
Yet, the results suggest that a different model might represent disruptive and constructive messages in textual discussions better.

\paragraph*{Zusammenfassung}
Diese Arbeit beschäftigt sich damit, linguistische Erkenntnisse auf Wortebene über schriftlichen Diskussionen zu gewinnen.
Die Unterscheidung zwischen Botschaften, welche sich förderlich auf Diskussionen auswirken und jene, welche diese unterbrechen, spielte dabei eine besondere Rolle.
Hierbei lag ein Schwerpunkt darauf, zu ermitteln, ob Ich- und Du-Botschaften charakteristisch für die beiden Kommunikationsarten sind.
Diese Botschaften sind über Jahre hinweg zu Empfehlungen für erfolgreiche Kommunikation avanciert.
Ihre zugeschriebene Wirkung wurde zwar mehrfach bestätigt, jedoch geschah dies stets in kleineren Studien.
Deshalb wurde in dieser Arbeit mithilfe der Löschdiskussionen der englischen Wikipedia und der Liste gesperrter Nutzer eine vollautomatische Erstellung eines annotierten Datensatzes entwickelt.
Dabei wurden Diskussionsbotschaften entweder als förderlich oder schädlich für einen konstruktiven Diskussionsverlauf markiert.
Dieser Datensatz wurde anschließend im Rahmen einer binären Klassifikation verwendet, um charakteristische Worte für die beiden Kommunikationsarten zu bestimmen.
Es wurde zudem untersucht, ob anhand von Synsemantika (auch bekannt als Funktionswörter) wie Pronomen oder Konjunktionen eine Entscheidung über die Kommunikationsart einer Botschaft getroffen werden kann. 
Du-Botschaften wurden, übereinstimmend mit ihrer zugeschriebenen negativen Auswirkung auf Kommunikation, als schädlich in den durchgeführten Untersuchungen identifiziert.
Entgegen der zugeschriebenen positiven Auswirkung von Ich-Botschaften, wurde bei diesen ebenfalls eine schädlich Wirkung festgestellt.
Eine klare Aussage über die Relevanz von Synsemantika konnte anhand der Ergebnisse nicht getroffen werden.
Weitere charakteristische Worte konnten nicht festgestellt werden.
Die Ergebnisse deuten darauf hin, dass ein anderes Modell textliche Diskussionen potentiell besser abbilden könnte.

\cleardoublepage%
\tableofcontents
\cleardoublepage%
\chapter*{Acknowledgement}
I want to thank Ren\'{e} Pickhardt for introducing me to \iyms{}, and proposing the idea of analysing their effects on communication using Wikipedia deletion discussions.
Moreover, I am thankful for his regular feedback and support, as well as making me aware of the 2nd GESIS Winter Symposium 2015 where I was able to present this topic.
Further, I want to thank Prof.\ Dr.\ Steffen Staab for valuable input that made me consider function words and led to developing the sliding window approach.
Finally, I thank Aaron Kohn and Martin K\"{o}rner for proofreading this thesis.

\newpage
\pagenumbering{arabic}

\chapter{Motivation}
Already in the 1960s, Thomas \citet{gordon_theory_1967} coined the term \emph{\ims{}} while working as a psychologist with parents and their children.
An \im{} is described as an honest and direct message that expresses the sender's feelings.
Its name is derived from them mostly having the first-person singular as subject.
Furthermore, \ims{} are characterised as being non-judgemental and thus e.g.\ suited for conflict resolution \citep{doherty_parent_1980}. 
The concept became famous through Gordon's influential work introducing the \emph{Parent Effectiveness Training} (short: \emph{P.E.T.}) \citep{gordon_parent_2000}.
It describes an approach for parents communicating with their children for teaching them to solve their problems independently.
\citet{gordon_parent_2000} analogously calls messages with the second-person singular as subject \emph{\yms{}}.
He attributes the opposite characteristics of \ims{} to them, stating that they convey strong criticism.
\citet{gordon_leader_2001} later transferred these concepts onto leadership and reported many successful experiences from real-world applications.
The general idea of \iyms{} became widely used as recommendation for successful communication, even outside of leadership and teaching.

\I{} and \Y{} are themselves \emph{function words}, which is a set of common words that contribute little to the content of a sentence from a semantical perspective~\citep{chung_psychological_2007}.
These include among others articles, prepositions, conjunctions, auxiliary verbs and other pronouns.
However, \citet{chung_psychological_2007} showed that function words, especially pronouns, can encode emotions and information that are not explicitly articulated in the text.

The role of function words in discussions has not yet been analysed.
To our knowledge, the presumed effects of \iyms{} have never been evaluated in a big scale on discussions as well.
For these reasons, we wanted to analyse a great number of discussions for determining whether these concepts or certain words indicate \enquote{good} or \enquote{bad} communication.
\citet{mehrabian_silent_1981} found out that while communicating feelings and attitudes, the actual content merely makes $7$\% of the complete impression on the receiver of the message. 
The rest is shared by the tone of voice and gestures.
To ensure that the analysed communication is only affected by the message contents, we considered purely textual discussions.

Due to the amount of available data, we used \emph{Articles for Deletion discussions} (short: \emph{\glspl{discussion}}) for constructing a model.
These are pages featuring textual discussions between Wikipedia users that determine whether an \gls{article} should be deleted for reasons such as irrelevance or wrong information.
Hereafter, messages in these discussions will be called \emph{\glspl{post}}.
\enquote{Good} communication will be called \emph{constructive} in this context.
It can be described as positively affecting the collaborative process, e.g.\ by enriching the current discussion.
Likewise, \enquote{bad} communication will be called \emph{disruptive}.
It is characterised by being highly detrimental to the collaboration process like posts that harass or attack other users.
In the context of \glspl{discussion}, disruptive communication does not describe posts created with the intention to harm Wikipedia.
Instead, disruptive messages should form an exception to authors otherwise trying to improve Wikipedia.
To determine constructive and disruptive posts, we referred to a Wikipedia document containing records about users being (temporarily) blocked from editing due to misbehaviour.
Thus, the task at hand was reduced to a binary classification problem and a gold standard based evaluation was conducted.
Considering single posts, the dependent variable was whether such a post would lead to its author being blocked or not.
As this yielded unsatisfying results, we additionally conducted tests in which the contents from a subset of the user's post history were considered.

We expected the results to broaden our understanding of which words are heavy indicators for messages in textual discussions being either constructive or disruptive.
Of special interest were \iyms{} and function words.
Assuming that textual discussions in general and \glspl{discussion} are comparable, the following research questions were derived:
\begin{itemize}
  \item[\textbf{\RQ{1}:}] Do disruptive messages in textual discussions contain more \yms{} than constructive ones?
And likewise,
  \item[\textbf{\RQ{2}:}] Do constructive messages in textual discussions contain more \ims{} than disruptive ones?
  \item[\textbf{\RQ{3}:}] Is solely considering function words sufficient for determining whether a message is constructive or rather detrimental to a textual discussion?
  \item[\textbf{\RQ{4}:}] Which other words are typical for constructive and which for disruptive messages in textual discussions?
\end{itemize}

Such knowledge could be used to encourage friendly collaboration and fight cyber-bullying:
When users use too many words that are common for disruptive messages, they could be asked whether they really intends to send the message.
This could make them reconsider their wording which could otherwise be offending and detrimental to the ongoing discussion.
In the context of the Google Science Fair 2014, a young researcher conducted a comparable study with encouraging results for adolescents to rethink sending offending messages~\citep{prabhu_google_2014}.
Nevertheless, it is unclear, how effective such an approach would be for the mostly adult users of Wikipedia.
Answers to these research questions could also be used to analyse \glspl{discussion} on Wikipedia and alarm administrators when they should interfere to calm down a heated discussion.

Our results suggest that disruptive messages indeed contain more \yms{} and thus support \RQ{1}.
However, the opposite, as questioned by \RQ{2}, did not hold true for \ims{} in our classifications.
When only considering function words, the classifiers performed close to random guessing.
This is possibly related  to the full text classifiers' mediocre performance.
Therefore, \RQ{3} cannot be answered although one of the tests returned surprisingly good results for function words classification.
Inspecting typical words for disruptive and constructive messages did not return valuable results to answer \RQ{4}.
Instead, these words are mostly specific to Wikipedia and \glspl{discussion}.
The results thus suggest that another model might be more suitable for analysing the language used in textual discussions than \glspl{discussion} and the records of blocked users.

\Cref{s:psych} introduces the concepts of \iyms{} and function words in more detail, explaining their effects on and their role in communication.
Related work, in which conflicts and disputes in textual discussions were analysed, is presented in \Cref{s:relatedWork}.
Moreover, approaches to detect vandalism on Wikipedia are shown.
\Cref{s:preprocessing} presents the data we chose to build our model on, and the preprocessing steps that were needed to do so.
Our fully automatic approach of building an annotated data set is explained in \Cref{s:timeframes}.
In this \namecref{s:timeframes}, we also introduce our classifiers and the metrics used in the later evaluation.
Furthermore, characteristics of the newly created data set are studied.
\Cref{s:testsetup} describes the test setup with its goals and the chosen model validation.
The results of our tests are given in \Cref{s:results}.
\Cref{s:errors} illustrates assumptions and design decisions that we made and how they could have impacted the results.
Finally, \Cref{s:conclusion} concludes this thesis and the research questions are answered.

\chapter{Psychological Concepts}\label{s:psych}
This \namecref{s:psych} focuses on the psychological backgrounds to this thesis.
The sections further explain the concepts of \iyms{} as well as function words.
In both sections, studies are summarised which motivated us to inspect their impact within our model of textual discussions.
Moreover, first details are given about how the \glspl{discussion} will be tested for \iyms{} and function words.
Furthermore, it will be explained how their application to our model affects the research questions established in the first section.

\section{\I{}- and \enquote{You}-Messages}
Gordon's conflict prevention and resolution recommendations do not solely focus on \iyms{}.
Instead, he highlights that his concept is based on two more skills besides these messages.
The first skill is conflict resolution based on mutual agreement of both parties such that none of the parties feel to have \enquote{lost} the conflict~\citep{gordon_theory_1967,gordon_leader_2001}. 
\emph{Active listening} is described as a third important skill~\citep{gordon_theory_1967}.
It advises one party to repeat the understood emotions and content of the other party's messages to ensure that both parties understand each other.
Gordon recommends to refrain from using \yms{} and instead use \ims{}.
The latter consist of three parts~\citep{gordon_leader_2001}:
\begin{enumerate}
  \item A description of the intolerable behaviour of another party,
  \item one's feelings towards this behaviour, and
  \item the effect this behaviour has or will have on one's life.
\end{enumerate}
The message should not be blameful.
It should allow the listener to understand the speaker's feelings and the potential impact that the listener's actions may have on the speaker in general as well as on the speaker's well-being in particular.

A meta-evaluation by \citet{muller_zur_2001} confirmed many claimed positive effects of Gordon's training, especially for older children up to the age of twelve.
\citet{gordon_leader_2001} also highlighted successful experiences for applications in leadership.
It indicates that Gordon's concept can also positively influence the communication and problem-solving between adults.
Over time, the communication concept often got reduced to \iyms{} and claimed to be an effective tool for general communication as well~\citep{ii_exploratory_1993}.
The messages were even further simplified so that the term \ims{} was then no longer used for messages expressing behaviour, feelings and effects.
Instead, they are nowadays mostly identified by containing the pronoun \I{} while describing one's feelings towards someone else's behaviour.
Yet still, studies e.g.\ by \citet{kubany_verbalized_1992} showed a positive effect of the use of \I{}- over \yms{}.
Further studies showed similar traits simply by inspecting the use of the pronouns \I{} and \Y{}.
\citet{weintraub_verbal_1981} noted that angrier people tended to use \Y{} more frequently.
A study on marital interactions saw \Y{} correlated with negative and \I{} with positive effects on problem discussions~\citep{simmons_pronouns_2005}. 
Another study by \citet{slatcher_am_2008} showed a positive effect of the more frequent use of \I{} over \Y{} in couples' conversations on their relationship stability.
However, the found correlations were less apparent.

All in all, many studies support the positive and negative effects of \iyms{} and their simplified adaption.
However, these studies are restricted to small group sizes or little text corpora.
Hence, we want to determine whether we find evidence supporting Gordon's concept in \glspl{discussion}.
Creating an elaborate model to capture \iyms{} would be out of this master thesis' scope due to its potential complexity.
Instead, we reduce this task to counting of occurrences of the pronouns \I{} and \Y{} similar to other research~\citep[e.g.][]{chung_psychological_2007,simmons_pronouns_2005,slatcher_am_2008}.
The first two research questions can then be adapted to our model as follows:
\RQ{1} poses the question whether disruptive posts contain the word \Y{} more frequently than constructive ones.
Vice versa, \RQ{2} asks whether constructive posts contain the word \I{} more frequently than disruptive ones.

\section{Function Words}
Function words span pronouns, articles, prepositions, conjunctions, and auxiliary words~\citep{chung_psychological_2007}.
A common approach for analysing function words, as well as extracting implicitly encoded information from text in general, is counting words, which are believed to be connected e.g.\ with certain emotions.
\citet{pennebaker_linguistic_2001} developed the frequently used program \emph{Linguistic Inquiry and Word Count} (short: \emph{LIWC}).
It analyses text by counting words that it has internally mapped against emotions, content and function word categories.
For example, words like \enquote{love} and \enquote{nice} can indicate positive emotions~\citep{pennebaker_development_2015}.
The leisure content category is associated with words such as \enquote{cook} and \enquote{movie}~\citep{pennebaker_development_2015}.

By conducting multiple studies using LIWC and investigating studies of other researchers, \citet{chung_psychological_2007} found function words to encode different information.
For example, \citet{newman_lying_2003} found that lying people were using function words measurably differently than truth-speaking people.
Among other characteristics, liars were likelier to use fewer first-person singular pronouns and fewer exclusive words such as \enquote{but} or \enquote{except}.
A more frequent use of the pronoun \I{} was also associated with higher blood pressure~\citep{chung_psychological_2007} as well as depression~\citep{chung_psychological_2007,de_choudhury_predicting_2013}.
Summarising three different studies, \citet{chung_psychological_2007} found the use of pronouns to be an even better indicator for depression than the appearance of negative emotion words. 
Analogously, \citet{chung_psychological_2007} attributed the frequent use of third person pronouns such as \enquote{she} and \enquote{they} positive effects on a person's well-being.
By inspecting dialogues, the authors further found that people of lower status use the word \enquote{I} more frequently.
Thus the relationship between two communicating parties may be subtly encoded.
Most studies of \citet{chung_psychological_2007} analysed English text of US-American participants.
However, when comparing results with English text that was translated from Japanese they found differences in the pronoun use and concluded that there are cultural differences in their use.

In sum, a person's use of function words and especially pronouns may unknowingly encode interesting information.
In particular, \I{} and \Y{} have been shown to correlate with the absence or existence of disputes.
Due to the presumed effects of \I{} and \Y{} and the broad range of information function words seem to encode, we hope to gain further insights and potentially detect new correlations.
To our knowledge, no efforts have yet been made to determine whether connections between function words as a whole and constructive or disruptive communication in textual discussions exist.
Therefore, we decided to not only do full text classification but classification exclusively respecting function words as well.
The later classification may then give answers to \RQ{3}, which asked if function words alone contain sufficient information to deduce whether a message is constructive or disruptive.
As LIWC is a commercial software, we did not apply it to the \glspl{discussion}.
Nevertheless, results of studies conducted with LIWC highlight the importance of function words.


\chapter{Related Work}\label{s:relatedWork}

In this \namecref{s:relatedWork}, research is presented that has already been carried out on conflicts on Wikipedia, Wikipedia discussions and textual discussions in general.
Furthermore, information on research on \emph{vandalism} detection is given, which is related to our work.

Vandalism is defined by a Wikipedia policy~\citeW{Wikipedia:Vandalism} as the act of wilfully editing existing or creating new pages with the goal to harm Wikipedia.
The policy highlights that, independent of the quality, any edits made with the intention to improve Wikipedia are not to be understood as vandalism.
Therefore, articles that are obviously vandalism are not part of \glspl{discussion}~\citeW{Wikipedia:Articles_for_deletion}.
Instead, they may be immediately deleted~\citeW{Wikipedia:Criteria_for_speedy_deletion}.
This thesis aims to determine the impact of words in serious discussions that have the goal of reaching consensus.
Hence, there is no interest in vandalism to us but only in posts made in good faith.
We suspect the \glspl{discussion} to attract few vandals because vandalising these pages causes relatively little harm to Wikipedia considering that not many people view them.
Although limited, our experiences while working with these discussions support this assumption as we rarely saw any instances of vandalism.
So, despite vandalism also being detrimental to rational and objective discussions, it differs from our goals to analyse good-faith discussions.

If not specified otherwise, the word \emph{user} hereafter describes anyone who visits Wikipedia either for gathering information or for any form of contribution.
A \emph{contribution} is any active interaction made with Wikipedia such as writing articles, participating in discussions, reverting changes and the like.
The word \emph{\gls{editor}} will be used to refer to the subset of users that contribute to Wikipedia, e.g.\ by writing articles or participating in discussions.

\section{Conflicts and Disputes}
This section presents research done on textual conflicts and discussions.
The researchers' intentions varied from analyses to gain insights about the communication culture of users to the detection of disputes.

Our interest is to textually analyse posts to determine which ones no longer objectively address the relevance of an article and instead interrupt the discussion e.g.\ by containing personal attacks against other users.
\citet{yasseri_dynamics_2012} on the other hand set out to detect and analyse conflicts during the collaborative editing process on the article itself.
They have processed Wikipedia to detect collaboration conflicts called \emph{editorial wars} (short: \emph{Edit wars}).
Edit wars are characterised by groups sharing different opinions trying to enforce their opinion in an article.
Discussions related to the collaboration of an article happen on their respective talk pages.
A talk page is a separate page dedicated solely to the discussion of its associated article~\citeW{Wikipedia:TalkPages}.
The conflicts analysed by \citet{yasseri_dynamics_2012} take place on articles as well as their talk pages.
The authors moreover set out to separate edit wars from pure vandalism and focus on serious discussions in good faith like we do as well.
Edit wars are often accompanied by vandalism and reversions of the article to older revisions.

Deletion discussions end within a few weeks after a consensus was formed~\citeW{Wikipedia:Deletion_process}.
Edit wars however often span longer or even an indefinite time.
\citet{yasseri_dynamics_2012} distinguish three types of such conflicts.
The first is when the war ends and a consensus is reached.
A second type is identified by the authors as reoccurring phases of temporary consensus where a consensus seems to be reached multiple times but will always be broken up by another edit war phase.
The last type is never-ending wars, which the authors identified to be typical for highly controversial topics such as the Liancourt Rocks~\citeW{Liancourt_Rocks} that both Japan and Korea claim as their land.
\citet{yasseri_dynamics_2012} did not find a correlation between the edit frequency of articles and conflicts. 

To a reasonable extent, the length of the article's talk pages was an indicator for conflicts.
However, the authors found it to be true for the English but not e.g.\ the Hungarian Wikipedia.
Therefore, their approach to reliably detect conflicts in articles instead considers the number of editors $E$ contributing to an article and a measure based on reversions~\citep{sumi_characterization_2011}.
\autoref{eq:mutualReverts} shows the formula to calculate their controversy measurement $M$ with the assumption that a high number of editors active in a discussion is an indicator for an edit war.
$N^d_i$ is the number of edits of the disputed article by the editor $d$ of revision $i$.
Revision $i$ was reverted by the editor $r$ of revision $j$ for whom likewise $N^r_j$ is the number of their edits.
Although not captured by this formula, \citet{sumi_characterization_2011} restrict the pairs $(N^d_i,N^r_j)$ to those of \emph{mutual reversions}.
That is, the editor of revision $i$ and $j$ must both have reverted at least one revision of the other editor at some point.
Thus, single direction reversions which are often a result of restoring a vandalised article to a prior state, are not weighted.
Mutual reversions with at least one participant having vandalised the article contribute little weight due to considering only the minimum of each revision pair.
The reason for this is that vandalism likely leads to an editor being blocked and thus they may no longer participate in a discussion for some timeframe~\citeW{Blocking_policy}.
Conversely, conflicts between long-term editors are heavily weighted as the authors assume that such create more controversy.
\begin{equation}\label{eq:mutualReverts}
  M = E \cdot \sum_{(N^d_i,N^r_j)<\max}\min(N^d_i,N^r_j)
\end{equation}
Ignoring the highest pair of mutual reversions prevents two heavily---perhaps even personally---fighting editors from skewing the result.
\citet{yasseri_most_2013} were able to successfully apply this measure to ten different language editions of Wikipedia---including such diverse languages as English, Arabic and Czech.
They found the three most controversial categories to be politics, countries such as geographical locations or cities and religion. 

While \citet{yasseri_dynamics_2012} detected and analysed the characteristics of conflicts in articles, \citet{laniado_when_2011} set out to investigate editor interaction on article talk pages.
For every talk page, they built a tree with the article page as root and discussion comments as children.
Replies to comments were treated as children to their original comment.
Inspecting the height of the trees as well as the amount of nodes, \citet{laniado_when_2011} determined extensively discussed categories.
These categories are comparable to those found by \citet{yasseri_most_2013} although \citet{laniado_when_2011} only analysed the English edition of Wikipedia.
Yet, as they chose different categories, \citet{laniado_when_2011} additionally found philosophical and law related articles to attract lengthy discussions on their talk pages.
In regard to the trees of such pages, this means that they contain many leaves with high depth.
Furthermore, they found that editors, who reply to many others, mostly replied to inexperienced editors.
Users who received many replies from others, frequently engaged in discussions with others.
This analysis is restricted to a social networking level of interacting editors.
In contrast, the goal of this thesis is the investigation on a content level instead of learning about the general Wikipedia communication culture.

\citet{hassan_whats_2010} developed a classifier to determine the attitude of Usenet users in threaded online discussions towards each other.
They distinguish between positive and negative attitude.
Among others, the first includes agreement and praise, whereas the negative attitude includes disagreement or insults.
Their task therefore differs from ours in that we distinguish in disruptive posts and constructive ones.
That is, insults often indicate disruptive posts.
Yet, we regard disagreement as legitimate part of discussions and hence classify such posts as constructive as long as they do not contain personal attacks or similar.
\citet{hassan_whats_2010} analyse data replies to other users extracted from various Usenet discussion groups.
They constructed a graph from the words of each sentence in which they link words semantically related to each other or related by statistical co-occurrence.
It is used to build Markov models, which are stochastic models.
Those are used in a support vector machine (short: SVM) which achieves an F1 score of $80.2$\% and an accuracy of $80.3$\%.
Both are evaluation metrics which are introduced in more detail in \Cref{s:metrics}.
The performance is notably better than our results.

\citet{wang_piece_2014} built a binary classifier for online dispute detection on Wikipedia talk pages.
They performed a sentiment analysis on a sentence level and determined disputes according to the relation of positive to negative sentiments.
Different to our approach, \citet{wang_piece_2014} predicted whether there is a dispute or not for whole discussions instead of single posts.
The authors compiled a text corpus for testing by talk pages that are tagged with labels indicating disputes such as \enquote{\texttt{DISPUTED}}.
For non-disputed data, they referred to the absence of such tags on talk pages.
Training took place on the \emph{Authority and Alignment in Wikipedia Discussions} (short: \emph{AAWD})~\citep{bender_annotating_2011} corpus.
The corpus consists of $365$ discussions from talk pages that have been manually annotated by two or more annotators each.
Relevant for the classifier developed by \citet{wang_piece_2014} are the \emph{alignment moves} annotated by \citet{bender_annotating_2011}.
These are labels on a sentence level that categorise whether they express agreement or disagreement towards one or more other participants of the discussion.
Besides the sentiment analysis, \citet{bender_annotating_2011} also considered information about the discussion in total.
Among others, their classifier additionally respected the length of replies, the number of editors active in the discussion and the topical category this discussion takes place in.
\citet{bender_annotating_2011} note that arguments lead to longer answers.
Similar to \citet{sumi_characterization_2011}, they argue that the participation of more editors in a discussion increases the likelihood of a dispute appearing.
The authors furthermore made their classifier respect the category of the article being discussed as they argue that topics like politics or religion are likely to attract disputes.
They used an SVM as classifier that respected all these features and returned an F1 score of $78.25$\% and an accuracy of $80.00$\%.

\section{Vandalism Detection}
Software tools that aid editors and especially administrators in detecting and removing vandalism play an important role in countering deliberate attempts of damaging Wikipedia~\citep{geiger_work_2010}.
Thus, developing algorithms that can accurately detect vandalism is an active field of research with manifold approaches of which a few are subsequently presented.
Vandalism in posts is similar to disruptive posts in that both are detrimental to discussions.
Yet, we assume good faith in what we call disruptive posts, whereas vandalism aims at wilfully harming Wikipedia.
The presented vandalism detection solutions textually analyse individual posts' contents like in our approach.
However, most also consider additional features such as metadata or structural information to improve their predictions.

\citet{chin_detecting_2010} employ a bigram language model on article text to detect vandalism independent of any contributors.
Their language model returns various measurements and statistics such as a post's number of words or perplexity, a commonly used measurement for evaluating language models.
These are then classified by an SVM, logistic regression or decision trees.
Instead of using a big data set, the authors tested this approach only on two of the most vandalised articles.
They found decision trees to perform best.
Moreover, \citet{chin_detecting_2010} found little overlap in the vandalism detected by the SVM and logistic regression.
Thus, they conclude that combined methods could further improve their vandalism detection approach.

\citet{harpalani_language_2011} use trigram language models together with other features including the use of words from colloquial and vulgar language as well as objectiveness measurements.
However, the authors saw the biggest prediction improvements when considering features based on a \emph{probabilistic context free grammar} (short: \emph{PCFG}).
PCFGs are grammars whose rules are assigned a probability derived from the frequency of the rules being used in the training data~\citep{charniak_tree-bank_1996}.
Similar to a language model, \citet{harpalani_language_2011} used a PCFG to detect a common writing style of vandals.

\citet{adler_wikipedia_2011} on the other hand built a classifier respecting multiple features including a trust/reputation model of users.
It calculates reputations per user and per country.
The later is determined by the contributor's IP-address.
Contribution metadata is another feature that the classifier by \citet{adler_wikipedia_2011} respects.
Among others, it considers the comment length and the time since the last edit.
Extraordinarily short or long comments as well as frequently edited articles are an indication for vandalism~\citep{west_detecting_2010,adler_wikipedia_2011}.
Another feature is the analysis of the contributed text for a high amount of uppercase letters, which vandals use for their posts to gain attention~\citep{potthast_automatic_2008,mola-velasco_wikipedia_2012}.
The authors also considered language features, namely the use of pronouns and words indicating biased contributions such as superlatives or bad style, e.g.\ colloquial language.
Their motivation however is not based on the psychological effects of \iyms{} but on their indication for non-objective contributions.
This is rational considering that their focus was to detect vandalism on article pages where objective information should be compiled.
As article pages provide contents that are of interest for the majority of people visiting Wikipedia, these are the main goal for vandals to gain attention and create great harm to Wikipedia.
Therefore, current vandalism detection algorithms often focus on vandalism of article pages, whereas this thesis concentrates on discussions in which good-faith editors display misbehaviour.
Nonetheless, the related topic of vandalism detection shows that there are many different features of Wikipedia contributions which successful classifiers may consider.

In conclusion, Wikipedia is actively used for discussion and interaction with other editors.
The communication and interaction can remain peacefully constructive even over longer times.
But there are also disputes of varying lengths of which some halt productive collaboration.
In an effort to ensure good article quality and a minimum amount of article relevance, \glspl{discussion} are intended as a tool for such peacefully constructive discussions.
Nevertheless, some also transform into misbehaviour and are of particular interest to us.
These incidents are per policy not vandalism as they are mostly based upon good-faith contributions that drift into personal attacks or harassment.
Existing research on textual discussions and especially on Wikipedia discussions either focuses on the discussion as a whole or distinguishes between agreement and disagreement.
In this thesis, however, we regard disagreement as a natural part of discussions and focus on linguistic characteristics of messages that are detrimental to collaborative processes due to e.g.\ verbally attacking other editors.
Therefore, we solely consider text written by editors whereas other approaches often additionally or exclusively use structural approaches and other features.
For example, networks were built from posts or edit and reply frequency were evaluated in previous research.
To the best of our knowledge, this thesis marks the first efforts to evaluate textual discussions in a large scale for determining characteristics of language that indicate disruptive posts.
Likewise, evaluations for the effect of \iyms{} and function words in this setting were yet missing.
The following chapters explain our approach from data preparation over test setups to results.

\chapter{Data Extraction and Preprocessing}\label{s:preprocessing}
We processed the Wikipedia data dumps from the second of June 2015\footnote{\url{https://dumps.wikimedia.org/enwiki/20150602/} -- last accessed 13 July 2015, 15:20} which were the most recent at that time.
They include all Wikipedia pages as well as the complete Wikipedia log, which documents actions including account creation, article deletion, blocking of users~\citeW{Help:Log}.
Relevant for this thesis are the documented blocks.
Hence, we filtered the log accordingly.
Hereafter, the filtered log will be referred to as \emph{block log} as it is done on Wikipedia as well~\citeW{Special:Blocklog}.
$326,538$ \gls{discussion} pages and $3,332,551$ registered occurrences of editors being blocked were found in the data dumps.
The following \Cref{ss:preprocessing_afds,ss:preprocessing_block} go into detail about data properties of the \afds{} and blocks respectively.
Moreover, they describe our approaches and design decisions for preprocessing the data.

\section{Articles for Deletion Discussions}\label{ss:preprocessing_afds}
\afd s are Wikipedia pages, which are used to discuss whether an article should be deleted.
Many arguments revolve around whether an article fulfills the relevancy criteria~\citeW{Wikipedia:Notability} of Wikipedia or not.
That is, articles that are not deemed sufficiently relevant by the Wikipedia community will be deleted.
Discussions often become heated due to disagreement between the two user groups called \emph{inclusionists} and \emph{deletionists}~\citeW{Wikipedia:AfD_war_zone}.
Most of the time, an editor cannot be labelled as belonging to only one of the two groups but leaning towards one in certain topics.
The inclusionists argue that most articles are worth keeping, i.e.\ should not be deleted, as long as they are relevant to a few people~\citeW{DeletionismInclusionism}.
This stance is motivated by the idea that contrary to a printed encyclopedia, the cost for one more article is negligible.
Deletionists on the other hand argue that an article must be interesting to many people for fulfilling the relevancy standards of Wikipedia and can thus be kept~\citeW{DeletionismInclusionism}.
As a result, \glspl{discussion} provide a great amount of human discussions featuring disagreement.

The procedure of an \gls{discussion} is as follows~\citeW{Wikipedia:Articles_for_deletion}.
If an article potentially fails to meet Wikipedia's quality standards, it may be nominated for deletion.
However, if an article clearly validates Wikipedia's rules, e.g.\ contains copyright infringement or was unambiguously invented, it can be flagged for immediate deletion~\citeW{Wikipedia:Deletion_policy}.
An \afd{} nomination is normally done together with a short text describing why the article should be deleted according to the nominator.
A dedicated \gls{discussion} page must then be created with the article's title as its title preceded by \enquote{\texttt{Wikipedia:Articles for deletion/}}.
Other editors may use it to discuss the quality of the article and to express their opinion about what should happen with the article in question.
Besides keeping or deleting the article, editors may also recommend actions such as merging it with another article.
Usually, the \gls{discussion} pages are actively used for seven days to reach consensus on what action to take.
Editors can only articulate recommendations, meaning that their statements are not votes and  the discussions are not resolved by a majority decision.
An administrator will consider all user-made recommendations and determine at their own discretion the most rational action that conforms to the Wikipedia policies.
According to the guideline, the deciding administrator should not take part in the discussion and should not be involved in the discussed topic to prevent bias as much as possible.
Nevertheless, it is not guaranteed that these guidelines are complied with by all participants.

\section{Identifying and Isolating Posts}\label{ss:wikiwho}
This \namecref{ss:wikiwho} describes how individual posts by editors are obtained.
The Wikipedia data dumps contain streamable, compressed archives of all pages together with all their revisions.
There are no separate dumps that only contain \afds{}.
The pages and revisions are encoded as XML-files and \afds{} are regular Wikipedia pages whose title starts with \enquote{\texttt{Wikipedia:Articles for deletion/}}.
Hence, we filtered the approximately $98$ Gigabytes of archives according to the pages' title.
We excluded summary pages which embedded old \afds{} for documentation purposes.
This was done to prevent duplicate data.
The result was $29$ Gigabytes of \afds{} stored in uncompressed XML-files.

The data dumps do not include article modification histories.
I.e.\ instead of computing and storing the differences between two revisions, every Wikipedia article revision consists of the complete text in this revision.
Yet, for analysing editor posts on \afds{}, it is crucial to separate them from the others.
There are guidelines and recommendations for editors on how to communicate on discussion pages so that each editor's contribution can be clearly distinguished~\citeW{Help:Using_talk_pages,Wikipedia:Signatures}.
However, both guidelines and recommendations are not being enforced and are not always adhered to.
Furthermore, Wikipedia pages are frequently being reverted to an earlier revision due to vandalism.
Reverting editors inevitably introduce a lot of changes in the text which should not be attributed to them as they did not author the content.
Likewise, if editors move text---e.g.\ a sentence---written by another editor to another place in the revision text, they should not be credited for this text.
Thus, determining what text an editor introduced in some revision is a non-trivial task that must take all prior revisions into account.

\citet{flock_wikiwho_2014} addressed this problem by developing the algorithm \emph{WikiWho} which associates words from a Wikipedia page revision with an editor that it assumes to be its author.
The authors created a gold standard from Wikipedia articles which was then used for an evaluation in which their algorithm achieved a $95$\% precision.
In so doing, WikiWho performed $10$\% better in this evaluation than the prior state-of-the-art algorithm by \citet{de_alfaro_attributing_2013} while also executing faster \citep{flock_wikiwho_2014}.
Hence, we chose WikiWho for attributing authorship of words in revisions when extracting user posts.
We only considered the contributions by registered users.
This is motivated by the fact that anonymous users are only distinguished by their IP address, which is not a unique identifier.
A later presented approach merges posts by the same editor for the analysis and thus requires each post's author to be uniquely identifyable.
Moreover, anonymous Wikipedia editors are likelier to vandalise pages than registered editors~\citep{kane_multimethod_2011}.


\section{Removing Non-Linguistic and Automatically Generated Contents from Posts}
After single posts have been extracted and those by anonymous users have been filtered, their contents have to be processed before they can be used in our later analysis.
The purpose of this processing is to remove non-linguistic content like symbols as well as content that was not created by the post's author.
Subsequently, we describe our removal of markup, templates and signatures.

As we are interested in determining which words negatively affect a constructive, objective collaboration, we want to ignore any forms of markup.
Wikipedia allows the use of a subset of elements from the \emph{HyperText Markup Language} in version $5$ (short: \emph{HTML5})~\citeW{Help:HTML} as well as \emph{Cascading Style Sheets} (short: \emph{CSS}) for formatting page contents~\citeW{Help:CSS}.
HTML element tags and attributes are removed, yet their contents are kept.
CSS rules embedded in HTML attributes are therewith removed as well.

Wikipedia also supports a dedicated markup language called \emph{wikitext}~\citeW{Wiki_markup}. 
Wikitext enables editors to format their text e.g.\ by using lists, making text bold or adding hyperlinks~\citeW{Help:Wiki_markup}.
It is converted to and rendered as HTML when a page containing this markup is viewed in a Web browser~\citeW{Wiki_markup}.
Therefore, after determining an editor's contribution to a discussion, it must be processed to only contain words.
The meaning of markup such as italics or bold is not standardised.
Hence, for reasons of simplicity, we refrain from allowing the formatting to influence the weights of their affected words.
However, it could be considered to e.g.\ attribute higher importance to a word in bold formatting and lesser to a struck through word in future work.

Na\"{\i}vely removing all non-alphabetical symbols is insufficient for example in the context of marked up external hyperlinks.
Hyperlinks are surrounded by square brackets like \enquote{\texttt{[http://ddg.gg]}}.
External links require one square bracket each, whereas Wikipedia internal links require two.
Depending on the implementation, the removal of non-alphabetical symbols would result in one or more words being attributed to the editor as intentionally used in their style of communication.
I.e.\ in this example it could falsely be assumed that an editor wrote the words \enquote{\texttt{http}}, \enquote{\texttt{ddg}} and \enquote{\texttt{gg}}.
Especially URIs with long paths or query strings would add many words
and thus, we ignore URIs marked up as links.
Wikitext furthermore offers editors to provide a text for a hyperlink by using a vertical bar symbol \enquote{\texttt{|}}.
The text is shown instead of its URI when viewing the page e.g.\ \enquote{\texttt{[http://ddg.gg|only show this text]}}.
While we ignore the URI, as it is most likely beyond the control of the editors, any describing text can be freely chosen by them and will be taken into account in our setup.

After manually inspecting many \glspl{discussion}, it became clear that links are often used to refer to internal Wikipedia pages like articles or policy pages.
Linking to policy pages is frequently done to draw another editor's attention to why their behaviour might be inappropriate for the ongoing discussion.
These links are in the form of \enquote{\texttt{[[Wikipedia:No personal attacks]]}} and \enquote{\texttt{[[WP:NPA]]}}.
They can contain a hyperlink text similar to external links.
We process internal Wikipedia links differently than external ones.
The internal links beginning with \enquote{\texttt{Wikipedia:}} or \enquote{\texttt{WP:}} are retained and symbols including spaces removed.
Thus, they will appear as one word to the classifiers, yet can be distinguished from regular communication contents.
For example, \enquote{\texttt{[[WP:NPA]]}} would become \enquote{WPNPA}.
If there is a link text, it will be ignored as long as it is not different to the actual link.

Wikipedia templates~\citeW{Help:Templates} also require special processing.
They are used to easily insert commonly used text into pages such as page headers that describe the current state of an \gls{discussion}.
As their contents have most likely been written by editors other than the editors who embed one or more of these in their posts, they should not be considered in our analysis.
Templates can be included via transclusion~\citeW{Help:Transclusion}.
This is done by using two braces before and after the template name.
For example, the \enquote{Like} template can be included by inserting the text \enquote{\texttt{\string{\string{Like\string}\string}}}.
The template's content will replace the template code \enquote{\texttt{\string{\string{Like\string}\string}}} when the page is rendered but the source code of the page will still show the template code.
These templates are quite easy to detect and are removed during our content processing steps.

However, templates can also be substituted~\citeW{Help:Substitution}.
The referred template's content is then inserted in its most recent revision into the source code of the page.
For example, the \enquote{Like} template is substituted by adding \enquote{\texttt{\string{\string{subst:Like\string}\string}}} to the page.
Most \afd{}-specific templates are to be substituted~\citep[cf.][]{Wikipedia:Articles_for_deletion,Wikipedia:Substitution}.
The task to detect substituted templates is thus non-trivial:
All revisions of every template existing prior to the creation of the page in question would have to be matched against the full text of all \gls{discussion} revisions.
That is, a template detection would have to be applied before the WikiWho algorithm, because WikiWho does not guarantee to reconstruct revisions correctly.
The algorithm moreover discards any spaces.
We refrained from implementing an algorithm which would detect any substituted templates due to the associated computational overhead.
Beyond that, most templates are rarely used or even substituted in \glspl{discussion} in our experience.
This is because templates are mainly embedded in other spaces than in \glspl{discussion} such as the user or article space; cf.\ the list of Wikipedia templates~\citeW{Category:Wikipedia_templates}.
Nonetheless, there are a few templates built to be used in \glspl{discussion}~\citeW{Category:Articles_for_deletion_templates}.
They are mainly used to open such a discussion or state that it has been finished together with its outcome.
As they are a recurring part of \afds{}, we have implemented a rudimentary detection and removal of these templates.
Due to the complexity of this task, attempts to automatically build detection mechanisms from the revisions of all \afd{} templates without heavily impacting the performance failed.
Instead, we identified the most commonly used templates in \glspl{discussion} posts by grouping identical posts and counting them.
Thereafter, we built regular expressions to match and remove frequently used \afd{} templates in posts.

As per guideline~\citeW{Wikipedia:Signatures}, many editors append a signature to their posts.
There is also a bot called SineBot~\citeW{User_talk:SineBot} dedicated to adding signatures for editors who forgot to add or purposely left out their signature.
Posts by this bot alongside with other posts by registered editors whose user names ends in \enquote{Bot} are ignored because we are interested only in human communication.
This action follows the naming scheme described in the Wikipedia user name policy~\citeW{Wikipedia:Username_policy}.
Signatures should be removed for two reasons.
First, they can be customised by registered users and thereby introduce new words to their posts, which could influence the results.
Second, customised signatures as well as signatures in general can be used to identify authors.
However, classifiers should not learn to detect editors and decide upon their behaviour but make predictions only on the actual discussion contents.
Consequently, we match and remove all regular signatures as well as many customised ones.
Due to the vast possibilities of customising signatures using wikitext, HTML and CSS, we settled for a removal of many but not all customised signatures.
This design decision and its potential effects are further discussed in \autoref{s:errors}.

Performing the removal of automatically generated and non-linguistic contents after applying the WikiWho algorithm was performance motivated:
Without identifying single contributions first, the removal would have had to be done on the most recent and all previous contributions for each revision.

\section{Preprocessing of blocks}\label{ss:preprocessing_block}
Besides extracting and preprocessing the posts of \glspl{discussion}, similar has to be done for the information about blocked users retrieved from the block log.
At first, all blocks issued on anonymous users were removed as these are distinguished only by their IP address which is not unique.
This is done analogously to our post removal by anonymous editors.
Without this precaution, multiple people could issue edits from the same IP address, with some being constructive and others disruptive.
As a result of the filtering, the number of blocks was reduced from $3,332,551$ to $821,074$.

Per block, the block log contains a timestamp, the blocked user name, the user name of the administrator issuing the block together with the administrator's internal user ID, and a comment.
As we ignore anonymous editors, the user names are unique.
Hence, it is not a problem that an internal ID for blocked users is missing.
Each comment should legitimise its associated block.
We use these comments to determine whether a block is relevant for our analysis.
The comment can be freely chosen by the blocking administrator or even left blank.
Therefore, some comments are more helpful than others when trying to determine which blocks are of interest.
For example, comments such as \enquote{personal attacks} or \enquote{harassment} are the result of disruptive contributions in which an editor must have verbally attacked another user.
Conversely, \enquote{Willy} as a reason cannot be understood without a context.
Here, the comment refers to the user \enquote{Willy on Wheels} who is claimed to have created more than a thousand accounts for the purpose of vandalising Wikipedia pages~\citeW{WillyOnWheels}.
Judging from the block log alone, these claims seem realistic.

There are two options to determine which blocks should be considered based on their comments.
One is a whitelisting approach, where only blocks are extracted that match a set of words which are indicative for preceded disruptive contributions.
The other option is to blacklist words that indicate prior misbehaviour unrelated to an editor's style of communication.
For whitelisting, we considered the words \enquote{personal attacks}, \enquote{harassment}, \enquote{vandalism}, \enquote{hating} and \enquote{legal threats}.
We included \enquote{vandalism} as vandalism has disturbing effects on discussions and is thus more similar to disruptive than to constructive posts.
As the comments are freely chosen texts by administrators, they may incorrectly claim a disruptive post to be vandalism because on first sight, both might look similar and may result in a block.
Wikipedia shortcuts~\citeW{Wikipedia:Shortcut} were also considered.
These are internal links that in this case link to the respective policies.
For example the policy explaining that personal attacks should be avoided and may lead to blocks~\citeW{Wikipedia:No_personal_attacks} is abbreviated using the shortcuts \enquote{WP:NPA} and \enquote{WP:PERSONAL}.
Even with additionally respecting word stems such as \enquote{vandal} or \enquote{harass}, the whitelist approach resulted in merely $16.548$ blocks issued on registered Wikipedia editors.
The number does not take into account that some of the editors will not have been active in an \afd{} discussion at all.
Therefore, even fewer data on blocked editors would be available.
Furthermore, it disregards more freely worded and context-sensitive comments such as \enquote{\texttt{being a dick after being blocked for the same thing}}, where \enquote{the same thing} refers to an earlier block issued due to \enquote{attacks, incivility} by the same administrator on the same editor. 

For the blacklisting approach, only those blocks were removed which were unlikely to be issued because of a disruptive communication style by humans.
Blocks whose legitimation comment contained the word \enquote{bot} were removed as our interest lies in human communication and not e.g.\ in bots malfunctioning.
Naturally, blocks that were issued only for testing purposes were also ignored.
Sometimes editors want to force themselves to take a break off contributing to Wikipedia and hence request being blocked~\citeW{Wikipedia:Block_on_demand}.
As the motivation behind such a block is unlikely to be related to disruptive editing, we ignore it as well.
All blocks of editors that concerned copyright infringement have been removed.
This was done, because these claims cannot be automatically detected without matching the contributions against a great amount of external data sources.
Moreover, the detection of copyright infringement is not the goal of this thesis.
Finally, we discarded the blocking of editors associated with not citing any or solely untrustworthy sources.
This misbehaviour must have happened on actual article pages which we do not consider and thus a certain style of communication on \afd{} pages cannot be derived from it.
The blacklisting approach returned $800,737$ blocks and also yielded better results for all classifiers.

We determined $746,349$ registered editors that have been blocked at least once according to the block log with respect to our previously described criteria. 
On the other hand, $25,743,749$ registered users~\citeW{Wikipedia:Wikipedians} have never been blocked. 
In other words, merely $2.82$\% of all registered Wikipedia users have been blocked at least once. 
But of the $25,743,749$ users, there are also users who registered an account but have never been active.
However, considering all $111,759$ editors who have been active in at least one \gls{discussion}, $16,314$ or $14.6$\% of all \afd{} participants have been blocked at some time. 
Therefore, the \glspl{discussion} offer a large pool of human textual discussions that partly created enough arousal for an administrator to intervene and block an editor.

\chapter{Building an Annotated Data Set}\label{s:timeframes}
For learning and testing classifiers, we need an annotated data set.
That is, we need posts that are labelled as disruptive or constructive.
However, none of the entries in the block log is associated with a user contribution.
Thus, it is unclear, what contribution led to a block being issued.
Moreover, such a data set did not yet exist for \glspl{discussion}.
The existing annotated data set AAWD~\citep{bender_annotating_2011} was built from general Wikipedia talk pages.
It contains merely $3000$ additions of text and was annotated using data from 2008.
However, to prevent overfitting we were interested in a bigger data set.
It is also unclear, whether the language of discussions has changed within the years and thus, whether annotations from 2008 still represent current discussion styles.
At least when considering the oldest \glspl{discussion}, we found that our classifiers perform notably different as later presented in \autoref{s:results_chrono}.
Moreover, regarding the grading scale by \citet{landis_measurement_1977}, the annotators only showed \enquote{moderate agreement} with a Cohen's kappa coefficient of $0.50$.
In addition, only agreement and disagreement were labelled.
A subset of what \citet{bender_annotating_2011} identified as disagreement, matches our definition of disruptive posts.
Therefore, we decided against using AAWD.

Manually annotating posts was not an option, because it would have only been feasible with a small subset of data.
Yet, we were interested in gathering information from the analysis of large-scale textual discussions.
We assumed that disruptive behaviour must have preceded a block.
In this \namecref{s:timeframes}, we describe how we generated an annotated data set of constructive and disruptive posts with this assumption in mind.

Assuming that either all or only the last $n$ posts by editors before they were blocked led to said block, may result in false predictions.
For example, there could exist a scenario in which formerly constructive editors took a longer break from editing and then were blocked due to their new contributions.
As they were not blocked for their old posts, it is unlikely that these had been disruptive.
Yet, the old posts could then be classified as such in our data set.
Therefore, we chose to assume that the posts made by editors shortly before they blocked were the cause of it.
This raises the question: How much time should be considered before and up to a block, in which we assume posts by the blocked authors to be disruptive?
Hereafter, we call this timespan a \emph{timeframe}.

For every post, we calculated the time between the creation of this post and the next time its author was blocked.
Posts whose authors were never blocked afterwards, were considered to have been constructive regardless of any applied timeframe.
This data was then used to run tests using different timeframes and compare the performance of our full text classifiers.
The classifiers are described in more detail in \autoref{s:classifiers}.
In the then ensuing section, the metrics we chose for evaluating classifier performance throughout this thesis are presented.
Successively, the test setup and results for determining the best timeframe are shown.
Finally, some analysis is done on the newly generated annotated data set without the use of classifiers but by inspecting the term occurence frequencies.

\section{Classifiers}\label{s:classifiers}
For choosing a timeframe and for our later analyses, we decided to use a support vector machine (short: SVM), a na\"{\i}ve Bayes (short: NB) classifier and a language model (short: LM) classifier.
Regarding the NB classifier and support vector machine, two commonly used classifiers have been selected.
SVMs are known to be well-suited for text classification and to perform better than NB classifiers \citep{joachims_text_1998,yang_re-examination_1999} in this context.
Therefore, we expected to see similar performance differences for our corpus.
The language model classifier was chosen because it fits our analysis well in which we want to better understand language used in discussions.
Both, the LM and the NB classifier, are probabilistic models.
They are the same when considering their simplest implementations~\citep[cf.\ e.g.][]{peng_augmenting_2004,mccallum_comparison_1998}.
Therefore, we expected them to perform at least comparably to each other.
Yet, our LM classifier does not consider words independently like an NB classifier but instead uses a small history of words.
Due to inspecting words in a local context, we suspected that an LM classifier may also outperform an NB classifier.
However, the results, as later presented in \autoref{s:results}, show that the NB classifier performs better than the LM classifier.
The SVM on the other hand did indeed outperform the other classifiers.

After preprocessing, the posts containing English words, markup, numbers, punctuation marks and other symbols had been reduced to words separated by one space symbol each.
We declare the set of terms as the vocabulary $V \coloneqq \left\{t_1,t_2,\dotsc,t_n\right\}$.
In this sense, posts are elements of the set of sequences $S \coloneqq V^{*}$ constructable from all terms.
A particular post will be denoted as sequence $s = \left(w_1 w_2\dots w_m\right)$ with a word $w_i \in V$.
Thereby, it is possible for $w_i = w_j$ with $i \neq j$, i.e.\ a post can contain multiple occurrences of the same term.
The set of sequences on which the classifiers are learnt will be referred to as $S_L$ and the one on which the classifiers are tested as $S_T$.
Thus, it holds that $S_L \subset S$ and $S_T \subset S$ with the possibility of there being an overlap between $S_L$ and $S_T$.
Such an overlap is likely for typical \gls{discussion} posts such as \enquote{\texttt{delete as per nom}}, which expresses that the author agrees with the proposed deletion and the reason given by the creator of this \gls{discussion}.
The vocabularies containing the unique terms from the sequences on which the classifiers are learnt and tested are denoted as $V_L$ and $V_T$ respectively.

The SVM and the NB classifier were both given each post as a \emph{term frequency-inverse document frequency} (short: \emph{tf-idf}) feature vector.
A high tf-idf value indicates an extra\-ordinarily high presence of a term within a post (term frequency) compared to its appearance in the complete corpus (document frequency).
When a term appears in many posts, its document frequency is high.
Using the inverse ensures that the tf-idf value shrinks consequently to penalise common terms like conjunctions that do not represent a post's content well.
This allows retaining stop words in posts, which are essential for the function word analysis, while reducing their otherwise heavy impact.
In small tests, we found little difference between the performance of either using a tf-idf or a tf feature vector.
In the context of \textsf{RQ4}, we decided to use a tf-idf feature vector for finding terms other than function words that are characteristic for disruptive or constructive messages.
This did not interfere with determining the effects of function words, as required by \textsf{RQ3}, because separate classifications that solely considered function words were done as well.

Given the great amount of posts to analyse, the feature vectors for the SVM and NB classifiers have a high dimensionality while being sparse.
Thus, we applied stemming e.g.\ to reduce conjugated verbs to their word stem.
In many different classification tasks, very short words like one letter words or stop words are being pruned to reduce the vector size.
We refrain from doing so to ensure that the effect of function words, which include stop words and are often very short such as \I{} or \enquote{a}, can be studied.
RapidMiner's snowball stemmer was used for the task of stemming as its result were slightly better than the Porter and equal to the Lovins stemmer which are also included in RapidMiner.
The following \namecrefs{ss:svm} present the general concepts of our chosen classifiers individually together with some design choices we made.
We used RapidMiner\footnote{\url{https://rapidminer.com/} --- last accessed 22 January 2016, 15:30} Studio $5.3$, a software frequently used for machine learning and data mining tasks, for our SVM and NB classifications.
The LM classification process was developed using the Generalized Language Model Toolkit\footnote{\url{https://github.com/renepickhardt/generalized-language-modeling-toolkit/tree/cebbff8} --- last accessed 22 January 2016, 15:30}.

\subsection{Support Vector Machine}\label{ss:svm}
Support vector machines \citep{cortes_support-vector_1995} are supervised learning models for solving binary classification tasks.
They treat elements of a class---in our case each represented by a tf-idf feature vector---as points in a high-dimensional space.

\begin{figure}
  \begin{minipage}{0.45\textwidth}
    \centering
    \includegraphics[width=1\linewidth]{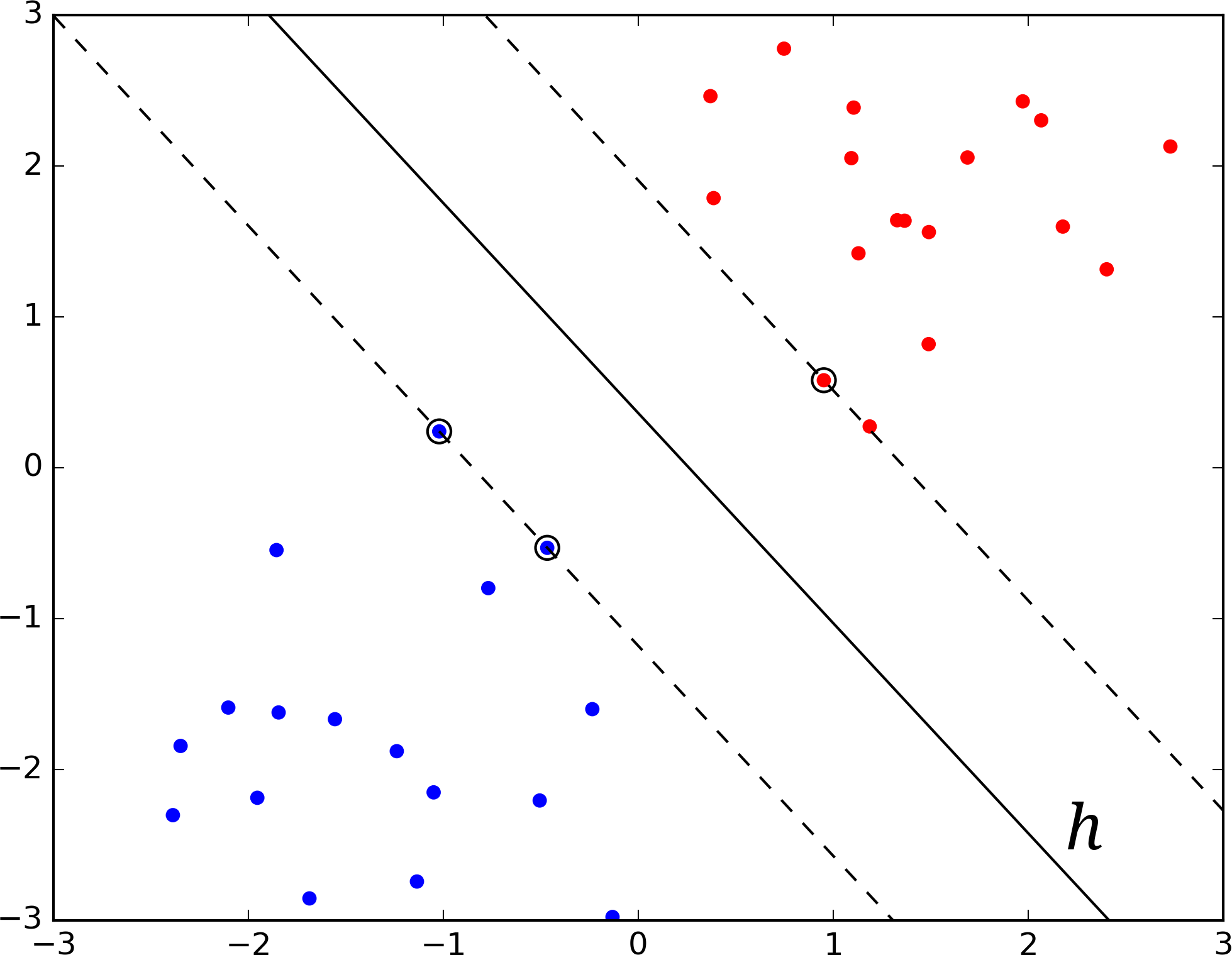}
  \end{minipage}
  \begin{minipage}{0.45\textwidth}
    \centering
    \includegraphics[width=1\linewidth]{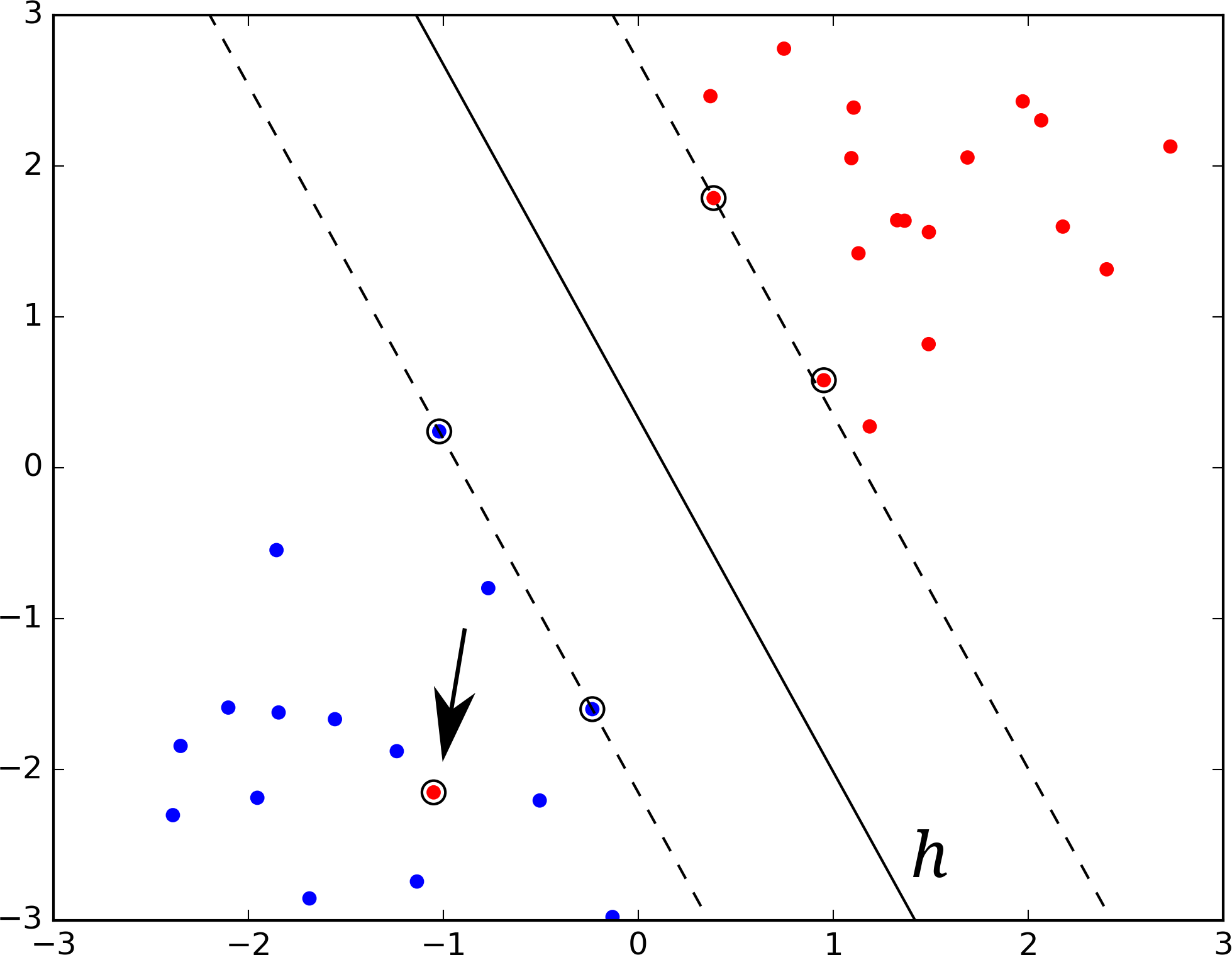}
  \end{minipage}
  \caption{Two elements of different classes are separated by an optimal hyperplane $h$ as the margins (dashed lines) are maximised.
  Circled elements are support vectors.
  In the right plot, the use of a soft margin allows tolerating an error, indicated by the black arrow.}
\label{fig:svm1}
\end{figure}
Using training data, an SVM tries to calculate the optimal hyperplane $h$ for separating the points of both classes as shown in \autoref{fig:svm1}.
A hyperplane is optimal when it maximises the margin between the elements of each class closest to the hyperplane.
Only the three circled elements in the left plot of \autoref{fig:svm1} determine the margins and as a consequence thereof the hyperplane.
Thus, they are called \emph{support vectors}.

In many cases, the elements of the two classes are not linearly separable without errors.
Then, a soft margin hyperplane can be used that tolerates a minimal amount of errors.
This is shown in the right plot of \autoref{fig:svm1}.

\begin{figure}
  \begin{minipage}{0.45\textwidth}
    \centering
    \includegraphics[width=1\linewidth]{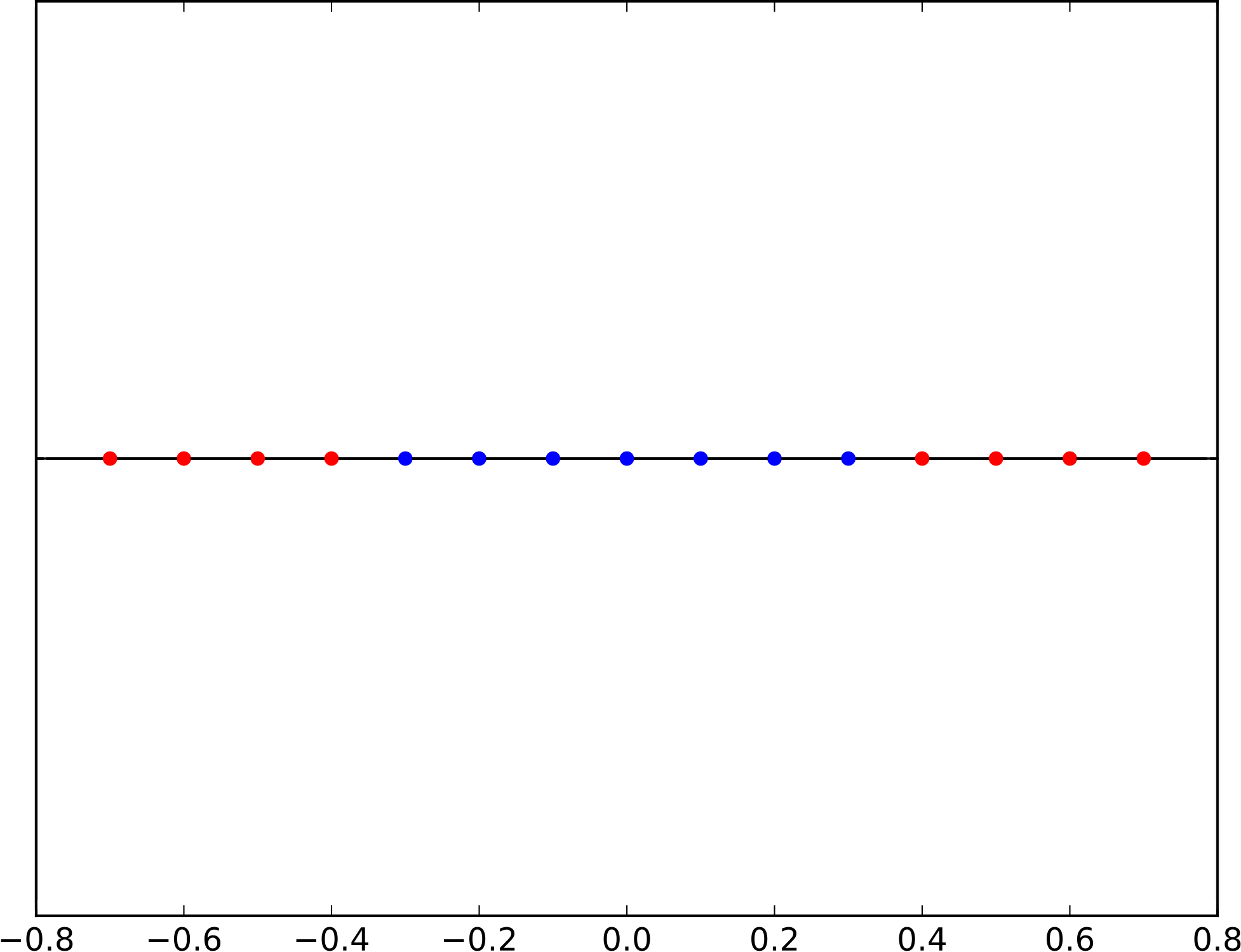}
  \end{minipage}
  \begin{minipage}{0.45\textwidth}
    \centering
    \includegraphics[width=1\linewidth]{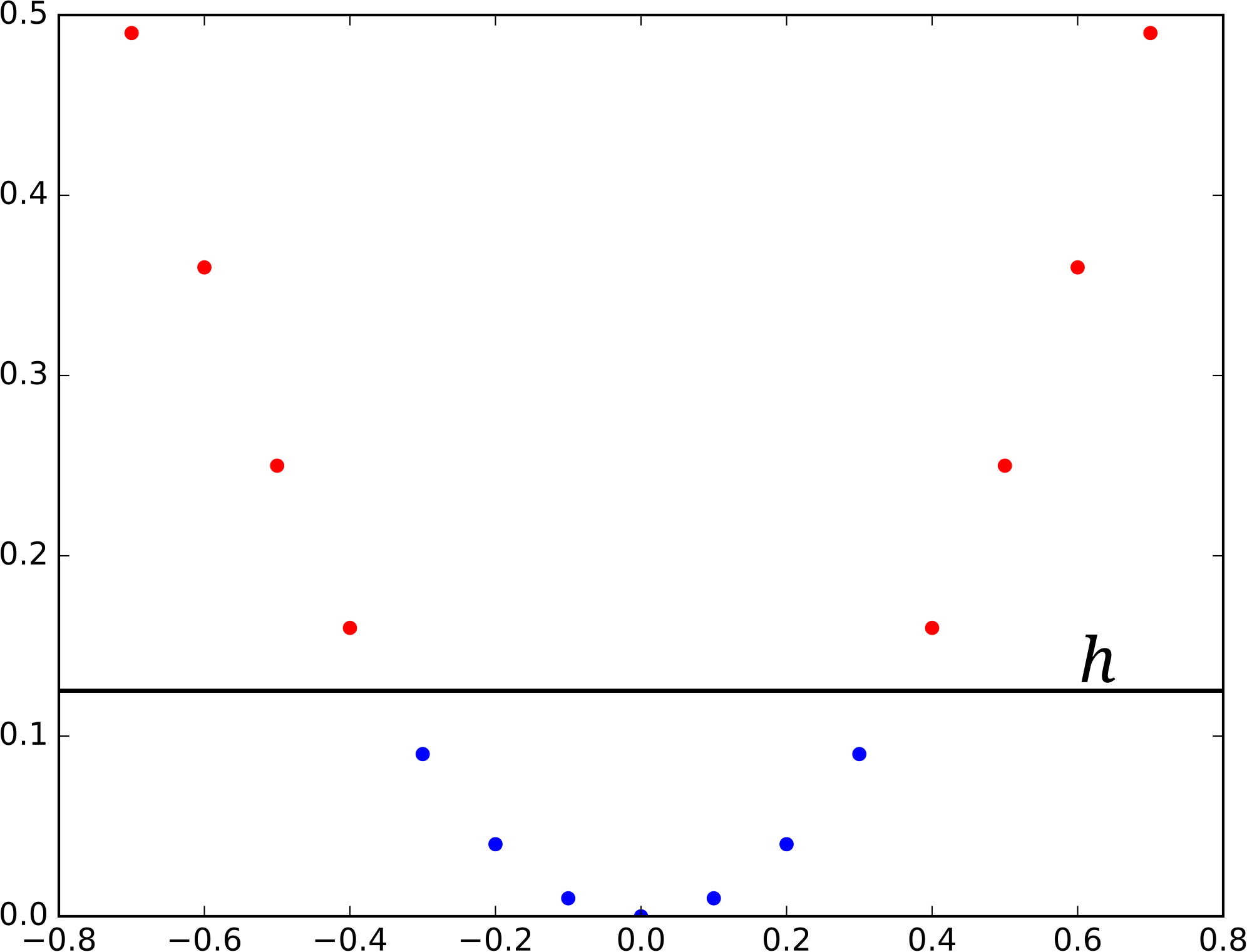}
  \end{minipage}
  \caption{The left plot shows a dataset of two classes which are not linearly separable in a one-dimensional space.
  In the right plot, the data has been transformed into a two-dimensional space by using a mapping function $\phi\colon \mathbb{R} \to \mathbb{R}^2,\ x \mapsto (x, x^2)$. The classes can thus be separated by a hyperplane $h$.}
\label{fig:svm2}
\end{figure}
In cases where the data is not linearly separable as depicted in the left plot of \autoref{fig:svm2}, \citet{cortes_support-vector_1995} propose the transformation of the vectors into higher features spaces.
The right plot of \autoref{fig:svm2} illustrates this by using a mapping function $\phi$.
Instead of transforming all points into a higher feature space, kernel functions are used.
They can calculate a dot product of two points in a higher dimensional space as needed for determining the hyperplane as well as for the actual nonlinear classification.
The interested reader may refer to \citet{cortes_support-vector_1995}.
Our classifier uses a dot kernel $k(u,v) = u \cdot v$, say the inner product of $u$ and $v$.
Due to limited time for this thesis, other kernels were not tested.

\subsection{Na\"{\i}ve Bayes Classifier}
A na\"{\i}ve Bayes classifier is a probabilistic classifier.
We use a multinomial NB classifier in this thesis.
Such a classifier is a unigram model, meaning that it determines probabilities for words individually instead of considering their usage context~\citep{mccallum_comparison_1998}.
\autoref{e:nb} shows the general concept of a multinomial NB classifier \citep[cf.~e.g.][]{mccallum_comparison_1998} using tf-idf weights~\citep{rennie_tackling_2003}.
\begin{equation}\label{e:nb}
  \gamma_{\mathrm{NB}}\left(s \in S_T\right) = \operatorname*{arg\,max}_{k \in \left\{0, 1\right\}} \left(P\left(C_k\right)\cdot\prod_{w \in s} P{\left(w\mid C_k\right)}^{\operatorname{tf-idf}\left(w\right)}\right)
\end{equation}
The classifier is applied to a specific post $s$ from the testing data with $C \coloneqq \{0,1\}$ being the prediction classes where $1$ indicates a post that led to a block and $0$ one that did not.
Essentially, one classifier is learnt per class and their predictions for $s$ are compared against each other.
The $\operatorname{argmax}$ operator returns the class of which the words in $s$ had the highest probability to belong to.
$P\left(C_k\right)$ is the general probability for any sequence to belong to class $C_k$.
Using balanced data, the probability is $p = 0.5$.

A zero probability would result in the whole product becoming zero.
Due to the sparsity of the feature vector, zero probabilities are likely.
Laplace smoothing solves this problem by adding $1$ to the denominator and the number of unique words in the training data $|V_{L}|$ to the nominator \citep{cestnik_estimating_1990}: 
\begin{equation}\label{e:laplace}
  \widehat{P}_{\operatorname{Laplace}}{\left( w \mid C_k \right)}^{tf\left(w,s\right)} = \frac{\operatorname{tf}\left(w, S_{L_{C_k}}\right) + 1}{\sum_{v \in V_{L}}\big( \operatorname{tf}\left(v,S_{L_{C_k}}\right) + 1\big)}
\end{equation}
For illustration purposes, \autoref{e:laplace} uses term frequency instead of tf-idf for explaining an estimation of $P\left(w \mid C_k\right)$ using Laplace smoothing.
$S_{L_{C_k}}$ is the set of learnt sequences of class $C_k$.
The term frequency $\operatorname{tf}(x, S_{L_{C_k}})$ is the summed up occurrence frequency of the term $x$ in all sequences of $S_{L_{C_k}}$.
In this estimation, the term frequency of $w$ is set into relation to that of the learnt terms $v \in V_L$ while adding $1$ to prevent a result of zero~\citep{kibriya_multinomial_2005}.
With big enough training data, the addition's effect on the estimated probabilities becomes negligible.
Another option would be Lidstone smoothing which is a generalisation of Laplace smoothing using any value instead of just $1$.
However, using Lidstone smoothing for the NB classifier was not an already available option in RapidMiner.

\subsection{Language Model Classifier}
The third classifier uses the 4-gram language model with modified Kneser-Ney smoothing introduced as generalised language model by \citet{pickhardt_generalized_2014}.
An $n$-gram language model estimates a probability for a sequence.
It does so by analysing each word's probability given the local history of the last $n - 1$ successive words prior to the current word.
It is therefore a probabilistic model.
The decision to only consider $n$ words of a sequence follows a Markov assumption so that full sequences' probabilities are only approximated and thus computational efforts are reduced.
Simultaneously, it is a necessity as it is impossible to have learned all potential sequences.
A trivial language model will therefore calculate a sequence's probability by calculating the conditional probabilities of the words $w_i \in s$ as shown in \autoref{e:lm}.
\begin{equation}\label{e:lm}
  P\left(s \in S_T\right) = \prod_{i = 1}^{\lvert s \rvert} P\left(w_i \mid w_{i-n+1}\ldots w_{i-1}\right)
\end{equation}

Such a trivial implementation however would result in zero probabilities for unseen terms.
Likewise, the confidence of estimations on rare terms would be low.
Both cases are likely to occur, considering that human languages follow Zipf's law and thus most words in a text corpus appear only a few times~\citep{newman_power_2005}.
To overcome the problems introduced by data sparsity, smoothing techniques are applied.
The language model applied to the \glspl{discussion} uses a modified Kneser-Ney smoothing to solve this task.
Essentially, the smoothing interpolates higher order with lower order models; using relationally growing discounts for $n$-grams that occur more frequently.
The interested reader may refer to \citet{chen_empirical_1999}.

\citet{pickhardt_generalized_2014} embodied another concept called \emph{skip $n$-grams}.
Besides the interpolation with lower order models as is done in modified Kneser-Ney smoothing, wildcards are applied to $n$-grams.
A $3$-gram model would thus determine $P\left(w_3 \mid w_1 w_2\right)$ by not only interpolating with
$P\left(w_3 \mid w_2\right)$ and
$P\left(w_3 \right)$ but also with
$P\left(w_3 \mid w_1 \_\right)$ where the \enquote{$\_$} symbol indicates a wildcard.
This reduces data sparsity problems for higher order models and hence allows a language model to learn relations between words which are not successive~\citep{goodman_bit_2001}. 
For example, if a $3$-gram model would be learnt on the sentences \enquote{I love you} and \enquote{I like you}, skip $n$-grams would enable it detecting a relation between \I{} and \enquote{\texttt{you}} separated by a single word.
The model thus learns a context and would calculate a higher probability for the previously unseen sentence \enquote{I envy you} than a model that does not use skip $n$-grams.

Similar to the approach of the multinomial NB classifier, we train two separate language models.
One is learnt on posts that led to a block and another on posts that did not.
The class prediction of a statement $s$ is then made by determining the better performing model for it.
To evaluate the performance, we use the perplexity $e^H$ with $H$ as given in \autoref{e:ppl} where $\widehat{P}_{\operatorname{GLM}}$ is the generalised language model by \citet{pickhardt_generalized_2014}.
Perplexity~\citep{gibbon_handbook_1997} is a common metric for evaluating language models with lower values indicating a better model.
\begin{equation}\label{e:ppl}
  H\left(s \in S_T\right) = {\frac{- \sum_{i = 1}^{\lvert s \rvert} \log\left(\widehat{P}_{\operatorname{GLM}}\left(s\right)\right)}{\lvert s \rvert}}
\end{equation}

\section{Evaluation Metrics}\label{s:metrics}
Accuracy, precision, recall, area under the curve (short: AUC) and F~score were chosen as evaluation metrics.
They are commonly used for binary and text classification evaluation~\citep{sokolova_systematic_2009}.
This \namecref{s:metrics} introduces these metrics
and explains how they will be utilised in our classification evaluations.

Accuracy is the ratio of correct predictions compared to the number of all predictions.
It is the only of our chosen metrics that incorporates the correct prediction of both disruptive and constructive posts.
The others can be calculated for positive predictions---posts that led to a block---as well as negative predictions which are posts that did not lead to a block.
This thesis' focus is the detection of disruptive posts for determining the effects that function words and \iyms{} have on this task as well as other terms.
Thus, the positive predictions are of special interest.
Positive precision is the probability that a post, which has been predicted to be disruptive, truly was disruptive.
The negative equivalent is often also called false omission rate and is calculated analogously.

Positive recall is the probability that given a disruptive post, it is predicted as such.
Negative recall is also called fall-out and is calculated as the probability that given a constructive post, it is predicted as such.

In accordance with our goals, a high positive precision is more important than a high positive recall for discovering language characteristics.
That is, a classifier might predict fewer disruptive posts but these predictions would then be correct more often.
Therefore, characteristic terms of disruptive posts can rather be retrieved from a classifier with high prediction.

A high precision or recall can easily be achieved by a bad classifier:
If a classifier only predicts a single disruptive post but does so correctly, it will achieve a positive precision of $100$\%.
Likewise, if a classifier predicts every post to be disruptive, it will achieve a positive recall of $100$\%.
However, these two perfect values negatively influence each other and cannot coexist when a classifier uses this na\"{\i}ve approach.
To capture this, the F~score can be used as it incorporates both values.
We use the commonly utilised F$1$ score which is the harmonic mean of precision and recall:
\begin{equation}
  \operatorname{F1} = \frac{2 \cdot \textrm{precision} \cdot \textrm{recall}}{\textrm{precision} + \textrm{recall}}
\end{equation}
Consequently, there is a positive and a negative F$1$ score.

The AUC is the area under the receiver operating characteristics (short: ROC) graph.
This graph's $x$-axis contains the false positives rate whereas the $y$-axis contains the true positives rate.
All points are sorted by prediction confidence from highest to lowest.
Whereas the other metrics are highly affected by the class distribution, ROC curves are insensitive to heavily skewed distributions.
Naturally, when balancing the data, this benefit becomes irrelevant.
The AUC describes a classifier's ability to correctly predict disruptive posts while ignoring any constructive posts.
Theoretically, one could also plot false negatives on the $x$-axis and true negatives on the $y$-axis.
However, this is uncommon and not in our interest.
Therefore, we do not distinguish between a positive and a negative AUC\@.
The interested reader may refer to \citet{fawcett_introduction_2006} for an extensive description of both ROC and AUC\@.

Determining the performance of the classifiers with different timeframes is done on balanced data.
Likewise, all later tests but one are executed on balanced data as well.
Thus, accuracy will be used as metrics for the overall performance of classifiers.
As for judging the performance of classifying disruptive posts, the positive F$1$ score is of special interest.
Similarly, the AUC will give information about the performance of disruptive post classification in terms of prediction confidence.

\section{Deciding on a Timeframe}\label{s:timeframe_decision}
For constructing an annotated data set, a timeframe must be chosen.
Multiple tests were run using our classifiers on data with different timeframes.
The arithmetic mean of their performances was evaluated to ascertain the best timeframe length.
As the classifiers run for multiple hours and as there are many testable timeframes, our testing was limited to a few timeframes.
In this \namecref{s:timeframe_decision}, we explain which timeframes were chosen to be tested and which of these led to the best overall performance.

\begin{figure}
  \begin{minipage}{0.45\textwidth}
    \centering
    \includegraphics[width=1\linewidth]{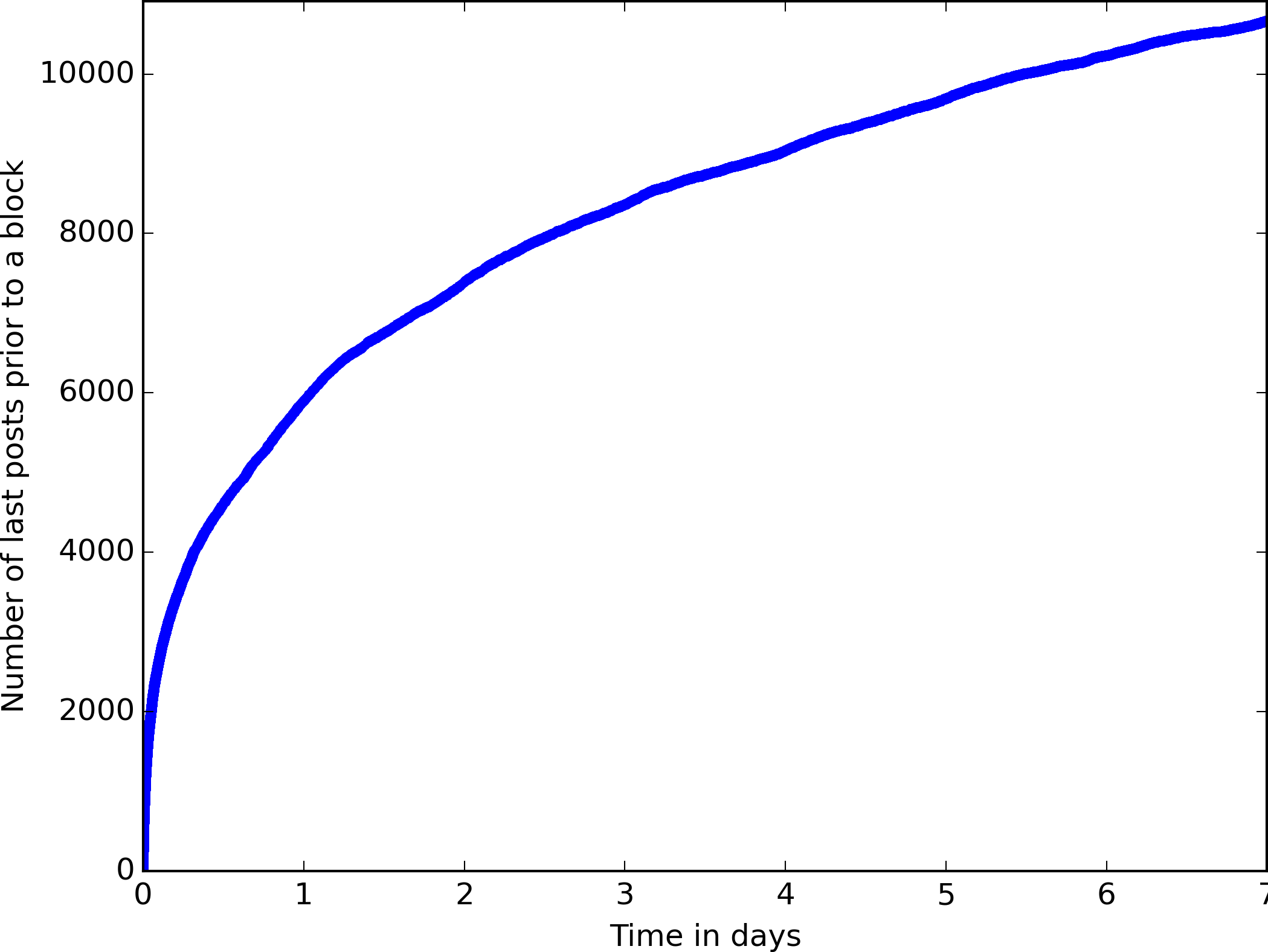}
  \end{minipage}
  \begin{minipage}{0.45\textwidth}
    \centering
    \includegraphics[width=1\linewidth]{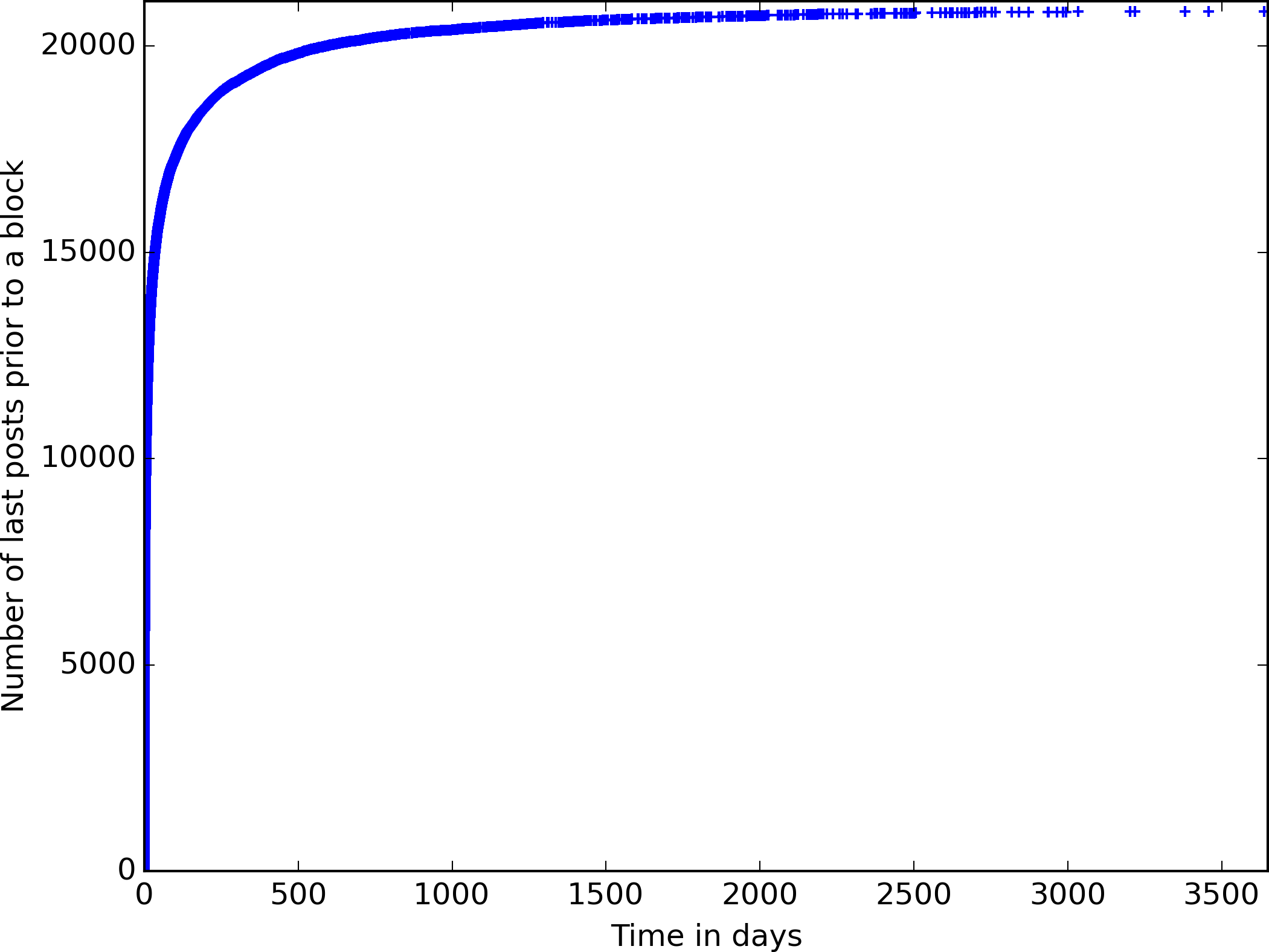}
  \end{minipage}
  \caption{Number of last posts in an \gls{discussion} prior to their authors being blocked within a given timeframe in days.
  The left graph shows the first seven days of the right graph in more detail.}
\label{fig:deltas}
\end{figure}
To not overfit the data and thus to allow detecting general communication patterns, we were looking for timeframes in which at least $10,000$ blocks were issued on editors who contributed to an \gls{discussion} at least once.
Consequently, the lowest timeframe was set to $13$ hours in which there were $10,067$ such blocks.
We inspected the last post an editor had made in an \gls{discussion} before they were blocked with respect to the post's temporal distance to said block.
\autoref{fig:deltas} shows how a growing timeframe increases the number of these last posts which would be considered disruptive.
The number of posts quickly rises in the beginning with its growth speed decreasing.
This suggests that editors who disruptively act in \glspl{discussion} are blocked soon after their last disruptive posts.
According to the corresponding Wikipedia policy~\citeW{Blocking_policy}, blocks should only be issued to ensure a productive and disruption-free environment.
As such, they should not be used for punishing misbehaviour of users and thereby should not be issued on old posts which no longer affect active discussions.
Thus, the longer the considered time between the last post by an editor in an \gls{discussion} and their block, the likelier it is for the block to be related to a contribution elsewhere on Wikipedia.
\glspl{discussion} should typically be closed after seven days~\citeW{Wikipedia:Articles_for_deletion}.
Therefore, $6$ days was chosen as our longest timeframe for testing, so that an \gls{discussion} would still be active and benefit from an editor being blocked.
The other timeframes were decided to be $1$, $1.5$, $2$, $2.5$, $3$, $4$ and $5$ days long following the assumption that longer timeframes will eventually lead to worse data quality.

Within a timeframe of $13$ hours, only $0.70$\% of all posts are disruptive. 
A classifier that would always predict a contribution to be constructive would yield a good overall performance.
Thus, the data was balanced to prevent this effect.
As a consequence thereof, the amount of data a classifier has to process is reduced immensely which makes the process faster.
Otherwise, the unbalanced data would have to be sampled to a feasible subset.
Due to the then small amount of disruptive posts, these would likely have been overfitted.

To make the results between different timeframes more comparable, the classifiers were always learnt on the same sample size.
With longer timeframes, the number of assumed disruptive posts increases at the cost of the amount of constructive posts.
Therefore, all available disruptive posts for the timeframe of $13$ hours were used and for later timeframes, the same amount was randomly sampled.

\begin{figure}
  \centering
  \includegraphics[width=0.7\linewidth]{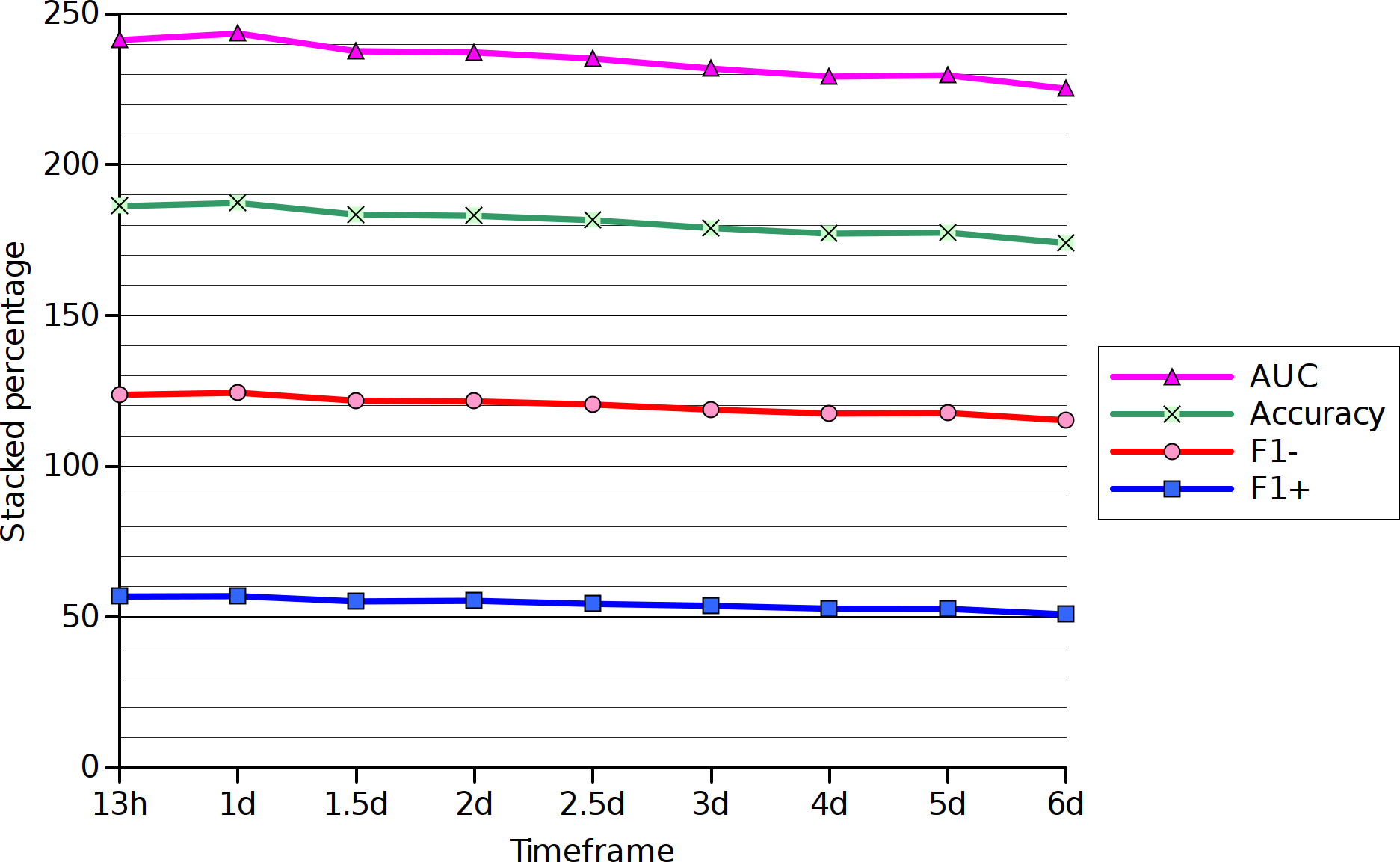}
  \caption{This stacked line plot shows the average performance of the classifiers with the various timeframes.
  The values are the arithmetic mean of the results from the SVM, NB classifier and LM classifier.
  All are given as percentages.
  }
\label{fig:timeframeResults}
\end{figure}
As can be seen in \autoref{fig:timeframeResults}, the timeframe of $1$ day returned the best results overall.
The figure shows how the values for all metrics peak at the $1$ day timeframe and worsen the longer the timeframe becomes.
All specific values used to generate the plot in \autoref{fig:timeframeResults} can be found in \autoref{t:timeframeResults_confusionMatrix} together with precision and recall values.
The positive precision value is the only one which was not the highest using a $1$ day timeframe.
It was $49.81$\% for $13$ hours and $49.29$\% for $1$ day.
Nevertheless, this is negligible considering that the $1$ day timeframe resulted in the greatest positive F$1$ score and accuracy.
Thus, successive tests were performed using $1$ day as timeframe length.

Due to the missing link between contributions and blocks, we made the assumption that posts shortly before a block were disruptive.
However, there are scenarios where this assumption is false.
This problem will be discussed in \autoref{s:errors}.

\section{Data Analysis}\label{s:dataAnalysis}
Besides through classification, the data was also manually inspected.
For this, all \glspl{discussion} have been separated into single posts, which have been preprocessed, e.g.\ by removing wikitext markup.
All posts have been labelled as either being disruptive or constructive according to the $1$ day timeframe.

\begin{figure}[ht!]
  \centering
  \includegraphics[width=0.9\textwidth]{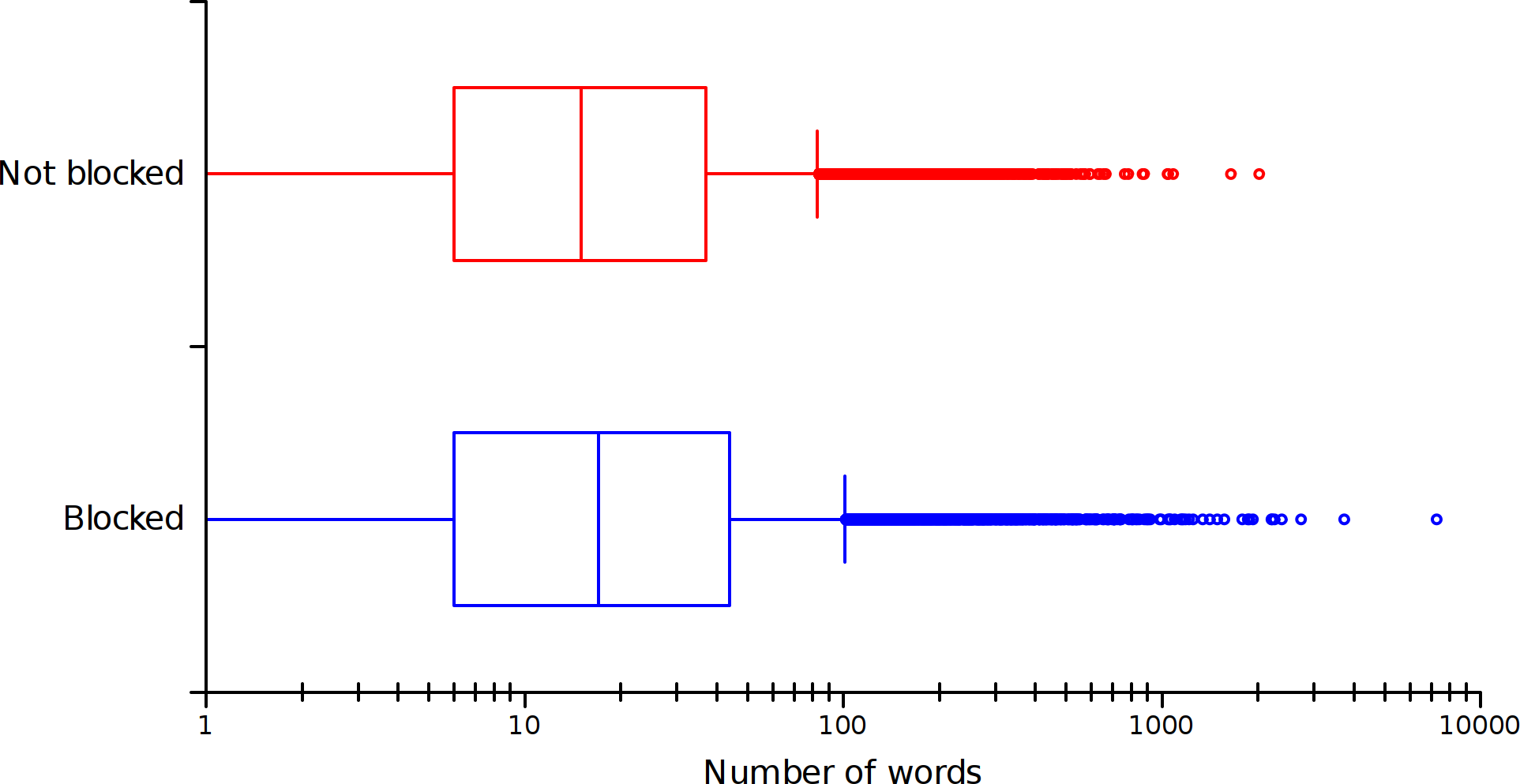}
  \caption{This log-scaled box plot shows the differences in lengths of posts which were assumed to have led to a block and those which were not.
  It illustrates that both data sets are skewed and have outliers consisting of many words.
  In spite of that, disruptive posts contain more words on average.}
\label{fig:dataDifference}
\end{figure}

First, the average length of disruptive and constructive posts was compared.
The length was defined as the number of words a post contained.
We considered all posts labelled disruptive within the $1$ day timeframe and randomly sampled the same number of constructive posts.
Here, we only discuss concrete values of a single subset of all available posts because of performance reasons.
Analysing and then plotting the length of $3,467,402$ posts was not an option.
Nonetheless, we repeated this test with different samples and found similar results.
In general, for every data set we found the average post length always to be significantly longer for disruptive posts.
The average disruptive post is $23.80$\% longer than a constructive one. 
That is, the average disruptive post contains $41$ words, whereas the average constructive post only contains $31$.
However, with the medians being $17$ and $15$ respectively, the data is heavily skewed, i.e.\ it contains multiple outlier posts with many words.
The box plot in \autoref{fig:dataDifference} illustrates this.

In general, shorter posts seem to be less disruptive than longer ones.
An explanation could be that many short posts solely state to keep, merge or delete the article.
They often lack an explanation why the article should be kept, merged or deleted.
Such a post is an unfounded expression of one's opinion about the quality and relevance of an article.
Naturally, this alone is no reason for being blocked.
Personal attacks and similar disruptive comments can also be made using few words but the data suggests that such posts are likelier to be verbose.

\begin{table}
  \begin{tabularx}{0.9\textwidth}{p{2cm} | Y Y}
    \bottomrule
    \centering{\textbf{term}} & \textbf{share of words from disruptive posts (\textperthousand)} & \textbf{share of words from constructive posts (\textperthousand)}
    \\\toprule
    \centering{fucking}  & \textbf{0.06}  & 0.00 \\
    \centering{fuck}     & \textbf{0.06}  & 0.01 \\
    \centering{shit}     & \textbf{0.09}  & 0.01 \\
    \centering{i}        & 6.40  & \textbf{10.70} \\
    \centering{you}      & \textbf{10.64} & 4.52 \\
    \centering{me}       & \textbf{2.43}  & 1.20 \\
    \centering{my}       & \textbf{3.00}  & 1.68 \\
    \centering{your}     & \textbf{3.05}  & 1.25 \\
    \centering{myself}   & \textbf{0.22}  & 0.13 \\
    \centering{yourself} & \textbf{0.20}  & 0.10

  \end{tabularx}
  \caption{The table shows how commonly a term appears in disruptive or constructive posts.
  A bold font indicates that the term appears more frequently in that class.}
\label{t:words}
\end{table}
\autoref{t:words} shows how often certain terms appeared in the disruptive or constructive posts in relation to all other words of the posts.
All $3,467,402$ posts have been considered for these values.
Common swear words like \enquote{fucking}, \enquote{fuck} and \enquote{shit} are hardly used in disruptive posts.
With only $6.39$\textperthousand\ of disruptive posts containing any of the three swear words, the few posts that do, contain the terms multiple times. 
However, when they are used, they are quite expressive.
For example, the term \enquote{shit} is $9.43$ times likelier to appear in a disruptive than in a constructive post.
In sum, with these swear words alone, a small recall but a high precision could potentially be achieved.
Hence, a collection of swear words would be unlikely to suffice for identifying many disruptive posts.

The use of \iyms{} can also be investigated without using a classifier.
As initially stated, correctly detecting these messages is out of the scope of this thesis.
Instead, terms indicative of \iyms{} can be counted like \I{}, \Y{} but also \enquote{myself}, \enquote{yourself} and others.
The relation of word usage in disruptive to constructive posts is a less stark contrast for these terms as can be seen in \autoref{t:words}.
However, they are a lot more frequently used: $36.60$\% of all disruptive and $26.89$\% of all constructive posts contain at least one of the terms \I{} and \Y{}. 
Additionally considering the terms \enquote{me}, \enquote{my}, \enquote{your}, \enquote{myself} and \enquote{yourself} increases this to $40.13$\% and $29.37$\% respectively. 
$43.93$\% of all disruptive and $32.11$\% of all constructive posts contain at least two of the terms.
Therefore, if these terms and especially \I{} and \Y{} were strong indicators for constructiveness or disruptiveness, they would improve the recall noticeably.

In accordance with the assumed effect of \iyms{}, \I{} appears significantly more often in constructive than in disruptive posts.
Likewise, \Y{} appears more than twice as often in disruptive posts.
Surprisingly, the terms \enquote{me}, \enquote{my} and \enquote{myself} appear more frequently in disruptive posts although they are rather typical for \ims{}.
A possible explanation could be that \afds{} benefit from objective posts.
These three terms, however, are often used in a subjective context to express the author's personal feelings or thoughts.
When they are expressed towards another editor, they can be framed as a personal attack, which would make them disruptive.

The relation of the occurrence frequency of \I{} to \Y{} also seems to be expressive.
That is, a constructive post contains on average $2.37$ times more \I{} than \Y{}.
Disruptive ones on the other hand have an average relation of $0.60$, i.e.\ \Y{} appears nearly twice as often as \I{} in a disruptive post.

On average, there is a significantly higher use of the term \I{} in constructive posts as questioned in \RQ{1} and the opposite holds true for \RQ{2}.
Thus, the results of this simple data analysis support the research questions when exclusively considering the terms \I{} and \Y{} as indicators for \iyms{}.
Interestingly, other terms that are also likely to encode \iyms{} are not in favour of \RQ{1}.
The results of the later classifications will not be able to support \RQ{1} especially in regard to the use of the term \I{}.
\RQ{4} raised the question which other terms are characteristic for constructive or disruptive messages.
This analysis implies that swear words are a strong indicator for disruptive messages although they are only rarely used.

\chapter{Test Setup}\label{s:testsetup}
This \namecref{s:testsetup} describes the tests and their analysis.
The first section introduces the goals of the later tests.
It is thereby explained how our analysis could answer the research questions.
\autoref{s:setupSW} describes how the tests were conducted in general.
Furthermore, our own approach is introduced which allows the analysis of a short history of an editor's posts.
As the results in \autoref{s:results_sw} show, the classifiers perform mostly worse with this approach.
Finally, \Cref{s:validation} describes our model validation and how the annotated data set was sampled for the tests.

\section{Goals}
The motivation of this thesis was to study the effects of function words as well as \iyms{} on the disruptive character of messages in textual discussions.
For this, we employed the classifiers presented in \autoref{s:classifiers} on \glspl{discussion} as our model.
In respect to function words, this means comparing the performance of full text classifiers against that of classifiers only considering function words.
\textsf{RQ3} raised the question whether function words suffice for distinguishing disruptive and constructive messages.
If the differences between the function words and full text classifiers were negligible, this question would be answered in the affirmative.
This would also be the case, if the function words classifiers performed notably worse but still significantly better than a random classifier.
Naturally, little difference in the classifiers' performances can only answer the question when the full text classifiers already vastly outperformed a random classifier.

We have not developed elaborate measures to detect \iyms{} due to the complexity of this task.
Instead, as these concepts are frequently reduced to the usage of \I{} and \Y{}, we solely inspect the impact of these terms on the classifiers.
If the concept of \iyms{} would hold true on \afds{}, \I{} would appear among the most important features indicating that a post did not lead to a block.
This complies to \RQ{2}.
Likewise, \Y{} would be among the most important features which indicate a disruptive post, which complies to \RQ{1}.
It will be evaluated on the SVM and NB classifier but not on the LM classifier as it returns metrics for $n$-grams instead of single terms.
But more importantly, the LM classifier did not perform significantly better than a random classifier.
To determine other terms, which are characteristic for either disruptive or constructive posts and thus to potentially answer \textsf{RQ4}, the heaviest weighted features of the SVM and NB classifier are inspected.

\section{Test Setup and the Sliding Window Approach}\label{s:setupSW}
After having chosen a timeframe of $1$ day and preprocessing the data, a new balanced annotated data set was sampled from the annotated data set created, containing all $21,213$ disruptive posts and equally many, randomly sampled constructive posts.
The corpus used for determining the best performing timeframe was not reused as it contained only a subset of elements from both classes.
At first, all three classifiers were applied to the full text.
In a second run, the classifiers only considered function words.
We relied on the list of function words for the English language compiled by Leah Gilner and Franc Morales\footnote{\url{http://www.sequencepublishing.com/academic.html} --- last accessed 10 October 2015, 22:00}.
It includes pronouns (e.g.\ \I{}, \Y{}), prepositions (e.g.\ \enquote{for}, \enquote{at}), conjunctions (e.g.\ \enquote{but}, \enquote{and}), auxiliary verbs (e.g.\ \enquote{can}, \enquote{will}), determiners (e.g.\ \enquote{her}, \enquote{his}) and quantifiers (e.g.\ \enquote{some}, \enquote{none}).

In addition to those classifications that consider posts individually, tests have been made which consider parts of an editor's post history.
Subsequently, the later approach will be called \emph{sliding window} and the original one \emph{independent posts} approach.
In this approach, posts by the same author made shortly after each other are merged into a single new post.
A merged post is compiled from posts where the oldest and newest ones are not further apart than $1$ day.
We call this timespan a \emph{window}, which slides over posts and creates the longest merged post using every post as the oldest one once.
The time difference of $1$ day is identical to our previously determined best performing timeframe because it yielded the best results in the earlier tests using isolated posts.
Tests with different window sizes could not be conducted due to the limited scope of this thesis.

\begin{figure}[t]
  \centering
  \includegraphics[width=0.7\textwidth]{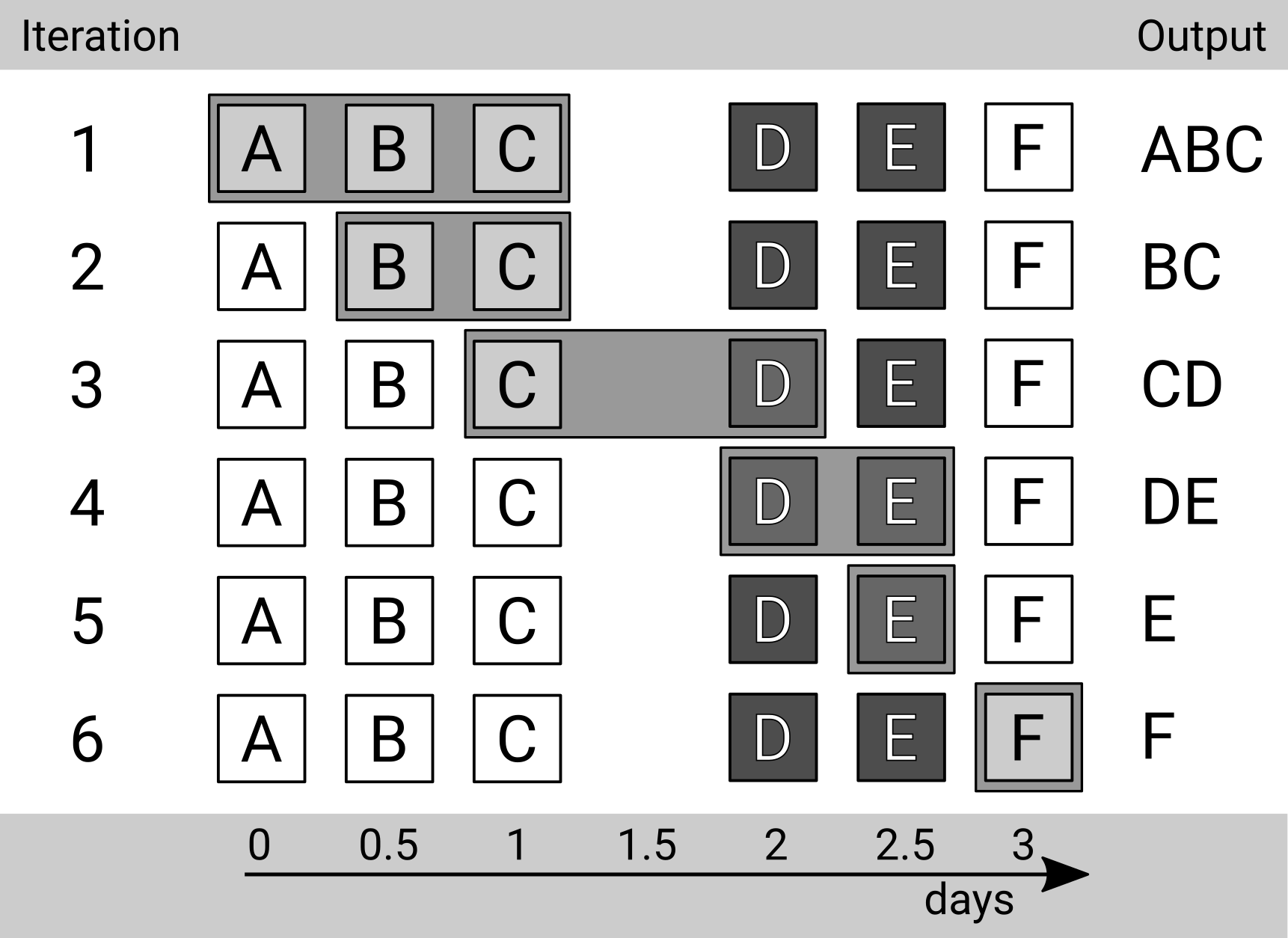}
  \caption{The graphic shows the sliding window algorithm throughout six iterations together with the merged posts that it outputs.
  Posts are represented by a square with the letter symbolising its content.
  All posts are assumed to be authored by the same editor.
  The grey rectangle depicts the current window.
  A disruptive post is indicated as dark grey square with its content in white font.
  The window size and timeframe are both set to be $1$ day.}
\label{fig:slidingWindow}
\end{figure}
The process of merging an editor's posts into a short history of their posts is depicted in \autoref{fig:slidingWindow} and can be described as follows:
All posts are separated by editor and sorted chronologically, starting with the oldest.
The first post of an editor is grouped together with all posts which they authored within $1$ day afterwards in a single window.
Thus, the time difference between the oldest and newest post in this window is not greater than $1$ day.
They are then merged keeping their chronological order.
Subsequently, the window is shifted to start at the second oldest post and the process is repeated until all posts are processed.
Therefore, single posts may appear in multiple merged posts.
As a result, the average post becomes longer while the total number of constructive and disruptive posts in test and training data remains the same.
In the example given in \autoref{fig:slidingWindow}, the post with content \enquote{\texttt{C}} will appear three times in the final data set.

If a post was made within the timeframe of $1$ day before the author was blocked, it is assumed to have led to this block.
In the example, a block must have happened at or after $2.5$ days but before $3$ days.
Every window containing such a post will also be regarded as having led to a block.
Regarding the example, the merged posts with contents \enquote{\texttt{CD}}, \enquote{\texttt{DE}} and \enquote{\texttt{E}} will be considered as disruptive.
Therefore, the number of posts identified to be disruptive may increase and the number of constructive ones may decrease analogously.
The original data set in \autoref{fig:slidingWindow} contained two disruptive posts and the one created by our sliding window approach contains three.
Likewise, the original annotated data set created from the Wikipedia dumps contained $179$ times more constructive than disruptive posts.
The data set built with the sliding window approach reduces this, so that there are approximately $162$ times more constructive than disruptive posts.
Thus, there are still sufficient constructive posts for creating a balanced data set.
To have the results somewhat comparable, the same sample size was chosen to be identical to the independent posts approach.

Assuming that blocking has an educational effect, editors might change their behaviour and make only constructive posts afterwards.
The algorithm respects this potential change in tone:
A window will only slide over a disruptive post when the chronologically next post belongs to the same block.
Hence, if a merged post is considered disruptive, at least its chronologically newest post must have been disruptive as well.
If the window cannot grow anymore due to a disruptive post being followed by a constructive one or a disruptive post from another block, the algorithm will continue as if this disruptive post was the last post within the $1$ day timeframe.
Therefore, increasingly shorter windows will be created until the window only spans the last disruptive post.
\autoref{fig:slidingWindow} illustrates this behaviour in iterations $3$ to $5$.
Afterwards, a new window will be constructed starting at the first post chronologically following this disruptive post.
In the example given, this is the post with content \enquote{\texttt{F}}.

Just like the independent posts tests, the sliding window approach was evaluated once using full text classifications and a second time using function word classifications.
All tests were evaluated using the metrics presented in \autoref{s:metrics}.

\section{Model Validation}\label{s:validation}
All classifiers are learnt and tested using a $10$-fold cross-validation.
In this validation, the data is being partitioned into ten parts.
The classifier is learnt on nine of them and tested on the left out tenth.
Thus, training and evaluation data are separate.
The process is repeated ten times, so that every partition has been the test corpus once.

We implemented the cross-validation of the LM classifier so that the set of constructive posts and the set of disruptive posts were linearly partitioned.
Thereby, we ensured that the data sets which the classifier was learnt and tested on were always balanced.
Regarding the SVM and NB classifier, we used stratified sampling for the independent posts approach.
This was done to ensure that the training and testing data was balanced in every cross-validation iteration.
Different to linear sampling, the data is randomly partitioned while ensuring a balanced class distribution.
Therefore, the posts used for training and testing of all three classifiers are identical but the partitioning differs between the LM classifier and the two other classifiers implemented in RapidMiner.
RapidMiner's implementation of linear sampling is less suitable for the input data as it would result in a class imbalance for every partition.

However, tests using the sliding window approach were performed using linear sampling for the SVM and NB classifiers.
Partitions featuring unbalanced class distributions had to be accepted because the sliding window approach allows posts to appear in multiple merged posts.
As stratified sampling creates partitions from randomly selected data, merged posts could appear in the test set that contain posts which were already in the training set.
Problems related to the differences in cross-validation are discussed in \autoref{s:err_results}.

\chapter{Results}\label{s:results}
In this \namecref{s:results}, the results of our classifiers throughout the various tests are shown.
Concrete performance values can be found in \autoref{app:results}.
For an easier visualisation, the AUC values have been transformed to percentages in all following charts.
As decided upon earlier, posts made up to $1$ day before a block of their author were considered disruptive for all tests.
For less cluttered charts, the names of the metrics have been abbreviated.
A plus symbol ($+$) indicates a performance metric calculated for disruptive and a minus symbol ($-$) for constructive posts, e.g.\ positive recall will be labelled \enquote{Recall$+$}.

The first two sections present the results of the independent posts and the sliding window approach.
After that, the results of applying the independent posts approach to the oldest \glspl{discussion} are shown.
This data led to eminently improved results, different to all other tested data sets.
For all classifications, the differences in performance of full text and function words classification are illustrated.
Moreover, terms which are characteristic for disruptive and constructive posts are discussed.

\section{Independent Posts Approach}\label{s:results_independent}
\begin{figure}[ht!]
  \centering
  \includegraphics[width=0.7\textwidth]{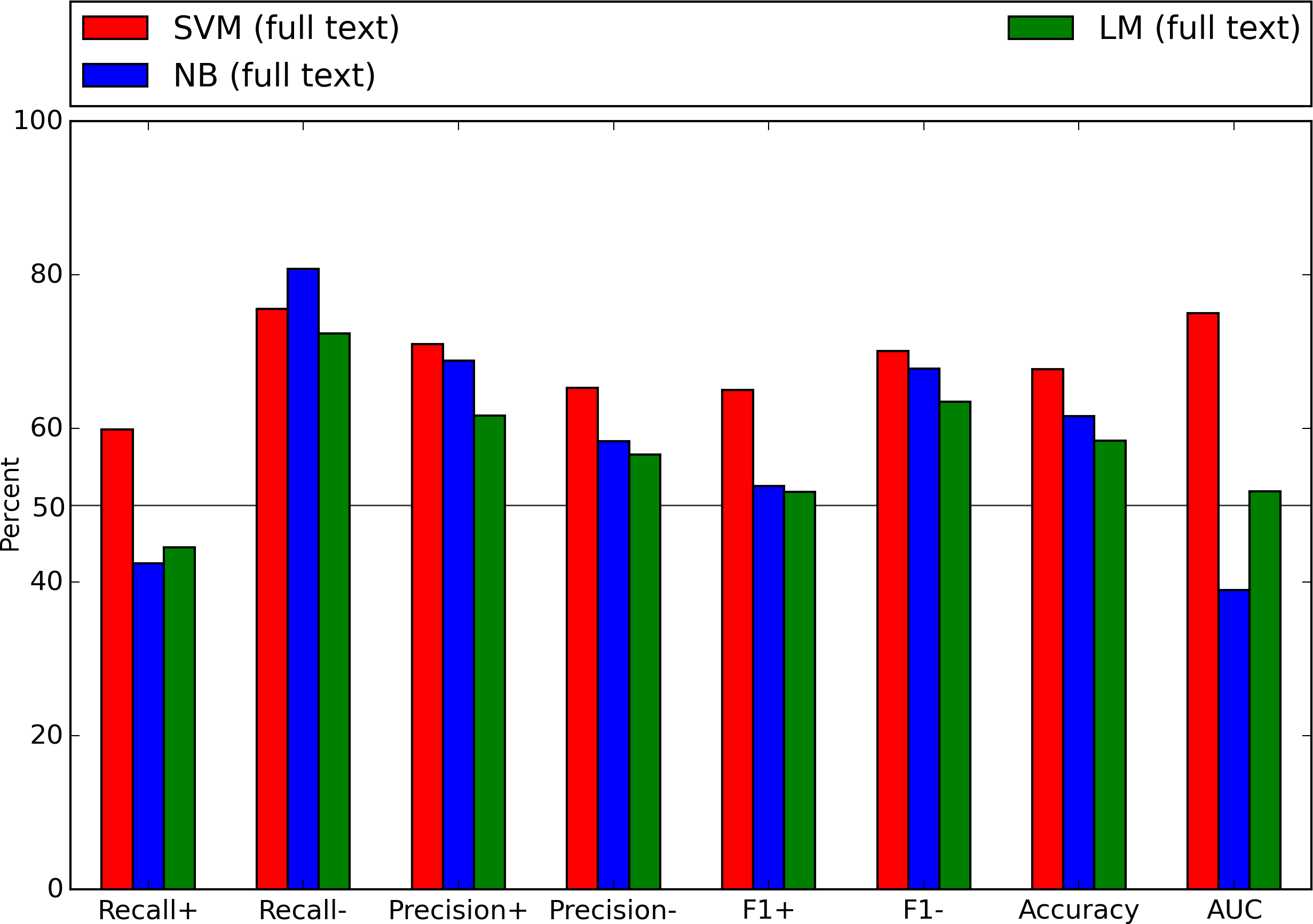}
  \caption{The performance of the full text SVM, NB and LM classifiers using the independent posts approach.}
\label{fig:1d_all}
\end{figure}

All values used to generate the charts in this \namecref{s:results_independent} are given in \Cref{t:val_1d_ip}.
\autoref{fig:1d_all} shows the performance of the three classifiers considering all words.
Overall, the support vector machine performs best, the na\"{\i}ve Bayes classifier ranks second and the language model ranks last.
The SVM is outperformed only in negative recall by the NB classifier.
However, due to lower negative precision, the NB classifier results in a worse negative F$1$ score than the SVM\@.
The SVM is significantly better in positive recall and AUC, whereas the NB classifier and the LM classifier repeatedly perform similarly.
When looking at the positive and negative F$1$ scores, it can be seen that all classifiers are better in predicting constructive posts than disruptive ones.
It was expected that the SVM would outperform the NB classifier.
Yet, we imagined that the language model would perform better.
All in all, the three classifiers perform hardly well enough to draw reliable conclusions from the results.
F$1$, AUC and accuracy scores of about $80$\% and above would be needed.

\begin{figure}
  \centering
  \includegraphics[width=0.7\textwidth]{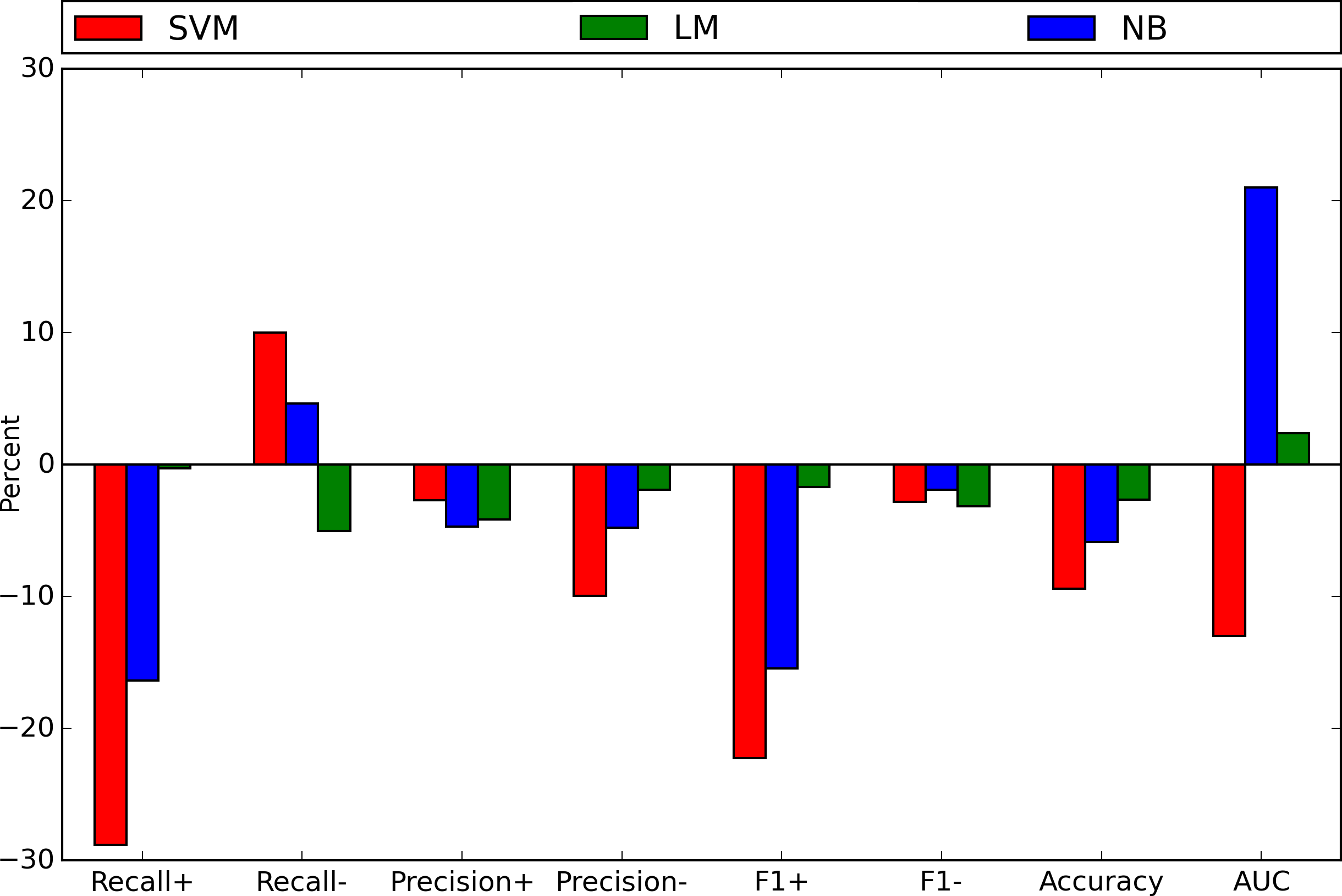}
  \caption{This figure shows the difference in performance between the function words classifiers and their full text equivalents.
  Positive percentages show a performance improvement over the full text classifier and negative ones a decline.
  For example, if a classifier considering all words achieved $50$\% AUC and that considering function words $70$\%, the function words classifier performed $20$\% better overall.
  }
\label{fig:1d_fw_all_comparison}
\end{figure}
\begin{figure}
  \centering
  \includegraphics[width=0.7\textwidth]{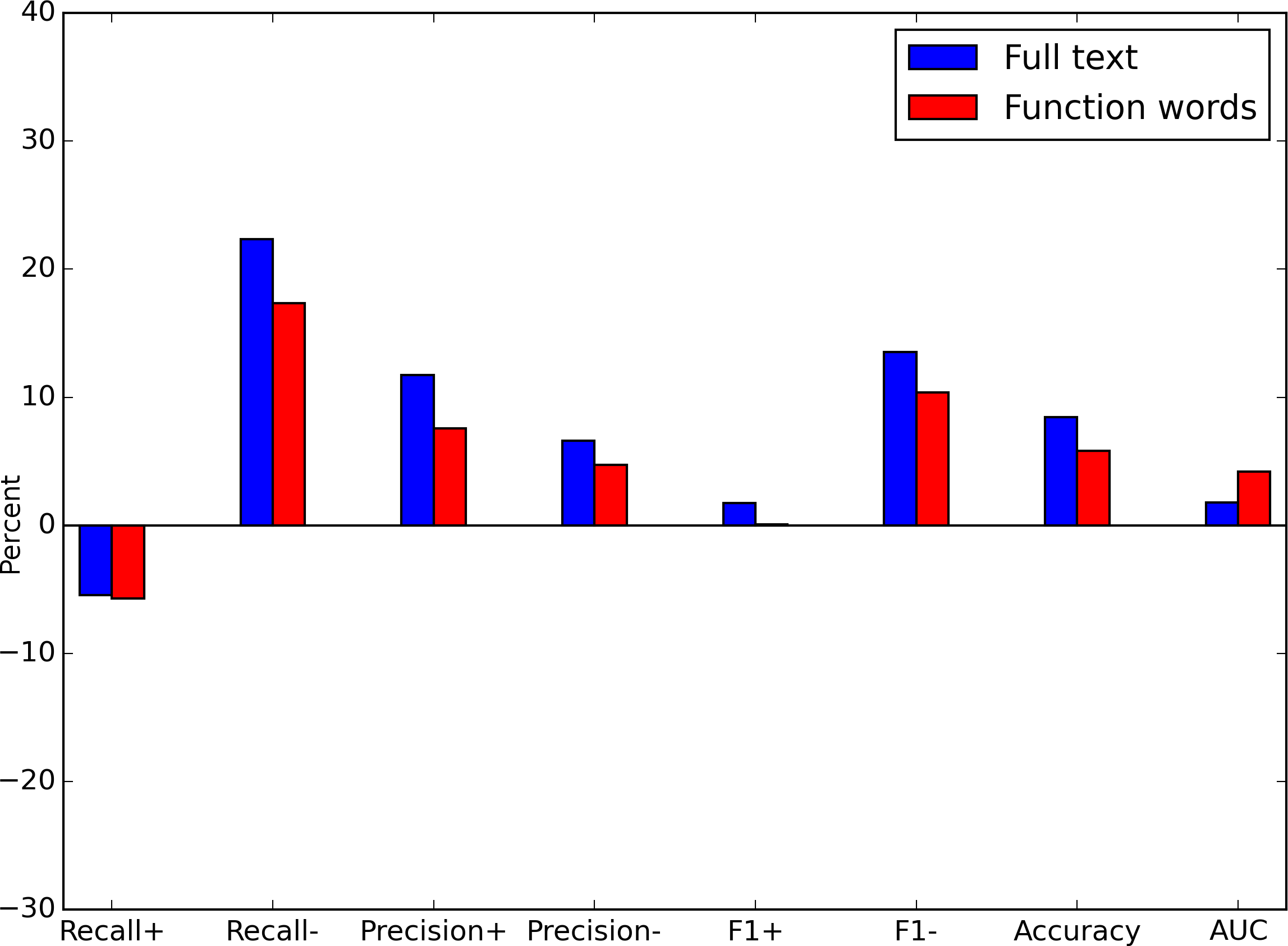}
  \caption{This bar chart shows performance metrics in relation to the performance of a random classifier with $p=0.5$ for both classes.
  Depicted is the full text LM classifier in contrast to the function words LM classifier.
  Values below $0$\% indicate a performance worse than a random classifier, whereas values above $0$\% mark a better performance.
  The percentages refer to the overall performance, meaning that $50$\% equates to a perfect result.
  }
\label{fig:1d_lm}
\end{figure}

The full text classifiers perform significantly better than their function words equivalents.
Except for negative recall of the SVM and LM classifier as well as the AUC for the NB and LM classifiers, all metrics indicate a worse performance when solely considering function words as visible in \autoref{fig:1d_fw_all_comparison}.
A decline was expected because the amount of information in a post was massively reduced.
Around $59.56$\% of words in a post have been ignored in the function words classifications.
The SVM's performance changes the most whereas the LM classifier's performance remains similar.
Nevertheless, the LM classifiers' performances cannot be used to answer \RQ{3} regarding the importance of function words.
The reason for this is that the full text classification's performance is already too similar to that of a random binary classifier that alternates its predictions with $p=0.5$ independent of any input.
This is visualised in \autoref{fig:1d_lm}, which shows the performance differences between the full text and function words LM classifier side by side in relation to such a random classifier.
The function words classifier achieved an accuracy of merely $55.73$\%.

\begin{figure}
  \centering
  \includegraphics[width=\textwidth]{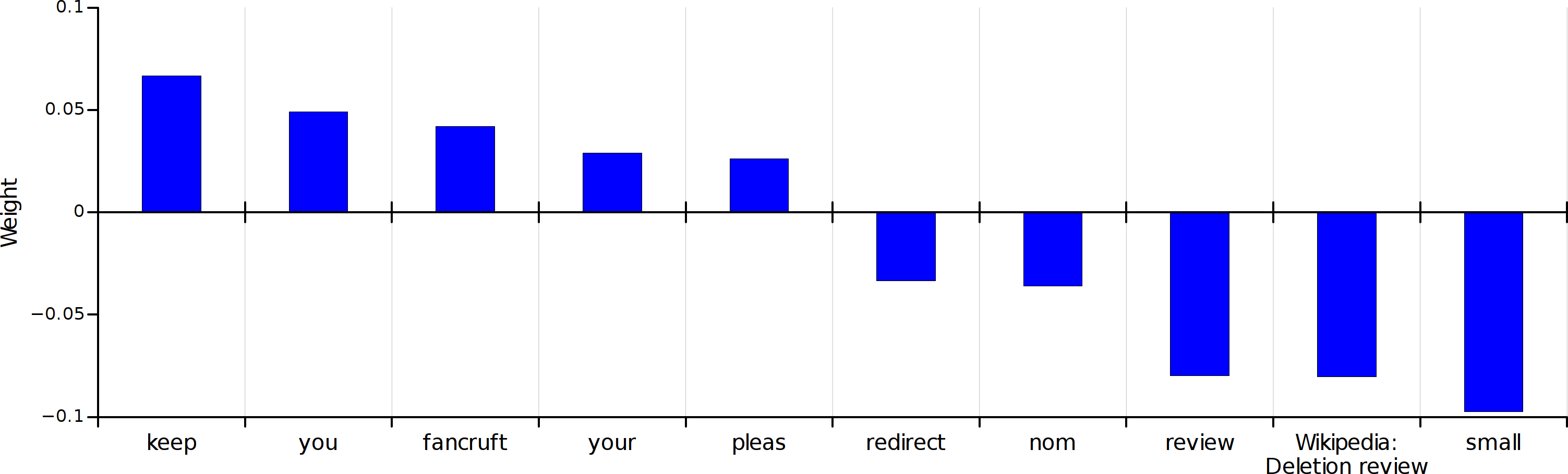}
  \caption{The top five lowest and highest weighted stemmed terms of the full text SVM\@.
  Positive weighted terms are characteristic for disruptive and negative weighted for constructive posts.
The penultimate word was formatted to contain a colon and a space for better readability.}
\label{fig:1d_words}
\end{figure}
\autoref{fig:1d_words} contains the five most characteristic terms for constructive and disruptive posts each.
The terms were retrieved by extracting the highest and lowest weighted terms according to the full text SVM as it performed the best.
Seven out of these ten terms are specific to Wikipedia and \glspl{discussion}.
This indicates that the SVM's results are probably not representative for textual discussions in general.
\enquote{Fancruft} is a Wikipedia term for contents that are only of interest for a small number of avid fans~\citeW{Wikipedia:Fancruft}.
\enquote{Nom} is short for \enquote{nomination} and often used in posts such as \enquote{\texttt{delete as per nom}} to express agreement towards the deletion nomination.
\enquote{Wikipedia:Deletion review} is detected as a single term due to our preprocessing step that converts internal Wikipedia links appropriately.
Deletion reviews offer Wikipedia users to appeal article deletions if they believe that the deletion was unrightful by Wikipedia standards~\citeW{Wikipedia:DeletionReview}.
\enquote{Small} often appears as an HTML element to markup contents in a smaller font size.
At this point, it is unclear why many of these elements have not been removed in the data preprocessing steps.
\enquote{Redirect} and \enquote{keep} are both potential outcomes for the discussed article.

Editors recommending to \enquote{keep} an article express that in their opinion the article's quality and relevance are both sufficient and that the article should not be deleted.
One explanation for this term being ranked the highest could be that it may be used in contexts where an editor was involved in creating the article and will subjectively defend it.
Some editors even create new accounts to act as different editors who argue to keep the article as well.
Such accounts are called \emph{sock puppets} on Wikipedia and are highly likely to be blocked once detected~\citeW{Wikipedia:SockPuppetry}.
Besides, manual inspection indicated that disruptive posts containing \enquote{keep} often also contained personal attacks.
The SVM seems to have optimised for a frequently appearing term, with it being contained by $22.75$\% of all disruptive and $19.90$\% of all constructive posts in this data set. 
On the other hand, the highest weighted swear word \enquote{fuck} appears in $0.77$\% to $0.05$\% of the posts respectively.
In relation to each other, \enquote{fuck} appears much likelier in disruptive than in constructive posts than \enquote{keep} does.
Yet, it was only ranked $17$th place.

The results support \RQ{1} in that \yms{} seem to be commonly used in disruptive posts.
\enquote{You} and \enquote{your} both appear in the top five of most characteristic terms for disruptive posts.
\I{} ranks $62$nd.
For comparison, \I{} is the $36,201$st heaviest weighted term identifying constructive posts.
Regarding these results, \I{} is thus not typical for constructive posts which was speculated by \RQ{2}.

\begin{figure}
  \centering
  \includegraphics[width=0.7\textwidth]{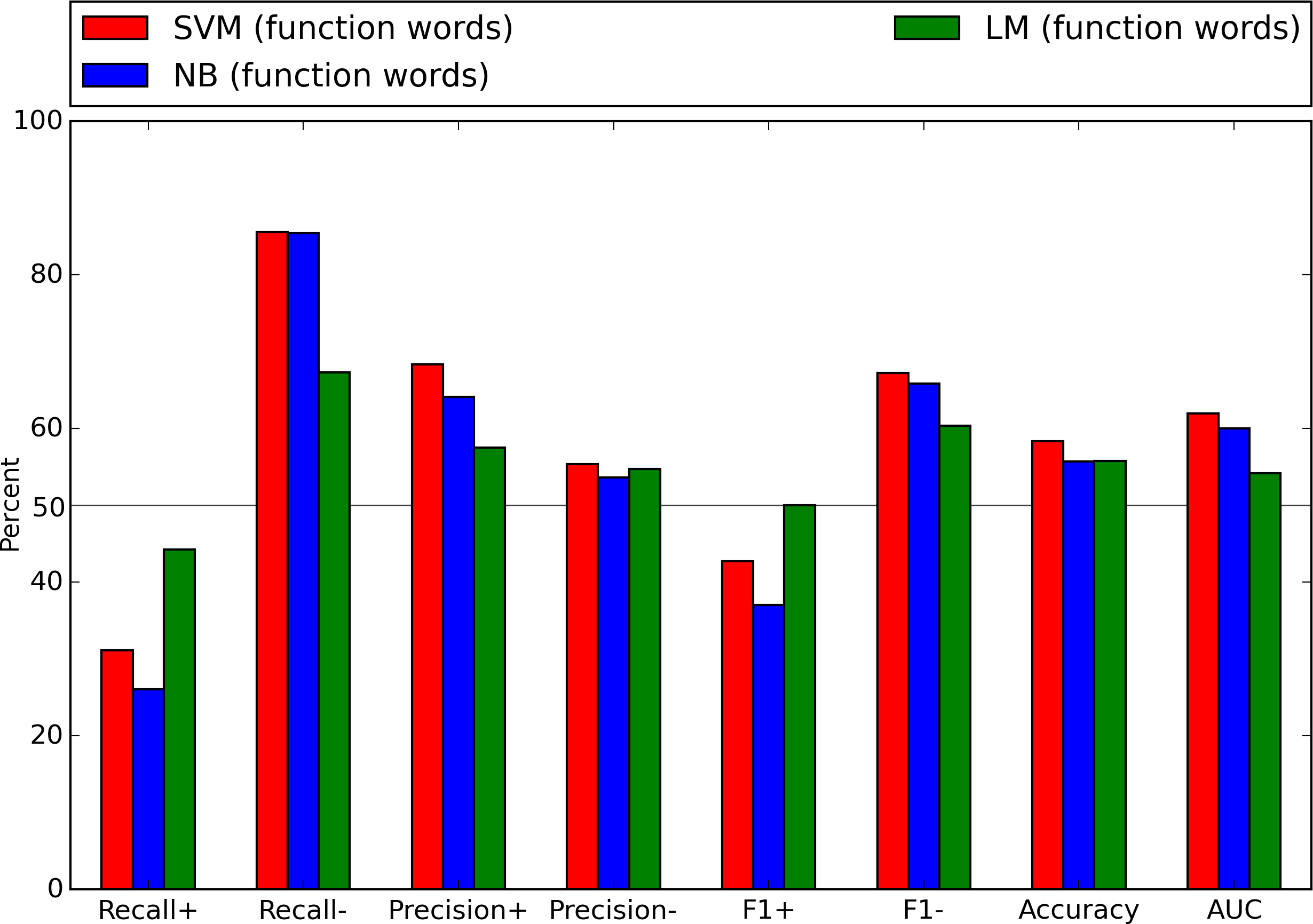}
  \caption{The performance of the SVM, NB and LM classifiers using the independent posts approach while only respecting function words.}
\label{fig:1d_fw}
\end{figure}
\begin{figure}
  \centering
  \includegraphics[width=\textwidth]{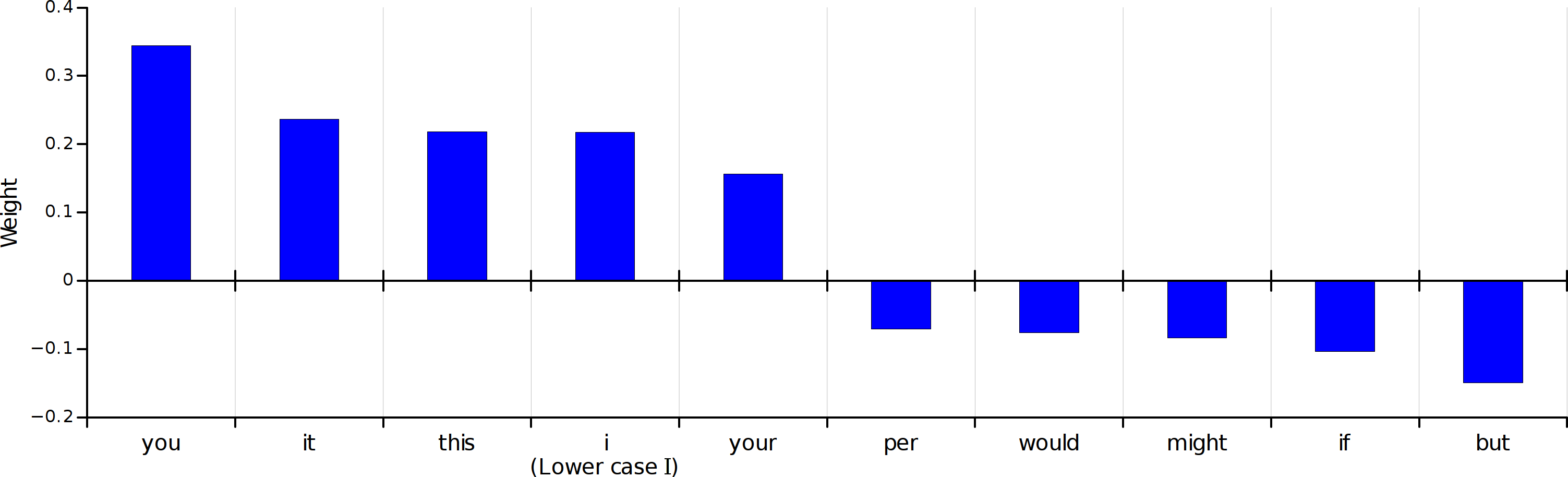}
  \caption{The top five lowest and highest weighted stemmed terms of the function words SVM\@.
  Positive weighted terms are characteristic for disruptive and negative weighted for constructive posts.}
\label{fig:1d_words_fw}
\end{figure}
As can be seen in \autoref{fig:1d_fw}, the SVM yielded the best results when considering only function words.
Consequently, we inspected the ten heaviest weighted stemmed terms.
They are shown in \autoref{fig:1d_words_fw}.
It is difficult to speculate why the function words have been weighted as they are.
However, \enquote{you} and \enquote{your} both being part of the top five terms most characteristic for disruptive posts in this data set indicates support for \RQ{1}.
The two terms are common parts of  \yms{}.
Moreover, \enquote{you} is by far the most characteristic term for disruptive posts among all considered $162$ function words.
In contrast, \RQ{2} questioned whether \ims{} would typically be used in constructive messages but \I{} was the fourth highest positively weighted term.
Hence, \RQ{2} seems to be negated.

Nonetheless, it is important to take account of the performances when drawing conclusions from the observed results.
The function words SVM achieved merely $58.34$\% accuracy.
Thus, \RQ{1} being supported and \RQ{2} being negated should be interpreted as a tendency.
Also, the nature of \glspl{discussion} might generally favour objective posts and thus bias how \iyms{} are perceived.
With an accuracy of only $67.73$\%, the SVM performs significantly better than a random classifier.
Yet, its results are not good enough for answering the research questions with certainty.
Thus, although \enquote{you} and \enquote{your} are in the top five of terms characteristic for disruptive posts, \RQ{1} cannot be safely answered.
The same holds true for \I{} being identified as typical term contained in disruptive posts, which would negate \RQ{2}.
However, both amplify the trends seen in the function words classification.
We found no support for function words being sufficient for making good predictions about whether a message is constructive or disruptive as was asked by \RQ{3}.
\RQ{4} addressed whether there are other characteristic terms.
No answer can be given that would be applicable to general textual discussions as other terms identified as being characteristic for either of the classes were related to Wikipedia and \glspl{discussion}.

\section{Sliding Window Approach}\label{s:results_sw}
Classifying posts independent of each other implies that only the posts within the $1$ day timeframe were of disruptive nature.
This might not always be the case and the results of that approach were not satisfying either.
Thus, the sliding window approach was tested as well.
It increased the number of posts in our data set that were assumed to be disruptive.

The number of constructive and disruptive posts was increased compared to the independent posts classifications.
This is conditioned by the sliding window algorithm which regards all merged posts containing a blocked post as disruptive.
As this approach considers an editor's post history, the data is different and the approach was evaluated on newly sampled data.
However, multiple runs of classifications using both the independent posts as well as the sliding window approach showed that the results always remained similar.
All values used to generate the charts in this \namecref{s:results_sw} are given in \Cref{t:val_1d_sw_lin}.

\begin{figure}[h!]
  \centering
  \includegraphics[width=0.7\textwidth]{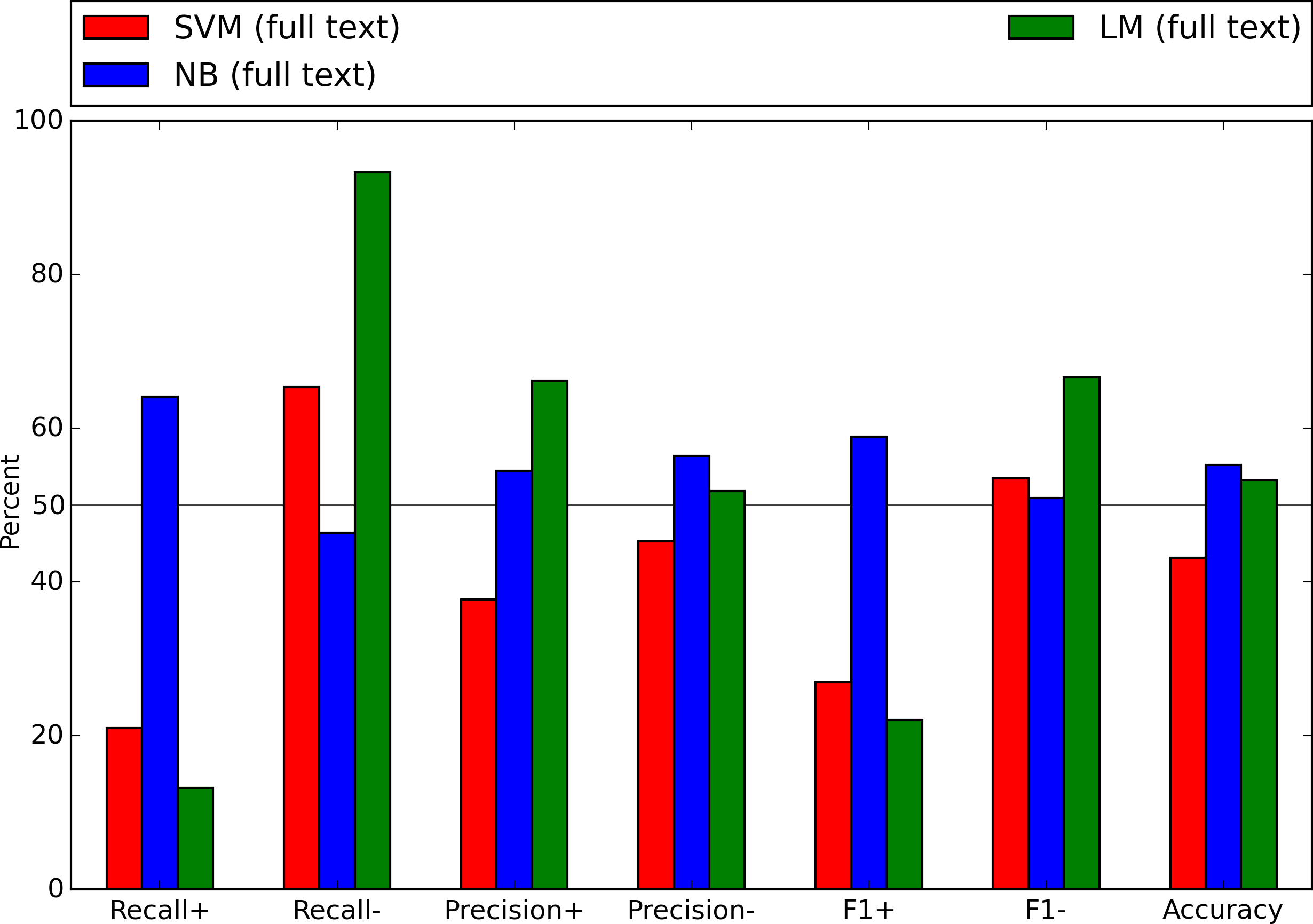}
  \caption{The performance of the full text SVM, NB and LM classifiers using our sliding window approach.}
\label{fig:1d_sw_all_lin}
\end{figure}
\autoref{fig:1d_sw_all_lin} shows the performance of the classifiers using the sliding window approach.
As RapidMiner was unable to calculate the AUC, it had to be left out.
When inspecting the accuracy, it becomes clear that the performance of all three classifiers has worsened.
The SVM was the formerly best performing classifier but has turned into the worst performing.
Again, the NB classifier performs slightly better than the LM classifier.
With the independent posts approach, the LM classifier was significantly better in predicting constructive posts as expressed by the F$1$ scores.
This tendency was amplified in the sliding window approach as its positive F$1$ score is now $22.03$\% and its negative $66.61$\%.

\begin{figure}[t]
  \centering
  \includegraphics[width=0.7\textwidth]{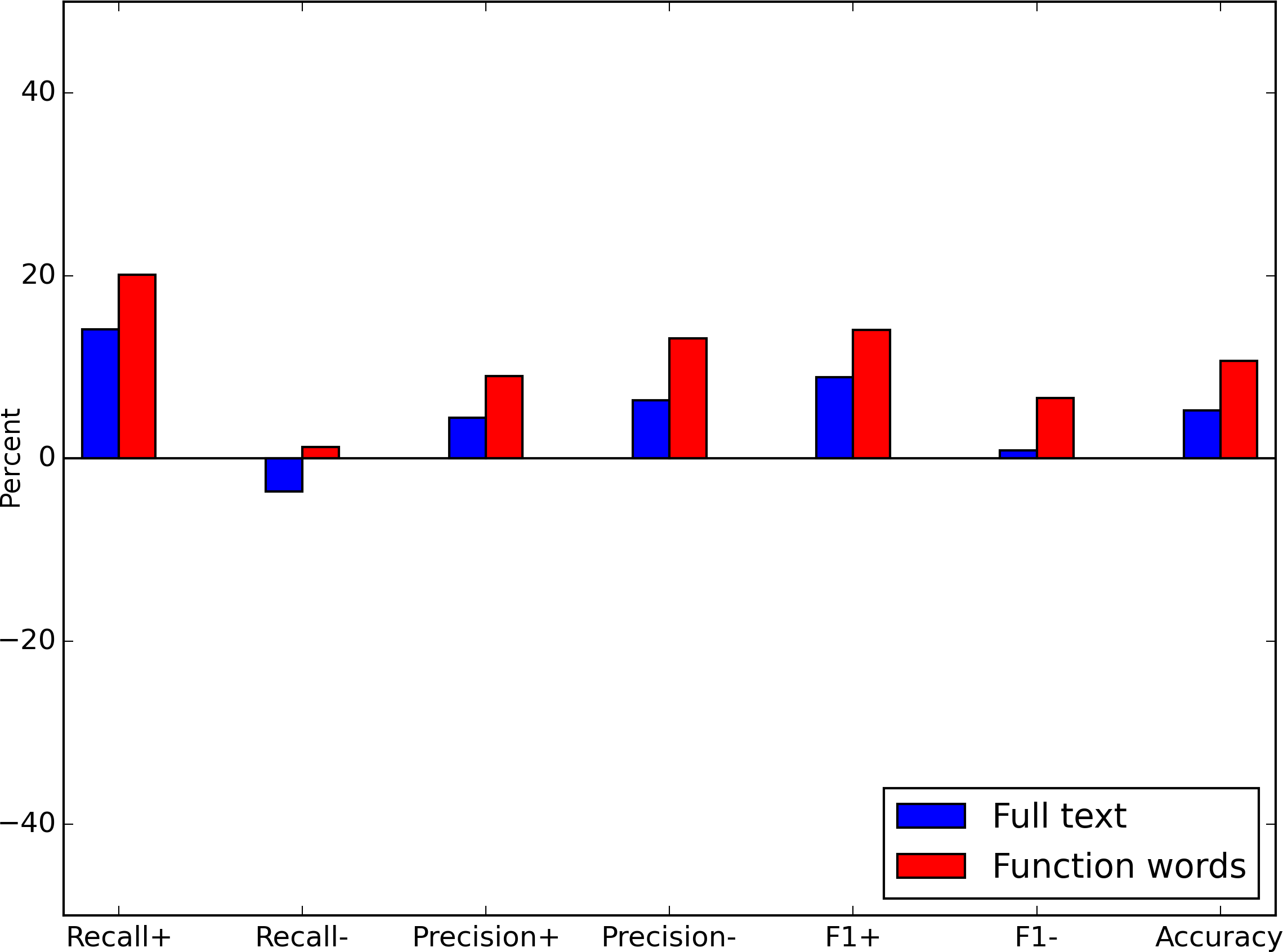}
  \caption{This bar chart shows the performance of the full text NB classifier in contrast to it only factoring in function words when using the sliding window approach.
  It is set in comparison to the performance of a random classifier with $p=0.5$.
  $0$\% expresses equal performance.
  The percentages refer to the overall performance, meaning that $50$\% equates to a perfect result.
  }
\label{fig:1d_sw_nb_lin}
\end{figure}
As all classifiers performed similar to or worse than a random classifier, the difference in performance between the full text and function words classification is less distinctive.
Surprisingly, the NB classifier forms an exception as all its values improved.
\autoref{fig:1d_sw_nb_lin} shows this improvement.
With an accuracy of $60.70$\%, it performed only slightly worse than the full text NB classifier using the independent posts approach which achieved $61.60$\%.
With a positive F$1$ score of $64.08$\% and a negative of $56.61$\%, it predicted disruptive better than the full text NB classifier of the first approach.
The latter had positive and negative F1 scores of $52.49$\% and $67.78$\% respectively.

\begin{figure}
  \centering
  \includegraphics[width=0.6\textwidth]{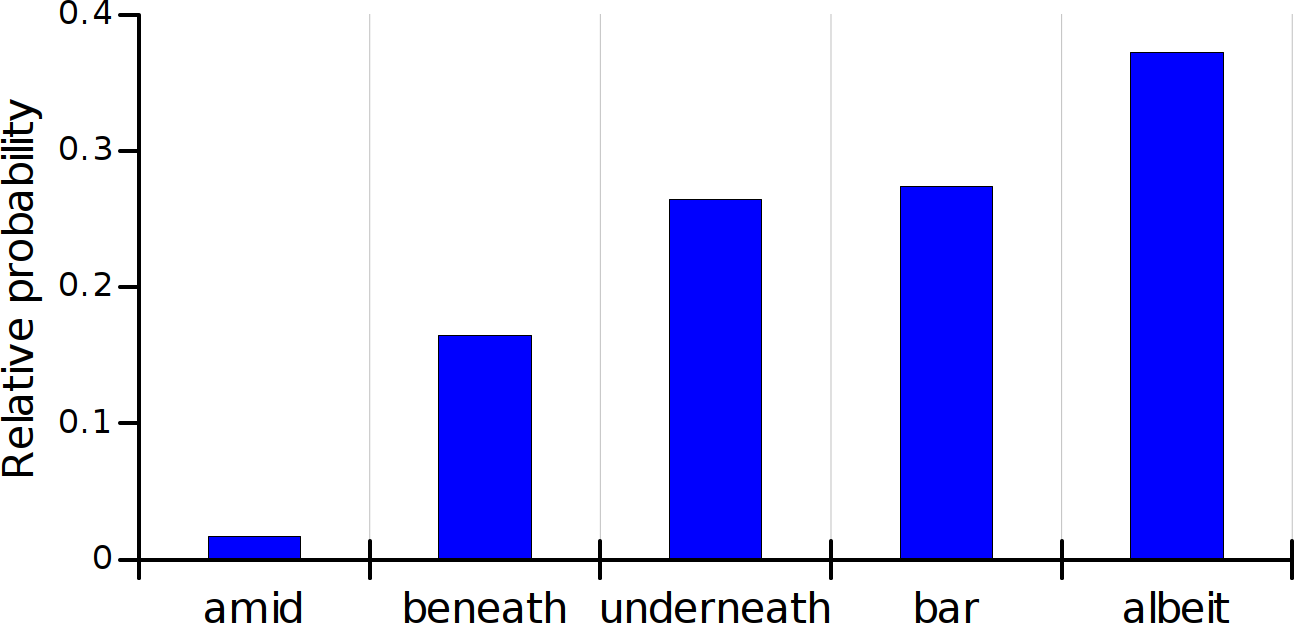}
  \caption{This chart shows a relation of positive probability divided by negative probability for a stemmed term.
  A low relational probability indicates a term characteristic for constructive posts.
  Shown are the terms most characteristic for constructive posts according to the results of the function words NB classifier using the sliding window approach.
  }
\label{fig:1d_nb_sw_constructive_words}
\end{figure}

\begin{figure}
  \centering
  \includegraphics[width=\textwidth]{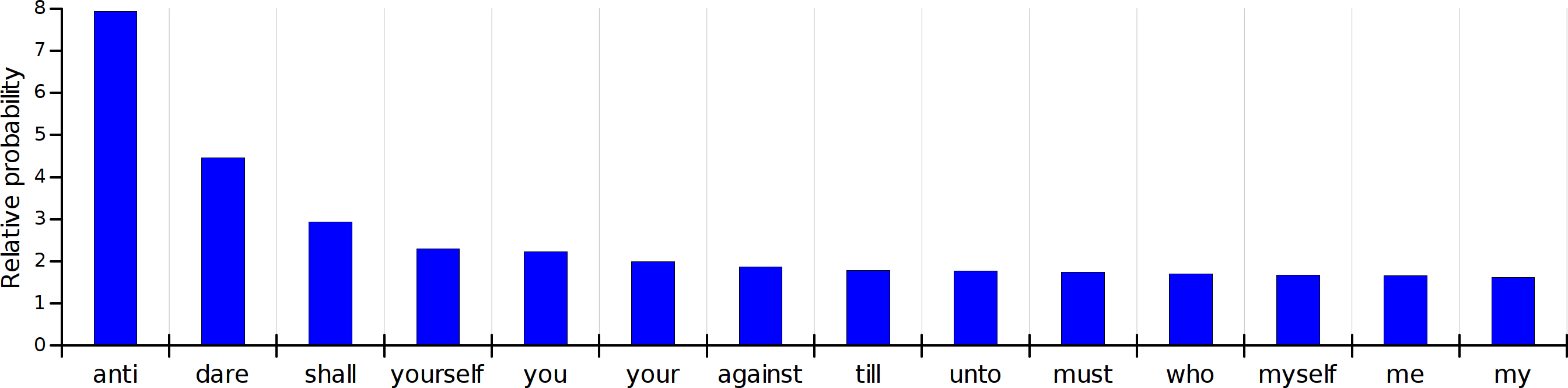}
  \caption{This chart shows a relation of positive probability divided by negative probability for a stemmed term.
  A high relational probability indicates a term characteristic for disruptive posts.
  Shown are the terms most characteristic for disruptive posts according to the results of the function words NB classifier using the sliding window approach.
  }
\label{fig:1d_nb_sw_disruptive_words}
\end{figure}

Consequently, we extracted the terms most characteristic for constructive and disruptive posts from the function words NB classifier.
\autoref{fig:1d_nb_sw_constructive_words} shows the five function words most typical for constructive posts.
All these terms have in common that they appear infrequently in the analysed posts.
We decided to only depict five terms as the following appear similarly meaningless.
On the other hand, the terms indicating disruptive posts are quite expressive.
They are given in \autoref{fig:1d_nb_sw_disruptive_words}.
Inspecting the data showed that \enquote{anti} was mostly used in contexts of strong disagreement.
Moreover, it was used in political and religious topics such as \enquote{anti-Serbian}, \enquote{anti-Muslim} and \enquote{anti-Semitic}.
These topics are known to be prone to controversial discussions on Wikipedia~\citep{yasseri_most_2013}, which themselves often lead to blocks.
\enquote{Against} and \enquote{anti} have synonymous meanings and suggest that disagreement and disruptive posts are potentially related.
As was asked in \RQ{1}, \yms{} seem to be an indicator for disruptive posts.
\enquote{Yourself}, \enquote{you} and \enquote{your} are all likely to be used in \yms{} and belong to the six function words most characteristic for disruptive posts.
Similarly, \enquote{myself}, \enquote{me} and \enquote{my} are likely part of \ims{}.
Them being identified to be among the strongest indicators for disruptive posts suggests that \RQ{2} must be negated.
\I{} appears as $38$th most characteristic term for disruptive posts.
In total, $162$ function words were considered.
Regarding the terms from \autoref{fig:1d_nb_sw_constructive_words}, it seemed as if their occurrence frequency substantially implied their ranking.
This, however, is not true as becomes clear when inspecting the occurrence frequencies of \enquote{dare} and \enquote{you} for example.
\enquote{dare}, ranking second most characteristic term for disruptive posts, appears merely $93$ times in the analysed posts. 
Conversely, \Y{} appears $49,997$ times and ranks fifth most characteristic term for disruptive posts. 

In summary, we saw the sliding window approach perform worse than the independent posts approach.
Due to the full text classifiers performing similar to a random classifier, a presentation of characteristic stemmed terms for disruptive and constructive posts was skipped.
Despite the low accuracy of $60.70$\%, the NB classifier yielded better results when considering function words exclusively than when considering all words in a post.
We count this as relatively strong support for \RQ{3} that function words can indeed play an important role in distinguishing constructive and disruptive posts.
The observed effects of \iyms{} in the independent posts approach were also visible in the function words NB classification.
Therefore, our presumptions are reinforced that \RQ{1} could be affirmed and \RQ{2} negated.
As asked in \RQ{4}, the list of most influential function words for disruptive posts hinted that terms expressing opposition could be an indicator for determining whether a post is disruptive.

\section{Analysis of the Oldest Articles for Deletion Discussions}\label{s:results_chrono}
Although we only present the data of a single run each, all tests have been executed multiple times with newly sampled data sets.
This confirmed that the observed performances were not the result of a sampling bias.
The results always remained comparable but for one exception.
When operating on a data set chronologically sampled from the earliest \glspl{discussion}, we found that the classifiers yielded significantly improved performance.
Chronologically sampling data from more recent \glspl{discussion} could not reproduce these results.
Instead, newer chronologically sampled data was comparable to the performance of our randomly sampled classifications as presented in the previous \Cref{s:results_independent,s:results_sw}.
However, due to the amount of existing data, we were not able to test all possible partitions.
All values used to generate the charts in this \namecref{s:results_chrono} are given in \Cref{t:val_1d_ip_chrono}.

\begin{figure}[hb]
  \centering
  \includegraphics[width=\textwidth]{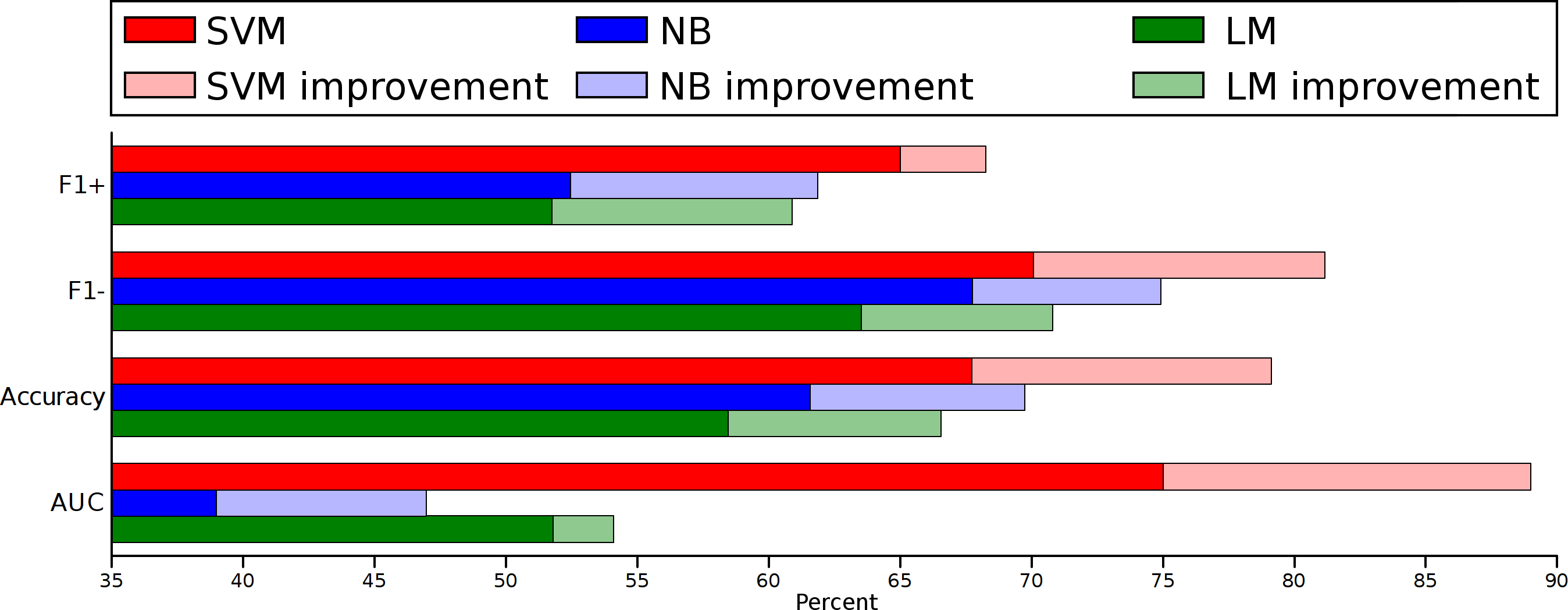}
  \caption{The darker coloured bars show the performance of the independent posts approach using the full text classifiers taken from \autoref{fig:1d_all}.
  When the analysed data is chronologically sampled from the oldest posts, the performance improves as indicated by the lighter coloured bars.
  To prevent the chart from becoming unclear, precision and recall have been left out.
  }
\label{fig:1d_chrono}
\end{figure}

As before, all available disruptive posts were considered.
Instead of randomly sampling the same number $n$ of constructive posts like in earlier approaches, we used the first $n$ constructive posts in chronological order starting with the oldest.
The performance of all classifiers improved using the independent posts approach, cf.\ \autoref{fig:1d_chrono}.

\vspace{1cm}

\begin{figure}
  \centering
  \includegraphics[width=\textwidth]{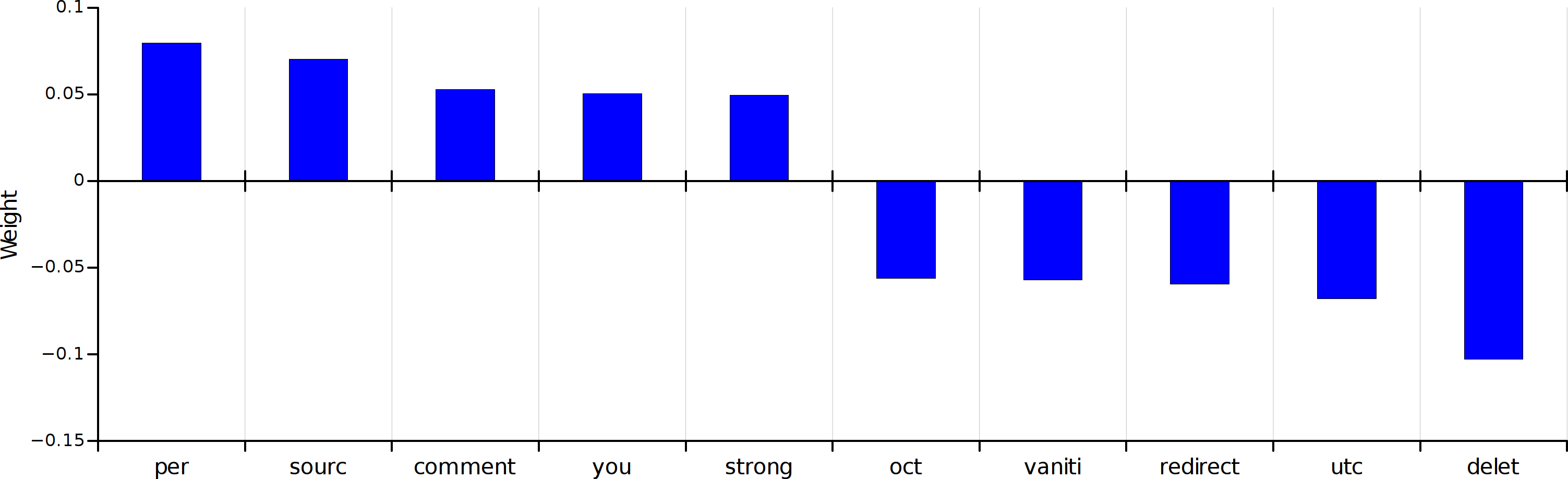}
  \caption{The top five lowest and highest weighted stemmed terms of the full text SVM using the independent posts approach on the chronologically first $19,237$ posts.
  Positive weighted terms are characteristic for disruptive and negative weighted for constructive posts.}
\label{fig:1d_words_chrono}
\end{figure}

Similar to the results observed when using randomly sampled data, the SVM performed best and the language model worst.
The SVM using the independent posts approach scored an accuracy of $79.12$\% and an AUC of $0.890$.
Moreover, it achieved a positive precision of $87.20$\% and a negative of $73.93$\%.
Therefore, the highest and lowest weighted terms are likely to be characteristic for constructive and disruptive posts in the context of this special data set.
These words are shown in \autoref{fig:1d_words_chrono}.

Again, most words are specific to \glspl{discussion}.
\enquote{Per} is another term appearing in the frequently used phrase \enquote{\texttt{delete per nom}}.
\enquote{Sourc} is a word stem from words such as \enquote{source} and is commonly found in posts which highlight that the article in question is lacking trustworthy sources.
\enquote{Strong} is often used in combination with \enquote{keep} or \enquote{delete} by editors to express strong belief that the article under discussion should be kept or deleted.

\enquote{Oct} is short for October.
\enquote{Utc} is frequently found in Wikipedia signatures as abbreviation for coordinated universal time~\citep[cf.][]{Wikipedia:Signatures}.
Although \glspl{discussion} are imaginable in which editors talked about the month October or the coordinated universal time, it is likelier that both words are leftovers from customised signatures.
Manual data inspection confirms this assumption.
This would then indicate that our attempts to remove user signatures were insufficient for one or multiple users.
As a result, the classifier might not have been learnt on the used language itself but has learnt implicit clues that identify one or more disruptive or constructive users.
This could partly be a reason why the classifiers yielded notably better results with this data set.

\enquote{Vaniti} is a stemmed term derived from the word \enquote{vanity}.
On Wikipedia, vanity was a term to describe contributions that are not of general interest but are made for the purpose of self-promotion~\citeW{Wikipedia:Vain}.
The term was used $36,830$ times until autumn 2004, when it got replaced by the broader behavioural guideline \enquote{conflict of interest}~\citeW{Wikipedia:COI}.
This guideline covers the potential bias that contributions may contain when their author is in any way related to its content.
The term \enquote{vanity} never appeared again after 2004.

The heaviest weighted stemmed term indicating constructive posts is \enquote{delet}.
Derived from the verb \enquote{delete}, it is commonly used by editors to advise for deletion of an article.
It could be speculated that words stemmed to \enquote{delet} are mostly used in posts that neutrally express to remove an article.
Manual inspection supports this presumption as authors of disruptive posts typically advocate to keep an article.
However, \enquote{per} being the most characteristic stemmed term for disruptive posts contradicts this as it mostly appears in the phrase \enquote{\texttt{delete as per nom}}.
The term \enquote{you} is again in the top five of highest weighted words, which indicate disruptive communication.
Thus, the associated \RQ{1} is further supported.
The term \I{} is the $64$th most representative term for disruptive posts.
Considering that this data set contained $34,627$ terms, the term is clearly still not typical for constructive posts.
As a result thereof, \RQ{2} is again negated.

\begin{figure}[p]
  \centering
  \includegraphics[width=0.7\textwidth]{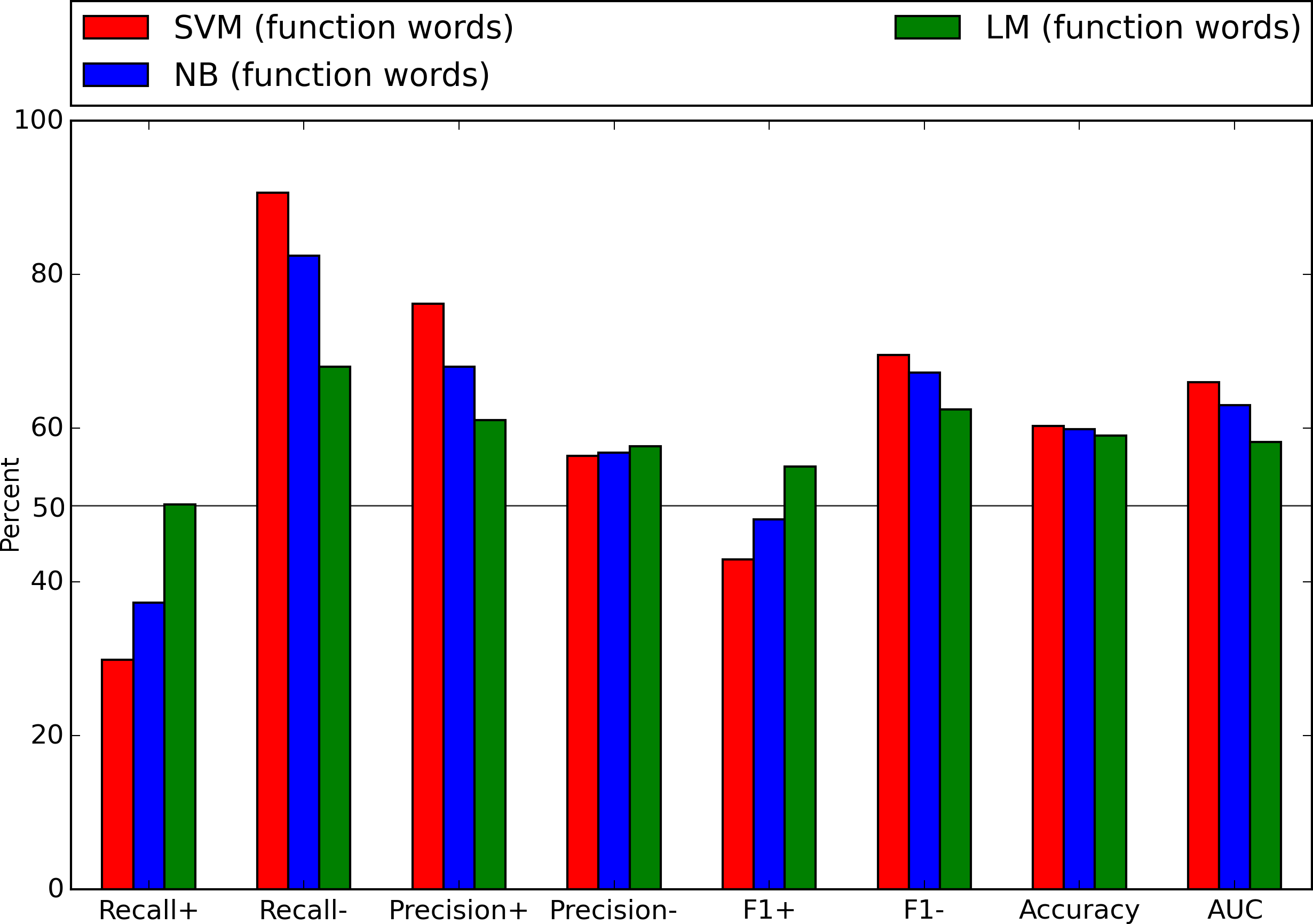}
  \caption{The performance of the function words SVM, NB and LM classifications using the independent posts approach on the chronologically first $19,237$ posts.
  }
\label{fig:1d_chrono_fw}
\end{figure}
\begin{figure}[p]
  \centering
  \includegraphics[width=\textwidth]{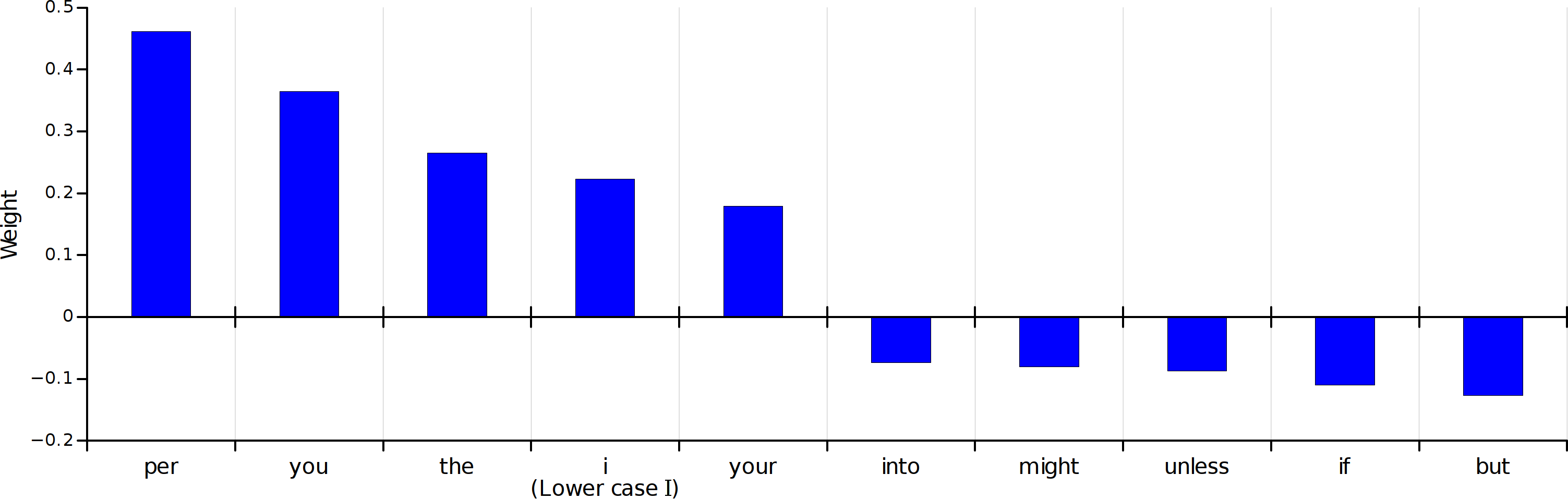}
  \caption{The top five lowest and highest weighted stemmed terms of the function words SVM using the independent posts approach on the chronologically first $19,237$ posts.
  Positive weighted terms are characteristic for disruptive and negative weighted terms for constructive posts.}
\label{fig:1d_words_chrono_fw}
\end{figure}
When only considering function words, the performance decreases significantly.
\autoref{fig:1d_chrono_fw} shows the performance of all classifiers.
It is slightly better than that of the independent posts approach with randomly sampled data.
The overall performance according to accuracy and AUC is again best for the SVM\@.
\autoref{fig:1d_words_chrono_fw} shows the heaviest weighted stemmed terms from the function words SVM\@.
Although the SVM has the lowest positive F$1$ score, it has the highest positive precision.
Therefore, it predicted only few disruptive posts but when it did, it was correct $76.21$\% of the time.
Interestingly, the absolute values of the positive weights is much higher than that of the negative weights.
\enquote{You}, \enquote{your} and \I{} appear in the five highest weighted terms indicating disruptive posts.
As before, this supports \yms{} being typically used in disruptive posts (\RQ{1}) but negates the use of \ims{} in constructive posts (\RQ{2}).

In essence, all classifiers perform notably better when analysing the oldest \glspl{discussion}.
This is probably due to terms that uniquely appear in this data set like \enquote{vanity} as well as not fully removed customised signatures.
The results suggest that the signatures belonged to highly active, constructively contributing editors.
Nevertheless, \enquote{you} has been identified as fourth highest weighted term indicating disruptive posts when considering all words.
Together with the term \enquote{your}, it appeared again among the highest weighted terms characteristic for disruptive posts when only considering function words.
\RQ{1} thus gained further support.
The term \I{} reappeared among the same terms.
Again, \RQ{2} would have to be negated.

\chapter{Design Decisions, Potential Errors and Improvements}\label{s:errors}
Multiple assumptions had to be made throughout this thesis, which could have noticeably influenced the results.
This \namecref{s:errors} focuses on these systematic errors and design decisions.
When possible, it is discussed how they could have influenced the results or how the associated errors could be resolved.
Although we put a lot of effort into all steps, the amount and extent of problems potentially impacting the classifications illustrate that the data is not well-suited for the analysis of textual discussions on a word-level.

\section{Difficulty of Building an Annotated Data Set}
A great difficulty was to build an annotated data set by determining which posts may have or have not led to a block.
This problem is rooted in the ambiguous information given about issued blocks.
Blocks are not associated with any contribution or page.
Thus, it is unclear which posts can be labelled disruptive and which ones constructive.
The block log even contains cases where the blocking was unrelated to any contribution.
For example, an administrator noted in a block log comment that receiving legal threats via email made him issue this very block. 

All posts within a fixed timeframe prior to their author being blocked were considered to be disruptive.
We determined the best performing timeframe which still contains a great amount of data.
However, this cannot ensure that all contributions within the timeframe prior to a block have actually made the administrator act.
For example, it is possible that an editor was simultaneously active in multiple \glspl{discussion} before being blocked.
In some \glspl{discussion}, they may have been contributing constructively, while they lost their temper in others.
Hence, the posts from all their active discussions shortly before their block would wrongfully be labelled disruptive although this is only true for some of them.
Our approach cannot detect this behaviour which reduces the correctness of our annotated data set.

Moreover, it is hard to determine relevant blocks based on a free text comment alone while ensuring that the amount of data is not too much reduced.
Therefore, the earlier presented blacklisting approach was chosen as a compromise.
However, there are reasons for blocks which do not allow to draw conclusions about the blocked editor's post contents.
Sock puppetry~\citeW{Wikipedia:SockPuppetry} was given $109,042$ times as a reason for blocking a registered editor.
Thus, $13.62$\% of blocks on registered editors have been issued due to editors acting as different people, e.g.\ by creating new accounts or by borrowing the accounts of friends.
The contents of posts by sock puppets vary from aggressive comments containing verbal attacks to reserved and seemingly objective input.
Considering the later, the intention of these posts is to influence others and change their minds while concealing that multiple contributions have been made by the same person disguised as different editors.
Their posts should therefore be classified as constructive from a word-level perspective.
Yet, when the sock puppets are identified as such, they will be blocked.
Regardless of choosing to ignore or consider blocks made due to sock puppets, there will be posts falsely labelled.
Future work building a model from similar data should therefore dismiss posts by editors who have at some point been identified as a sock puppet.

\begin{figure}
  \begin{minipage}{0.45\textwidth}
    \centering
    \includegraphics[width=1\linewidth]{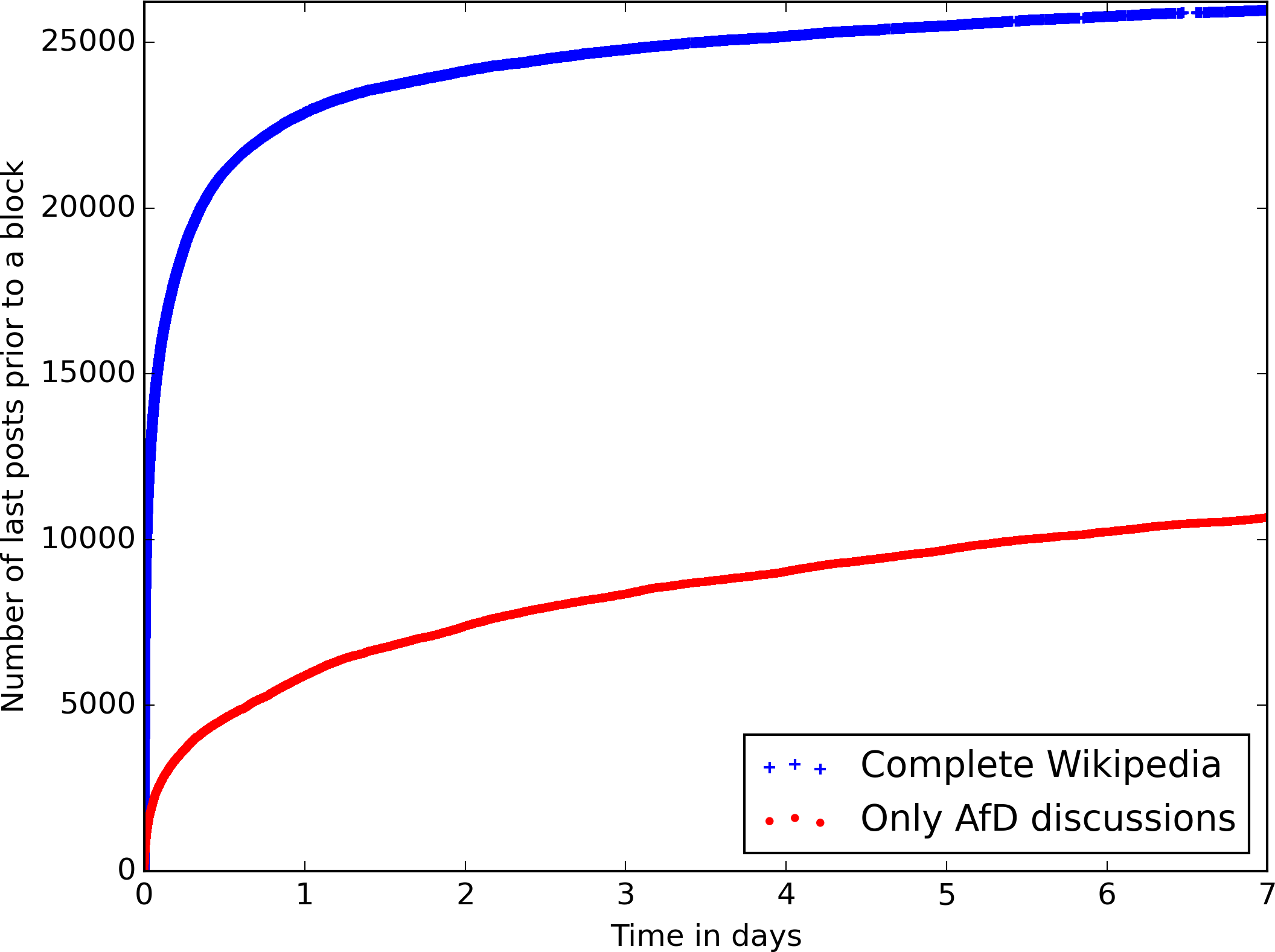}
  \end{minipage}
  \begin{minipage}{0.45\textwidth}
    \centering
    \includegraphics[width=1\linewidth]{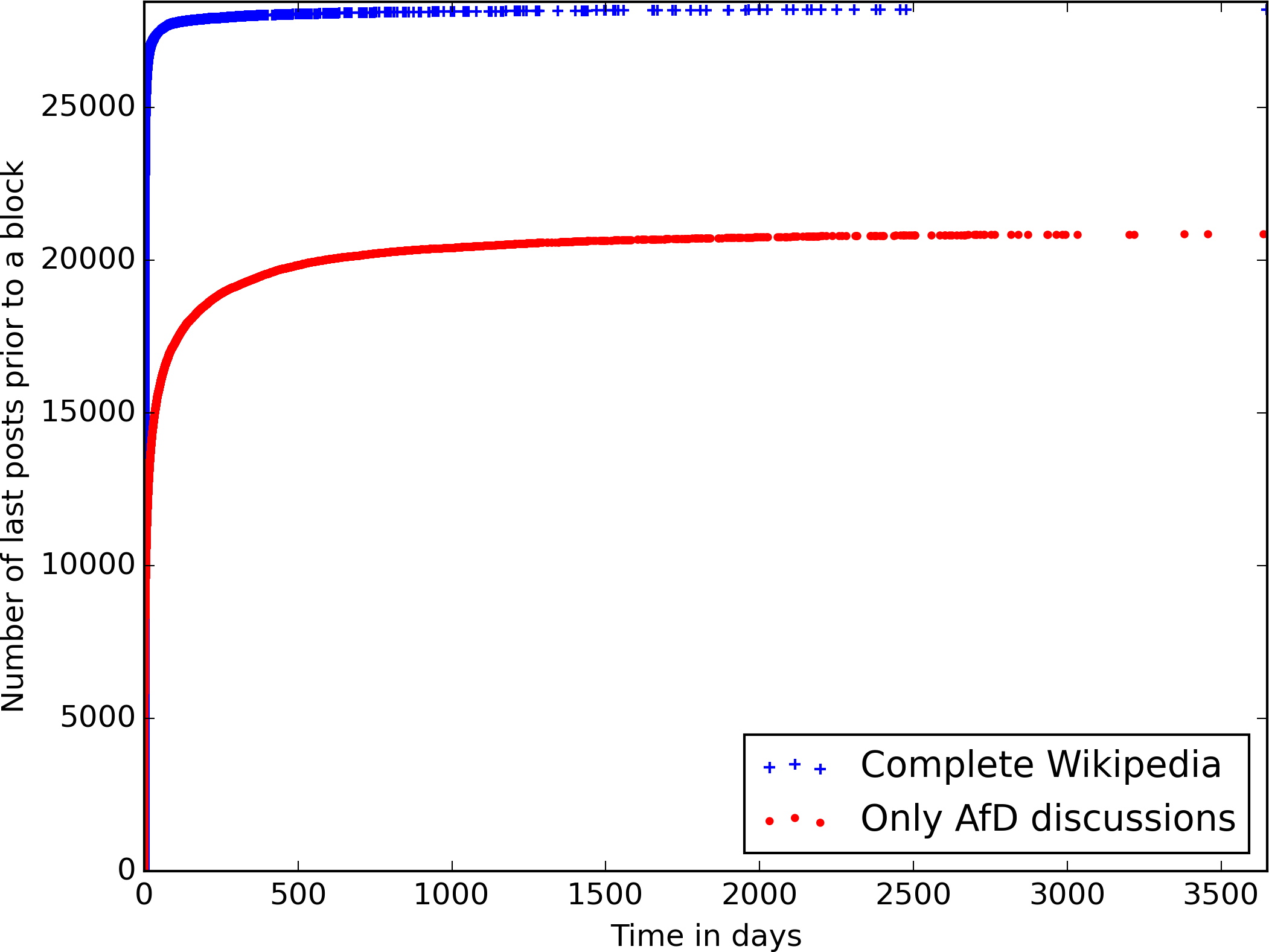}
  \end{minipage}
  \caption{Number of last posts in an \gls{discussion} and on any Wikipedia page prior to their authors being blocked within a given timeframe in days.
  Only registered editors who were active in \glspl{discussion} were considered.
  The left graph shows the first seven days of the right graph in more detail.}
\label{fig:deltasDiff}
\end{figure}
It is unclear whether blocks are a good indicator for disruptive communication.
They are often only used as ultima ratio in discussions when it is believed that the discussion would benefit from them~\citeW{Blocking_policy}.
This may frequently not be the case as \glspl{discussion} are mostly only active for seven days before a decision is made~\citeW{Wikipedia:Articles_for_deletion}.
Additionally, one of Wikipedia's behavioural guideline advises users to always assume contributions to be made in good faith unless there is hard evidence against it~\citeW{Wikipedia:GoodFaith}.
\autoref{fig:deltasDiff} visualises the time difference between the last post of a registered editor and them being blocked when only considering posts made in \glspl{discussion} compared to those made on any Wikipedia page.
Only registered editors were considered who created at least one post in an \gls{discussion}.
Blocks were filtered using the blacklisting approach presented in \autoref{ss:preprocessing_block}.
The total number of posts is higher when considering posts from all Wikipedia pages because $7,363$ registered editors have already been blocked once or multiple times before they ever contributed to an \gls{discussion}. 
As can be seen from \autoref{fig:deltasDiff}, the curve that considered all Wikipedia pages grows much faster.
This indicates that many registered editors, who have been active in at least one \gls{discussion}, have been blocked due to misbehaviour outside of these discussions.

A solution could be to use an annotated data set that was compiled by humans.
To our knowledge, there does not exist one for \glspl{discussion} yet.
Furthermore, the task is non-trivial as it is located on a post-level without any context.
This became especially clear, when we randomly picked thirty posts and asked four people to classify them.
The test cannot be seen as a serious user study but serves as mere impression on the difficulty of the task.
This difficulty is expressed in a Fleiss' kappa of $0.18$.
While only considering two classes (predicting a post to lead to a block or not within $1$ day), the low value implies that there was poor agreement among the participants.
Moreover, with an average accuracy of $55$\%, we saw the participants performing worse than the software classifiers in our independent posts approach.

\section{Difficulty of Preprocessing Posts}\label{s:errors_preprocessing}
This \namecref{s:errors_preprocessing} summarises the difficulties associated with identifying and processing posts.
It also shows problems which we had to ignore due to the limited scope of this thesis.

The first discussed difficulty is the identification of posts.
Although there are recommendations on how to highlight one's posts within a Wikipedia discussion, these are not always adhered to.
Moreover, editors may move or correct text by others or restore older revisions due to vandalism.
Both further complicate this task and thus, calculating the differences between one revision and its previous one does not suffice.
This problem was solved by using WikiWho, which is the current baseline for this task.
Nonetheless, it is still possible that some words do not get attributed to the correct author.
We also saw two instances where multiple posts had been identified by WikiWho as one.

It was important to remove non-linguistic and machine generated contents to ensure that only human communication is part of the analysis.
With the removal of the wikitext markup, potential stressing of words e.g.\ via bold font or italics was also removed.
We accepted this loss in information as there are no standardised semantics for the use of wikitext markup.
Although extending an existing solution for wikitext removal, some fragments still remained.
That is for one due to the complexity of wikitext, which also allows the use of HTML and CSS\@.
Likewise, it is related to many posts containing erroneous HTML syntax as well as not containing well-formed HTML\@. 
This problem is potentially amplified by errors made by the WikiWho algorithm.
For example, let us assume a post contained an italic emphasis \enquote{\texttt{<i>this</i> is great}}.
If WikiWho wrongfully associated the first less-than sign to a different post, the wikitext removal would no longer detect an element opening tag.
As a result, the word \enquote{\texttt{i}} would be added twice although it should not be added at all.
This is because a closing tag is only removed together with its matching opening tag.
Manual inspection showed mostly satisfying results but it has not been systematically evaluated.

To remove content that was automatically generated by algorithms of Wikipedia, signatures and templates had to be detected.
Transcluded templates were easily removed but detecting substituted templates is a difficult task.
We were able to remove the templates most commonly used in \glspl{discussion}.
Templates which were infrequently embedded in \glspl{discussion} are still present in the data which was used for analysis.
Due to them being seldom used, their effect should be negligible.
If, however, they share many words, their singular effects could add up and influence the results.

\begin{figure}
  \centering
  \includegraphics[width=0.7\textwidth]{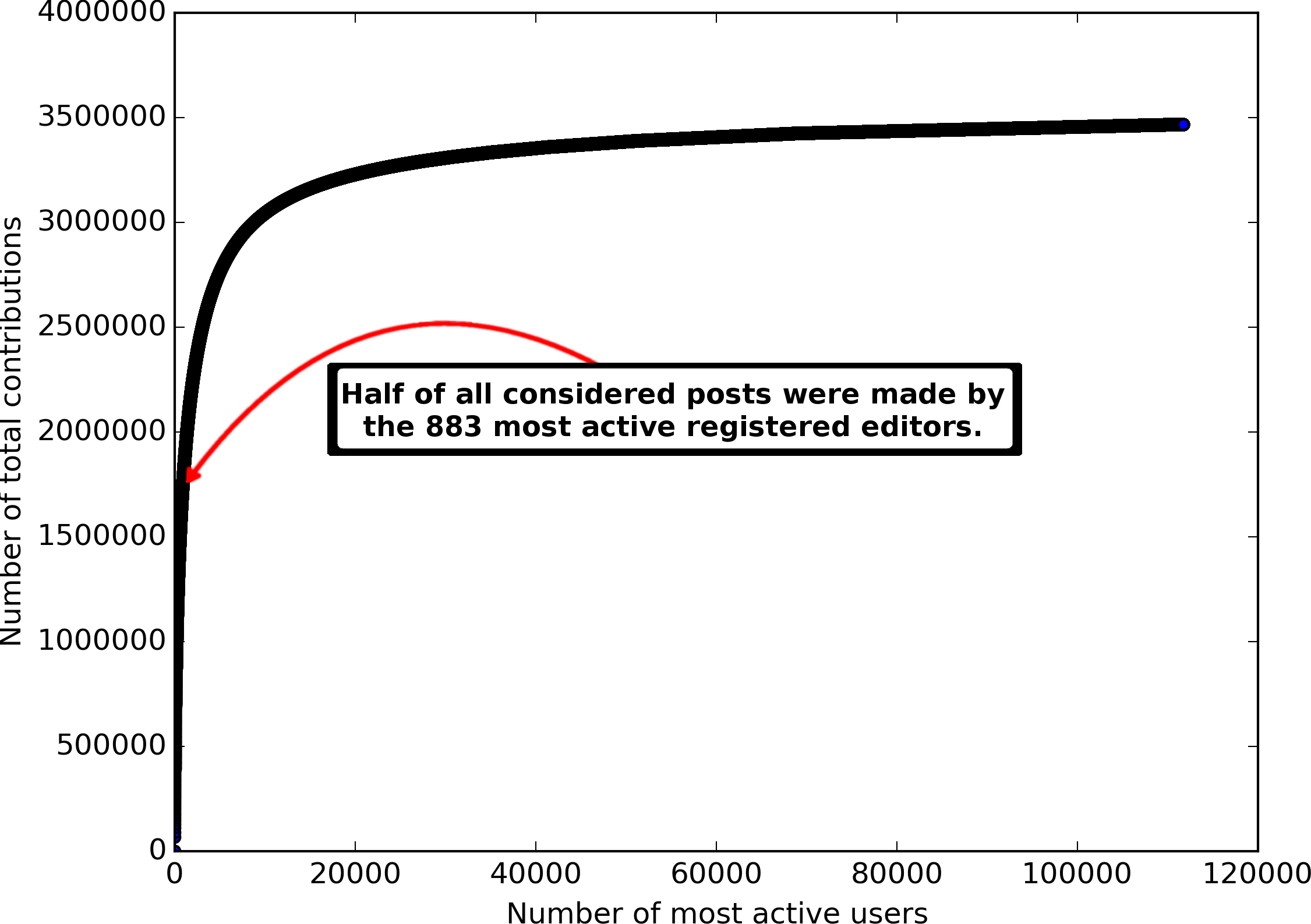}
  \caption{The graph shows the number of total posts made by the number of most active registered editors.}
\label{fig:userContribs}
\end{figure}

Signatures were removed using simple regular expressions.
Yet, they do not match all customised signatures.
\citet{ortega_inequality_2008} have shown that approximately $10$\% of all Wikipedia editors make $90$\% of all contributions.
Similarly, in \glspl{discussion} around $11.51$\% of registered editors have made $90$\% of all posts. 
\autoref{fig:userContribs} illustrates the effect which a few registered editors have on the number of posts.
A fictitious user called \enquote{Steve} might decide to modify his signature to contain \enquote{\texttt{I am Steve}}.
If this signature is not fully removed in the preprocessing step and if Steve is a heavily active editor, he can influence the results noticeably.
Following this example, if Steve makes many constructive posts, the term \I{} might be identified as important feature for constructive posts.
Moreover, the classifiers would learn to identify editors instead of language characteristics.

It can be assumed that signatures are mostly customised by very active editors because of multiple reasons.
First, not everyone may be aware of the possibility to customise one's signature and more complex customisation requires knowledge in wikitext, HTML and CSS\@.
Very active users are thus likelier to be aware of such extended features of Wikipedia and likewise to be more proficient in Wikipedia's own markup.
Second, very active users are probably also more interested in customised signatures for being noted and recognised.
Hence, the incomplete signature removal can impact the word distribution.
The full text SVM classification using the independent posts approach with the oldest \glspl{discussion} illustrated this.
It identified the two terms \emph{oct} and \emph{utc} among the top five most characteristic stemmed terms for disruptive posts.
These are assumed to be parts of signatures that have not been fully removed.
Manual inspection of posts extracted from more recent \glspl{discussion} also showed that some parts of signatures from highly active registered editors remained.

The further removal of the most common symbols including dashes, periods, commas, and brackets may have impacted our analysis.
It was a necessity to exclusively analyse a post's words and to not overcomplicate the task.
However, e.g.\ emotions in text expressed using smilies such as \enquote{\texttt{:)}} are lost.
The use of smilies can be ambiguous and is culturally dependant~\citep{park_emoticon_2013}.
Additionally, they can be used in ways where they contradict the actual content of the message e.g.\ as part of personal attacks like \enquote{\texttt{you are an idiot :)}}.
\enquote{Netspeak} is also not considered as it is an unstandardised Internet slang.
For example, the abbreviated \enquote{\texttt{b4}} instead of \enquote{\enquote{before}} will be reduced to \enquote{\texttt{b}} throughout our preprocessing steps.
Smilies and \enquote{netspeak} are both more prevalently used with messaging services that have a maximum message length to save characters and to communicate emotions.
\glspl{discussion}, however, focus on objective collaboration instead of expressing emotions and feelings.
Less than $0.40$\% of all considered posts, including customised signatures and links, contained \enquote{\texttt{:)}} or \enquote{:-)} before their removal. 
Therefore, we assume that ignoring smilies has not heavily affected the classification performance.

Due to its complexity, no efforts were made to detect irony or sarcasm.
Not only is detecting sarcasm in written text a difficult task for algorithms~\citep[cf.\ e.g.][]{liebrecht_perfect_2013,riloff_sarcasm_2013} but also for humans~\citep[cf.\ e.g.][]{abbott_how_2011,gonzalez-ibanez_identifying_2011}.
Heavy use of sarcasm could have influenced the results as the messages would seem constructive on a word-level but can be disruptive from a semantical perspective.

All words were transformed to lowercase in order to analyse the occurrence of terms while disregarding their capitalisation.
Therefore, the first word of a sentence---usually starting with a capital first letter---will not be different from when it had appeared in the middle of the sentence.
Words that are written in all capital letters are frequently perceived on the Internet as shouting and its author to being angry.
This information is lost as is the differentiation between homographs of different capitalisation.
Homographs are words which are spelled identically but have different meanings.
For example, \enquote{Dick} is a diminutive for the name Richard and will be grouped together with a slang word for the male genitalia which is often used as an insult.
Nevertheless, we argue that these cases only form rare exceptions.

To exclusively analyse human communication, posts by users whose name ended in \enquote{Bot} were ignored.
Posts by bots that do not follow the naming scheme for bots will still be left in the analysed data.
However, we estimate these chances low as bots for organisational tasks abide by the naming convention.
Additionally, bots built for vandalism purposes are assumed to be unlikely to operate on \glspl{discussion} as the harm done to Wikipedia is less visible there than on regular articles.

\section{Difficulty of Creating Comparable Results}\label{s:err_results}
As mentioned earlier, the classifiers use different implementations of the $10$-fold cross-validation.
Overall, they consider all given posts for training and testing.
However, the partitioning between our own implementation and that of RapidMiner differs.
Therefore, the SVM and NB classifier are comparable with each other but not fully with the LM classifier.
Although very likely, it can hence not be said that a language model classifier does indeed fit the task of distinguishing constructive and disruptive posts in \glspl{discussion} less.

\begin{figure}
  \centering
  \includegraphics[width=0.61\linewidth]{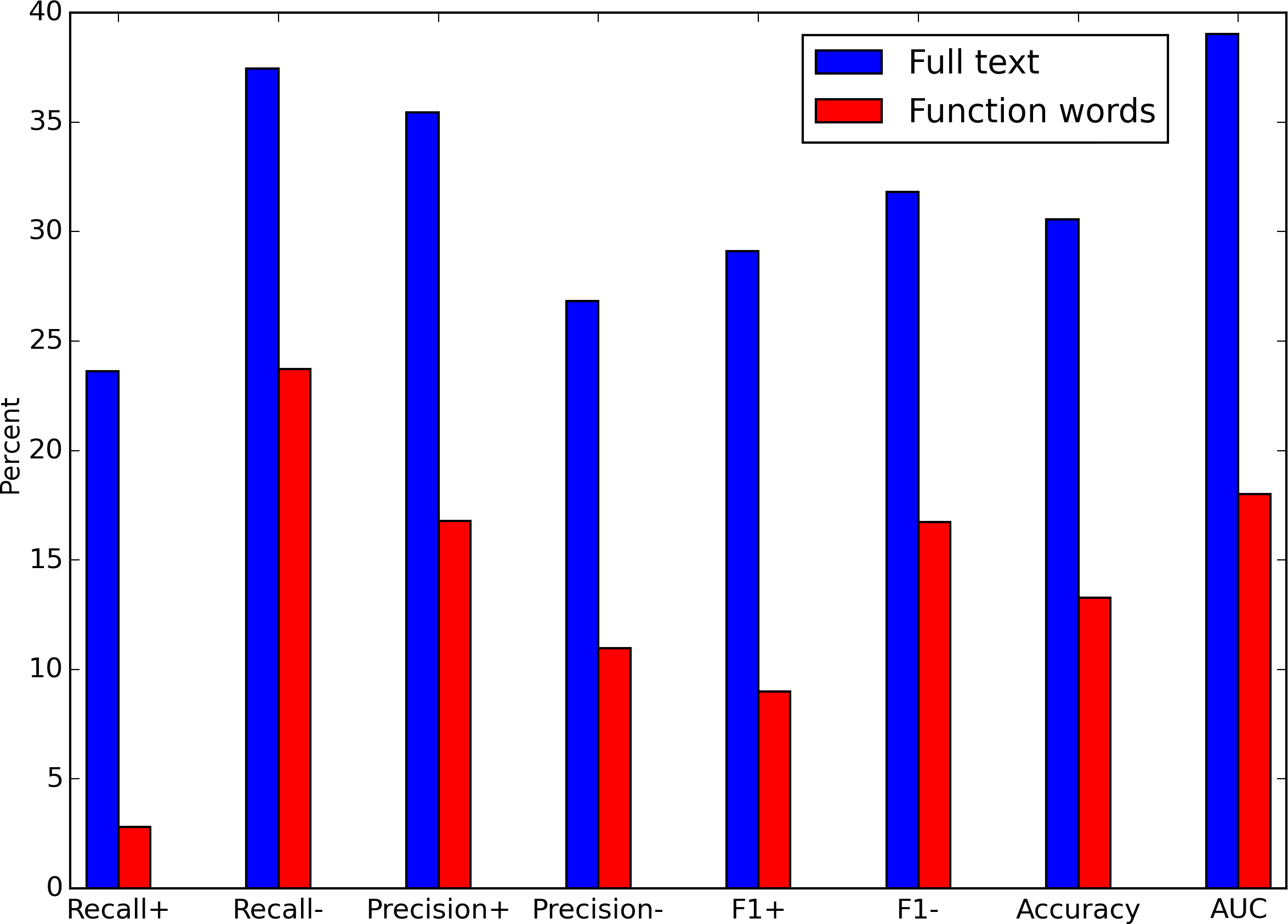}
  \caption{This bar chart shows the performance of the full text SVM in contrast to it only factoring in function words.
  The classifier used the sliding window approach with stratified sampling.
  Thus, it erroneously performed well.
  The values are in relation to a random classifier with $0$\% expressing equal performance.
  \Cref{t:val_1d_sw_strat} contains the values which were used to generate this chart as well as the results of the NB classifier.
  The percentages refer to the overall performance, meaning that $50$\% equates to a perfect result.
  }
\label{fig:stratified_svm}
\end{figure}
Due to different sampling, the sliding window and independent posts approaches as used by the SVM and NB classifier are not fully comparable either.
With the independent posts approach, we use stratified sampling as it is most similar to the sampling used for the LM classifier.
However, this was not an option for the sliding window approach or else the classifiers could potentially be tested on data they were partly learnt on.
\Cref{fig:stratified_svm} visualises the performance of an SVM using the sliding window approach with stratified sampling.
It is set in comparison to a random classifier with $p=0.5$ for each class.
Due to this sampling, the SVM performs better than a random classifier in all metrics.
As a result thereof, the tests using the sliding window approach were carried out using linear sampling.
Hence, the training and testing phases of the sliding window approach used unbalanced data.
Five iterations were run with a relation of $4$ disruptive to $5$ constructive posts.
Consequently, the other five were run with a relation of $5$ disruptive to $4$ constructive posts.
We estimate the impact of it on the results to be small but it must be considered when comparing the results.

Although related, the goal of this thesis was not to determine the best performing classifier.
Instead, the classifiers were used as a tool for analysing the data and thereby the usage of words in constructive and disruptive posts.
Therefore, in spite of the classifiers not being fully comparable, they were still suited for this task.

\chapter{Conclusion}\label{s:conclusion}
In this thesis, we set out to analyse the language used in textual discussions on a word-level, especially in regard to disruptive messages.
We chose to conduct the analysis using Wikipedia \glspl{discussion} because they provide a great amount of textual discussions.
Moreover, they feature sufficiently many instances of disruptive messages to perform a large-scale analysis.
The effects of \iyms{} and function words were of special interest.
Using the block log, we built a model that was evaluated using binary classifiers.
Thereby, we extracted terms typical for disruptive and constructive posts.

It must be noted that the subsequent answers to the research questions are based on mediocre classification results.
Nonetheless, we conducted different tests with different data samples and repeatedly saw similar results.
Therefore, we concluded that the questions could be answered.
The answers should not be interpreted as definite but rather as a strong tendency.

That is not only due to the classifiers' performances but also because it is unclear how well results from the analysis of \glspl{discussion} can be applied to general textual discussions.
Inspecting the heaviest weighted terms mostly showed terms specific to Wikipedia.
It is possible that \glspl{discussion} favour objective posts.
Thus, the probability for \iyms{} being classified as disruptive because of their subjective nature would increase.
Furthermore, we identified several difficulties such as the removal of customised signatures and transcluded templates.
In sum, they might have impacted the results notably.
Finally, a different model might be better at capturing constructive and disruptive textual messages.
Our solution for a fully automated creation of an annotated data set by using blocks might be unsuitable for the nature of \glspl{discussion}.
All things considered, we can now answer the research questions as follows:
\begin{itemize}
  \item[\textbf{\RQ{1}:}] Do disruptive messages in textual discussions contain more \yms{} than constructive ones?
\end{itemize}
  \vspace{0.3cm}
  Yes.
  Throughout all tests, terms that we deemed characteristic for \yms{} appeared in the top five of terms typical for disruptive posts.
  Despite the unsatisfying performances of the classifiers, the classifiers determined \Y{} to be more likely to appear in disruptive posts than over $30,000$ other terms.
  The results are in accordance with the occurrence frequency of \Y{} in the complete available data as was shown in \Cref{s:dataAnalysis}.
  \newpage
\begin{itemize}
  \item[\textbf{\RQ{2}:}] Do constructive messages in textual discussions contain more \ims{} than disruptive ones?\\
\end{itemize}
  \vspace{-0.2cm}
  Probably not.
  The term \I{} was found to be characteristic for disruptive posts in all tests.
  It was not as highly ranked in the lists of terms typical for disruptive posts as \Y{} was.
  Still, \I{} was clearly never a typical term for constructive posts throughout the tests.
  This is in contradiction to the observations made in \Cref{s:dataAnalysis}.
  In this \namecref{s:dataAnalysis}, we saw the term \I{} appearing a lot more frequently in constructive than in disruptive posts.
  One possible explanation could be that there might be a few constructive posts which contained the term \I{} many times whereas a great number of disruptive posts contained it only once or twice.
  \vspace{0.5cm}
\begin{itemize}
  \item[\textbf{\RQ{3}:}] Is solely considering function words sufficient for determining whether a message is constructive or rather detrimental to a textual discussion?\\
\end{itemize}
  \vspace{-0.2cm}
  Maybe.
  Due to the already mediocre performance of the full text classifiers, the function words classifiers often performed similar to or worse than a random classifier with $p=0.5$ for both classes.
  However, the NB classifier using the sliding window approach performed better when only factoring in function words.
  Consequently, the question can currently neither be affirmed nor negated.
  \vspace{0.5cm}
\begin{itemize}
  \item[\textbf{\RQ{4}:}] Which other words are typical for constructive and which for disruptive messages in textual discussions?\\
\end{itemize}
  \vspace{-0.2cm}
  The results make it difficult to answer this question as the most influential terms were mostly specific to \glspl{discussion} or Wikipedia in general.
  Results from the tests conducted using the oldest \glspl{discussion} suggest that terms expressing opposition and thus disagreement like \enquote{anti} and \enquote{against} could be characteristic for disruptive posts.
  Deducing from the results of discussion analysis from related work, terms associated with controversial topics such as religion could indicate disruptive posts as well.
  Yet, such words were not evident from our results.
  \vspace{0.5cm}

Future work should consider building a new model, e.g.\ by using manually annotated data.
It should also take the language used in \glspl{discussion} into account which is different from that used in general textual discussions.
Hence, data from more generic discussions on Wikipedia or a completely different data source could be used instead.
This could then bring new insights into the terms characteristic for constructive and disruptive messages in textual discussions as well as the correctness of our results.

\appendix
\chapter{Omitted Data and Graphics from the Test Results}\label{app:results}
\section{Effects of Different Timeframes on Classifier Performance}
\begin{table}[ht]
  \begin{tabular}{c  c  c  c  c  c  c  c  c }
\bottomrule
\textbf{time} & \textbf{recall$_{+}$} & \textbf{recall$_{-}$} & \textbf{precision$_{+}$} & \textbf{precision$_{-}$} & \textbf{F1$_{+}$} & \textbf{F1$_{-}$} & \textbf{accuracy} & \textbf{AUC}
\\\toprule
13 hours & \textbf{49.81} & 75.35 & 66.68 & 60.26 & 56.8 & 66.87 & 62.58 & 0.551\\
1 day & 49.29 & \textbf{76.63} & \textbf{67.83} & \textbf{60.27} & \textbf{56.93} & \textbf{67.42} & \textbf{62.96} & \textbf{0.562}\\
1.5 days & 47.45 & 76.03 & 66.46 & 59.22 & 55.18 & 66.52 & 61.74 & 0.542\\
2 days & 48.14 & 75.0 & 65.85 & 59.24 & 55.38 & 66.11 & 61.57 & 0.542\\
2.5 days & 46.57 & 75.77 & 65.88 & 58.72 & 54.35 & 66.09 & 61.17 & 0.536\\
3 days & 46.49 & 74.04 & 64.33 & 58.12 & 53.71 & 65.03 & 60.26 & 0.529\\
4 days & 45.35 & 74.03 & 63.72 & 57.59 & 52.76 & 64.7 & 59.69 & 0.521\\
5 days & 45.13 & 74.51 & 64.07 & 57.64 & 52.72 & 64.92 & 59.82 & 0.522\\
6 days & 42.95 & 74.6 & 63.02 & 56.7 & 50.85 & 64.35 & 58.77 & 0.512\\
\end{tabular}

  \caption{This table shows the average performance of the classifiers with the various timeframes.
  The values are the arithmetic mean of the classifiers' results.
  These are an SVM, a NB classifier and an LM classifier.
  A bold font highlights the best result considering this metric.
  The plus and minus symbols indicate whether the performance metric was calculated for disruptive ($+$) or constructive contributions ($-$).}
\label{t:timeframeResults_confusionMatrix}
\end{table}

\newpage
\section{Independent Posts Approach}
\begin{table}[hb] 
  \begin{tabular}{c  c  c  c  c  c  c  c  c }
\bottomrule
\textbf{Classifier} & \textbf{Recall$_{+}$} & \textbf{Recall$_{-}$} & \textbf{Precision$_{+}$} & \textbf{Precision$_{-}$} & \textbf{F1$_{+}$} & \textbf{F1$_{-}$} & \textbf{Accuracy} & \textbf{AUC}
\\\toprule
SVM & 59.91 & 75.56 & 71.03 & 65.33 & 65.00 & 70.07 & 67.73 & 0.750\\
SVM (FW) & 31.11 & 85.58 & 68.33 & 55.40 & 42.75 & 67.26 & 58.34 & 0.620\\
NB & 42.42 & 80.78 & 68.82 & 58.38 & 52.49 & 67.78 & 61.60 & 0.390\\
NB (FW) & 26.04 & 85.43 & 64.12 & 53.60 & 37.04 & 65.87 & 55.73 & 0.600\\
LM & 44.56 & 72.35 & 61.71 & 56.61 & 51.75 & 63.52 & 58.45 & 0.518\\
LM (FW) & 44.28 & 67.34 & 57.55 & 54.72 & 50.05 & 60.38 & 55.81 & 0.542\\
\end{tabular}

  \caption{This table shows the performance of the SVM, NB and LM classifiers using the independent posts approach.
  Function words classifications are marked with \enquote{FW}.
  All others were full text classifications.
  The plus and minus symbols indicate whether the performance metric was calculated for disruptive ($+$) or constructive contributions ($-$).}
\label{t:val_1d_ip}
\end{table}

\begin{figure}[h]
  \RawFloats%
  \centering
  \includegraphics[width=0.57\textwidth]{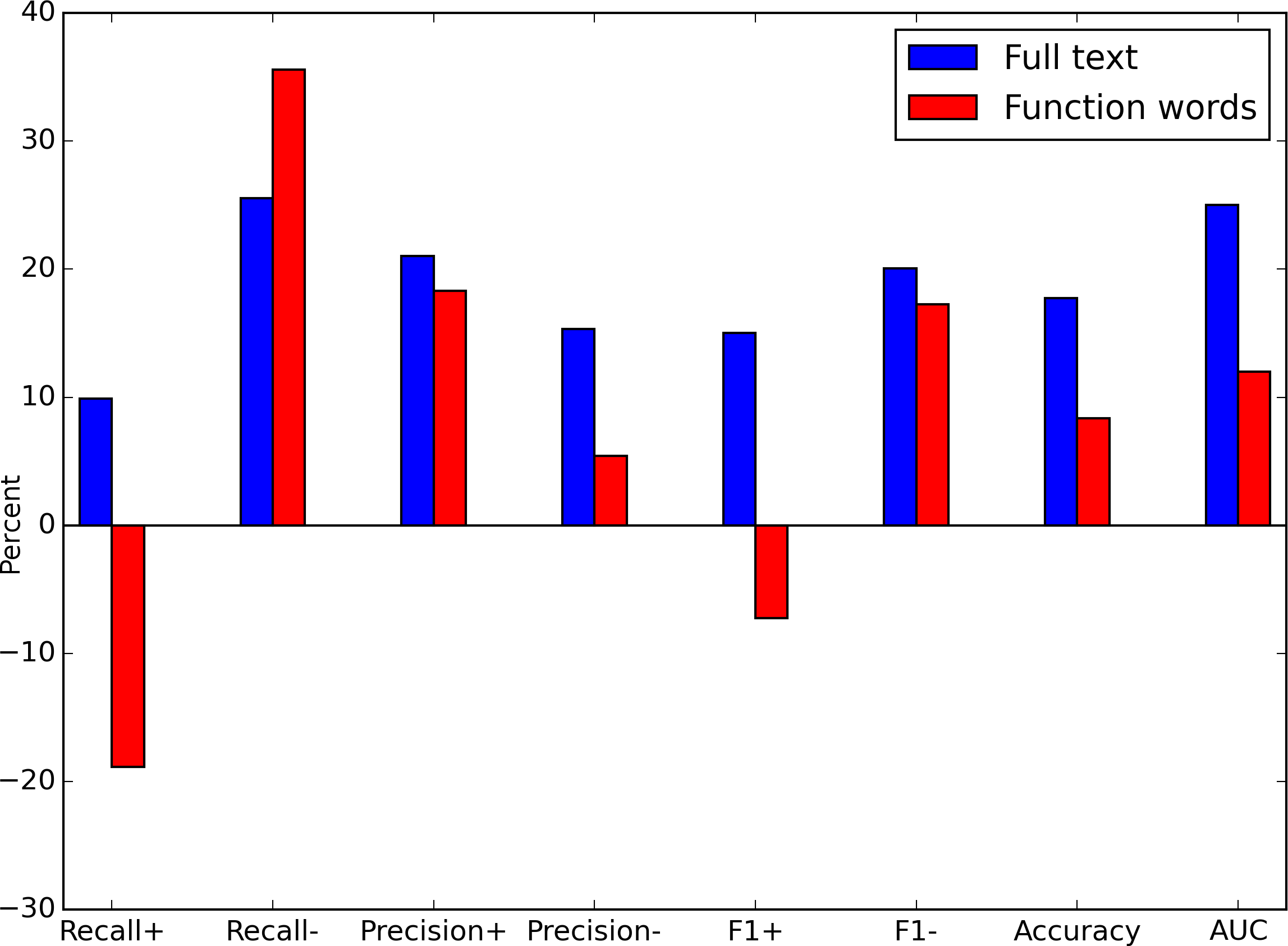}
  \caption{This bar chart shows the performance of the full text SVM in contrast to it only factoring in function words when using the independent posts approach.
  It is set into relation to the performance of a random classifier with $0$\% expressing equal performance.
  The percentages refer to the overall performance, meaning that $50$\% equates to a perfect result.
  \autoref{t:val_1d_ip} contains the visualised values.}
  \vspace{1.5em}
  \includegraphics[width=0.57\textwidth]{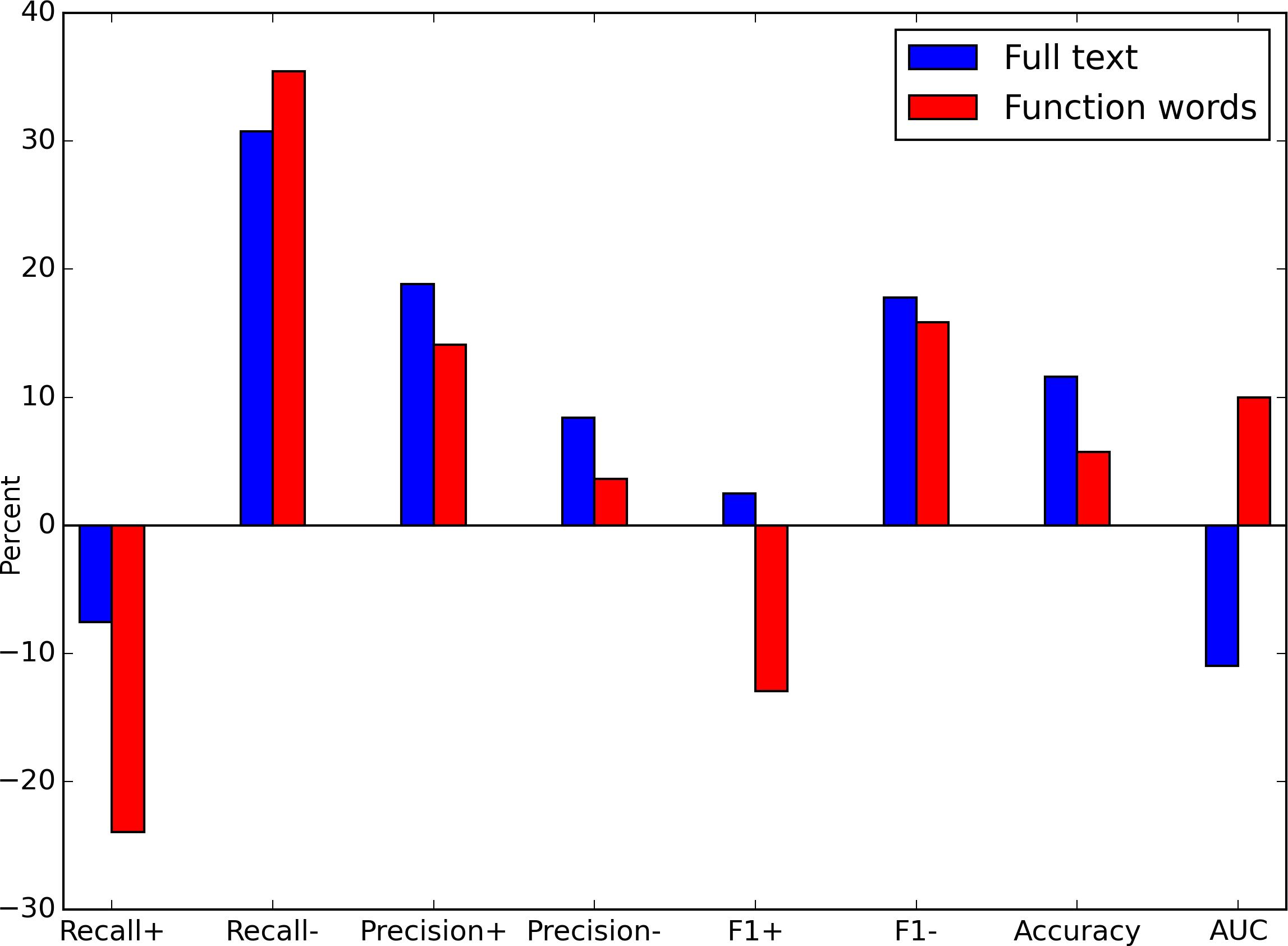}
  \caption{This bar chart shows the performance of the full text NB classifier in contrast to it only factoring in function words when using the independent posts approach.
  It is set into relation to the performance of a random classifier with $0$\% expressing equal performance.
  The percentages refer to the overall performance, meaning that $50$\% equates to a perfect result.
  \autoref{t:val_1d_ip} contains the visualised values.}
\end{figure}

\newpage
\section{Sliding Window Approach Using Linear Sampling}
\begin{table}[ht] 
  \begin{tabular}{c  c  c  c  c  c  c  c  c }
\bottomrule
\textbf{Classifier} & \textbf{Recall$_{+}$} & \textbf{Recall$_{-}$} & \textbf{Precision$_{+}$} & \textbf{Precision$_{-}$} & \textbf{F1$_{+}$} & \textbf{F1$_{-}$} & \textbf{Accuracy} & \textbf{AUC}
\\\toprule
SVM & 20.97 & 65.39 & 37.73 & 45.28 & 26.96 & 53.51 & 43.18 & --- \\
SVM (FW) & 29.63 & 37.92 & 32.31 & 35.01 & 30.91 & 36.41 & 33.77 & --- \\
NB & 64.14 & 46.39 & 54.47 & 56.40 & 58.91 & 50.91 & 55.26 & --- \\
NB (FW) & 70.11 & 51.28 & 59.00 & 63.18 & 64.08 & 56.61 & 60.70 & --- \\
LM & 13.21 & 93.26 & 66.23 & 51.80 & 22.03 & 66.61 & 53.24 & 0.497\\
LM (FW) & 4.69 & 96.88 & 60.05 & 50.41 & 8.70 & 66.31 & 50.78 & 0.559\\
\end{tabular}

  \caption{This table shows the performance of the SVM, NB and LM classifiers using the sliding window approach with linear sampling.
  RapidMiner did not return AUC values for the SVM and NB classifier, so they had to be left out.
  Function words classifications are marked with \enquote{FW}.
  All others were full text classifications.
  The plus and minus symbols indicate whether the performance metric was calculated for disruptive ($+$) or constructive contributions ($-$).}
\label{t:val_1d_sw_lin}
\end{table}
\begin{figure}[h!b]
  \centering
  \includegraphics[width=0.57\textwidth]{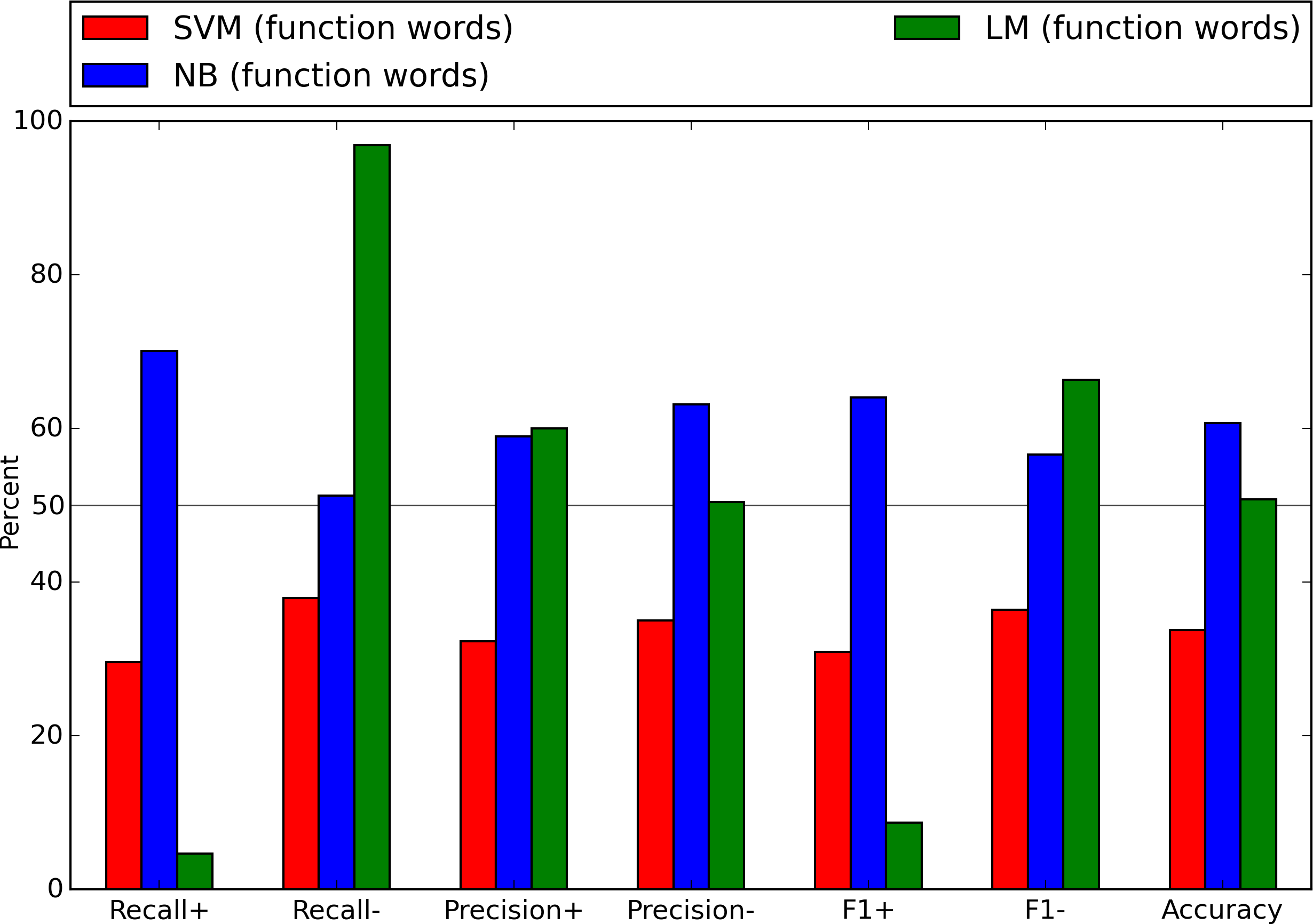}
  \caption{The performance of the full text SVM, NB and LM classifiers using the sliding window approach and linear sampling.
  The visualised values can be found in \autoref{t:val_1d_sw_lin}.}
\end{figure}
\begin{figure}[h]
  \RawFloats%
  \centering
  \includegraphics[width=0.57\textwidth]{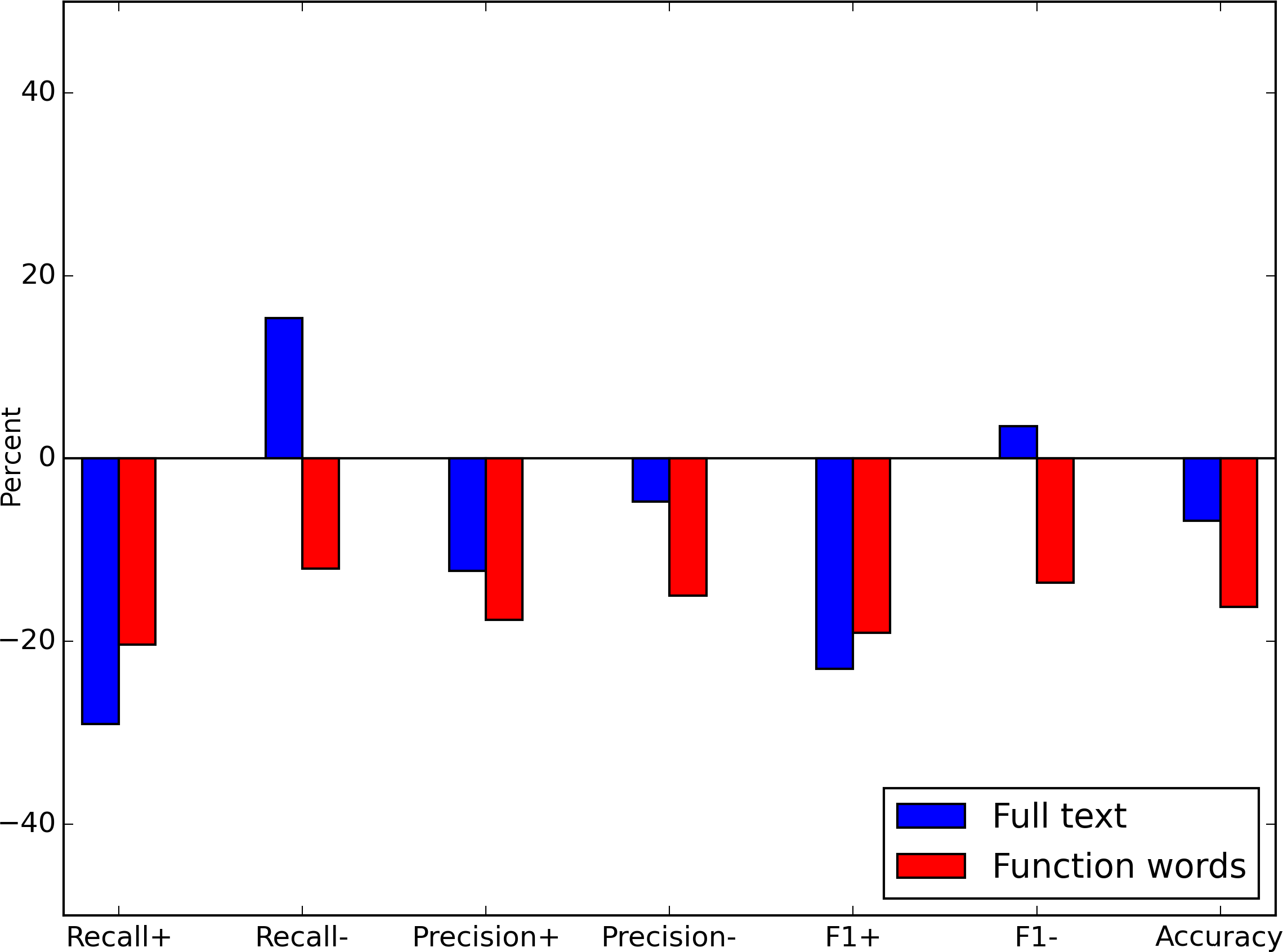}
  \caption{This bar chart shows the performance of the full text SVM in contrast to it only factoring in function words when using the sliding window approach and linear sampling.
  It is set into relation to the performance of a random classifier with $0$\% expressing equal performance.
  The percentages refer to the overall performance, meaning that $50$\% equates to a perfect result.
  \autoref{t:val_1d_ip} contains the visualised values.}
  \vspace{1.5em}
  \includegraphics[width=0.57\textwidth]{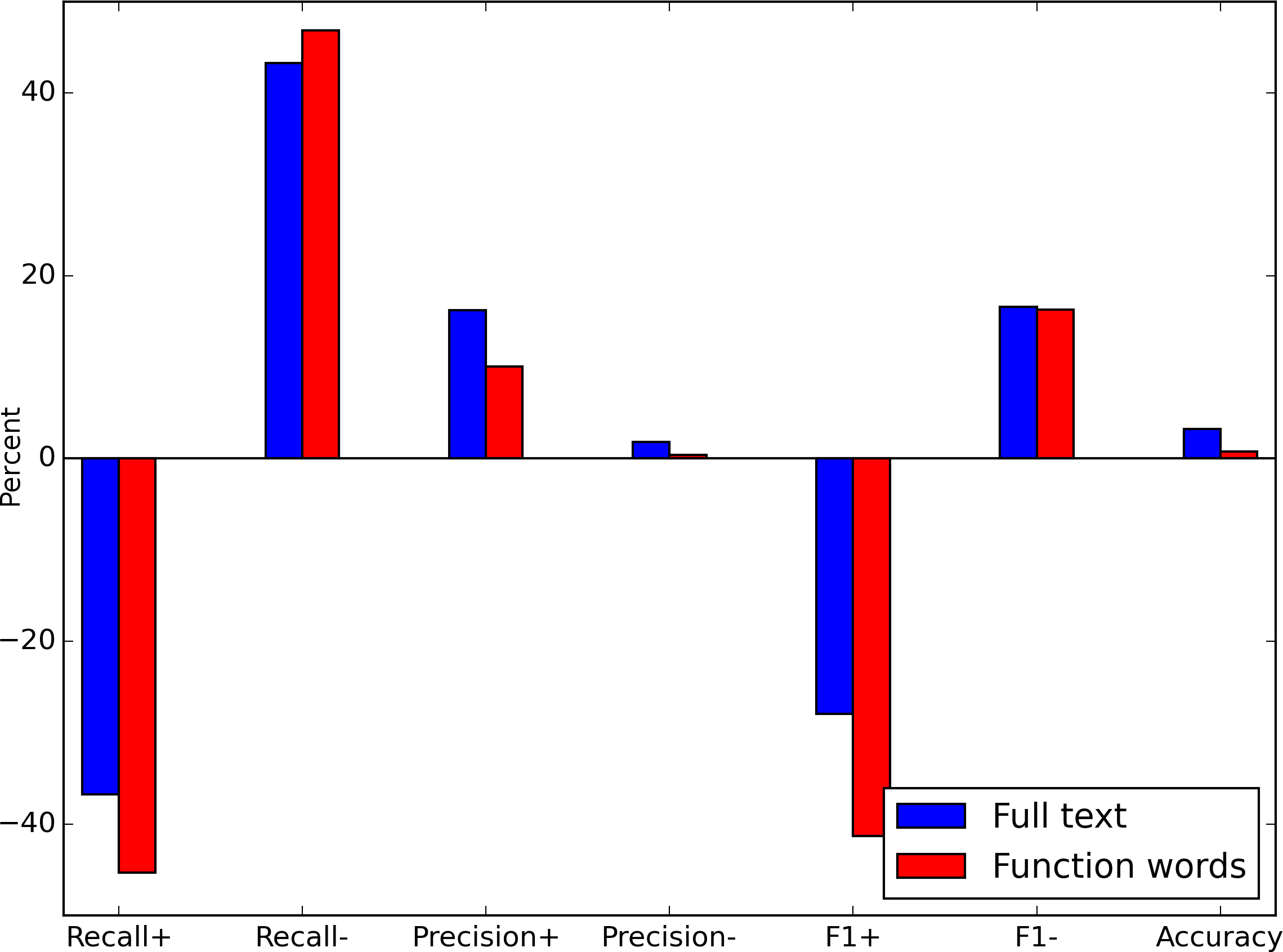}
  \caption{This bar chart shows the performance of the full text LM classifier in contrast to it only factoring in function words when using the sliding window approach and linear sampling.
  It is set into relation to the performance of a random classifier with $0$\% expressing equal performance.
  The percentages refer to the overall performance, meaning that $50$\% equates to a perfect result.
  \autoref{t:val_1d_ip} contains the visualised values.}
\end{figure}

\newpage
\section{Oldest Articles for Deletion Discussions}
\begin{table}[ht] 
  \begin{tabular}{c  c  c  c  c  c  c  c  c }
\bottomrule
\textbf{Classifier} & \textbf{Recall$_{+}$} & \textbf{Recall$_{-}$} & \textbf{Precision$_{+}$} & \textbf{Precision$_{-}$} & \textbf{F1$_{+}$} & \textbf{F1$_{-}$} & \textbf{Accuracy} & \textbf{AUC}
\\\toprule
SVM & 68.26 & 89.98 & 87.20 & 73.93 & 76.58 & 81.17 & 79.12 & 0.890\\
SVM (FW) & 29.86 & 90.68 & 76.21 & 56.39 & 42.91 & 69.54 & 60.27 & 0.660\\
NB & 49.10 & 90.44 & 83.70 & 63.99 & 61.89 & 74.95 & 69.77 & 0.470\\
NB (FW) & 37.30 & 82.43 & 67.98 & 56.80 & 48.17 & 67.26 & 59.86 & 0.630\\
LM & 52.04 & 81.08 & 73.33 & 62.83 & 60.88 & 70.80 & 66.56 & 0.541\\
LM (FW) & 50.08 & 68.04 & 61.04 & 57.68 & 55.02 & 62.43 & 59.06 & 0.582\\
\end{tabular}

  \caption{This table shows the performance of the SVM, NB and LM classifiers using the independent posts approach on the oldest \glspl{discussion}.
  Function words classifications are marked with \enquote{FW}.
  All others were full text classifications.
  The plus and minus symbols indicate whether the performance metric was calculated for disruptive ($+$) or constructive contributions ($-$).}
\label{t:val_1d_ip_chrono}
\end{table}
\begin{figure}[h!b]
  \centering
  \includegraphics[width=0.57\textwidth]{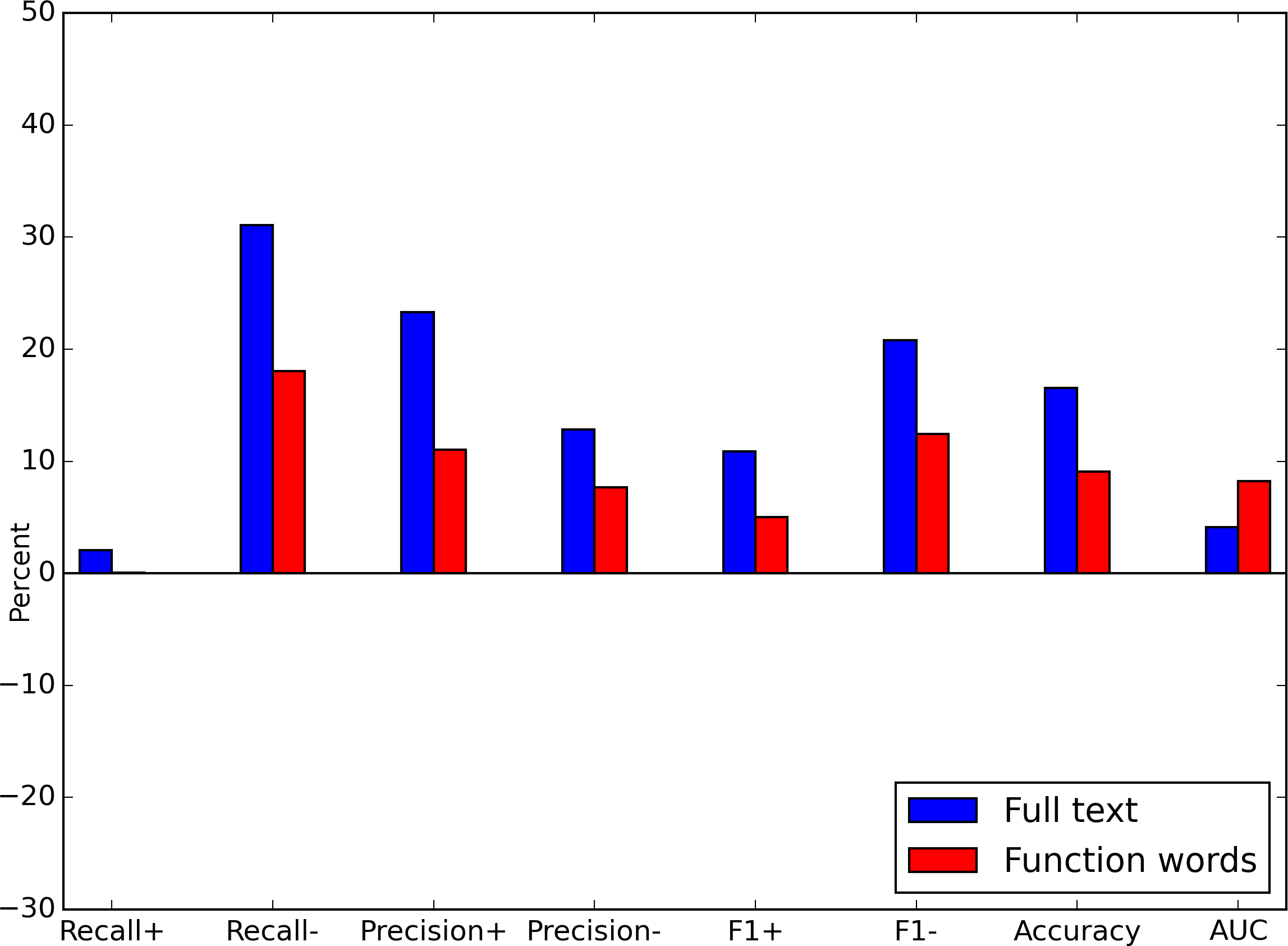}
  \caption{This bar chart shows the performance of the full text LM classifier in contrast to it only factoring in function words when using the independent posts approach on the oldest \glspl{discussion}.
  It is set into relation to the performance of a random classifier with $0$\% expressing equal performance.
  The percentages refer to the overall performance, meaning that $50$\% equates to a perfect result.
  \autoref{t:val_1d_ip} contains the visualised values.}
\end{figure}
\begin{figure}[h]
  \RawFloats%
  \centering
  \includegraphics[width=0.57\textwidth]{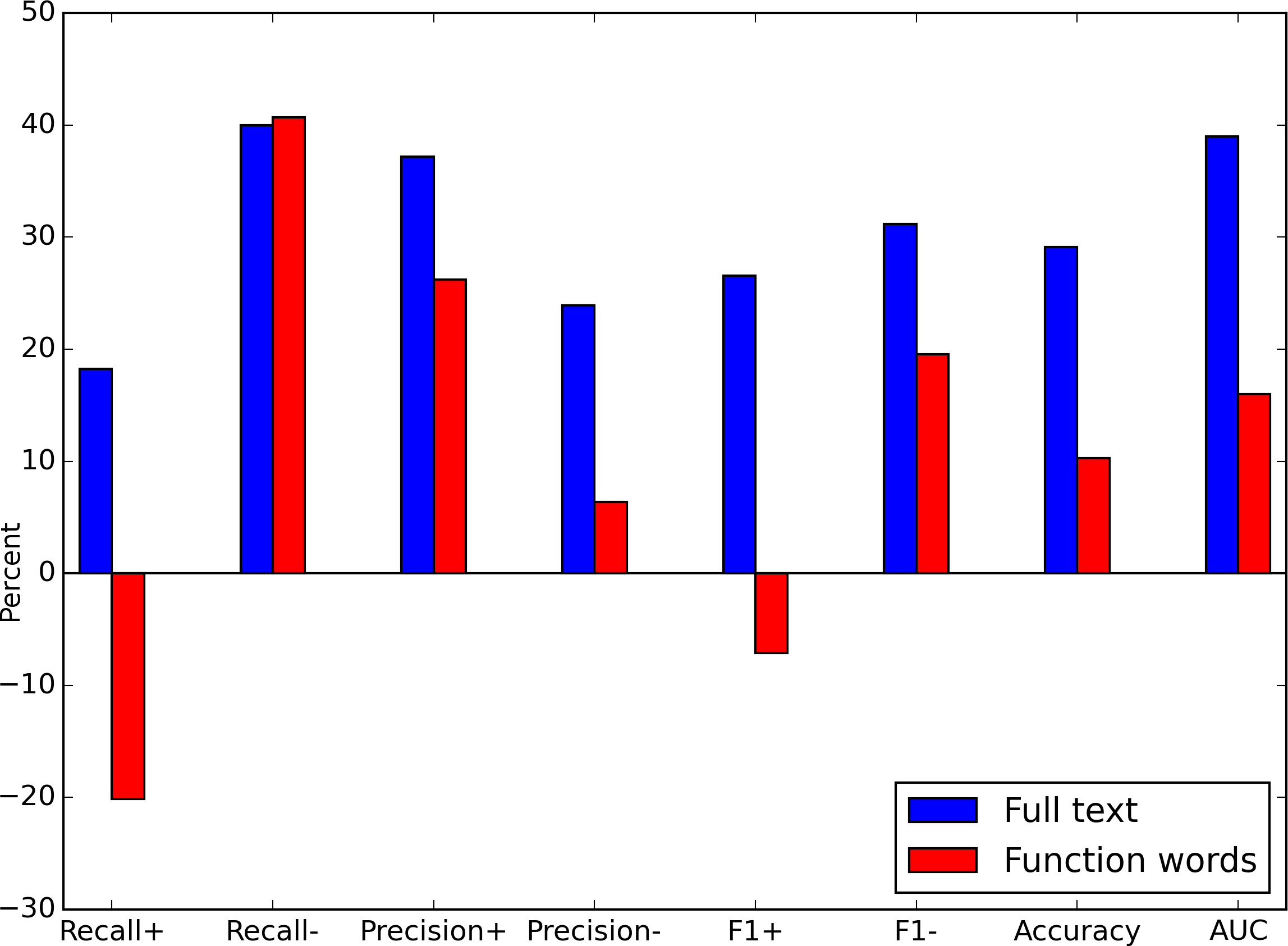}
  \caption{This bar chart shows the performance of the full text SVM in contrast to it only factoring in function words when using the independent posts approach on the oldest \glspl{discussion}.
  It is set into relation to the performance of a random classifier with $0$\% expressing equal performance.
  The percentages refer to the overall performance, meaning that $50$\% equates to a perfect result.
  \autoref{t:val_1d_ip} contains the visualised values.}
  \vspace{1.5em}
  \includegraphics[width=0.57\textwidth]{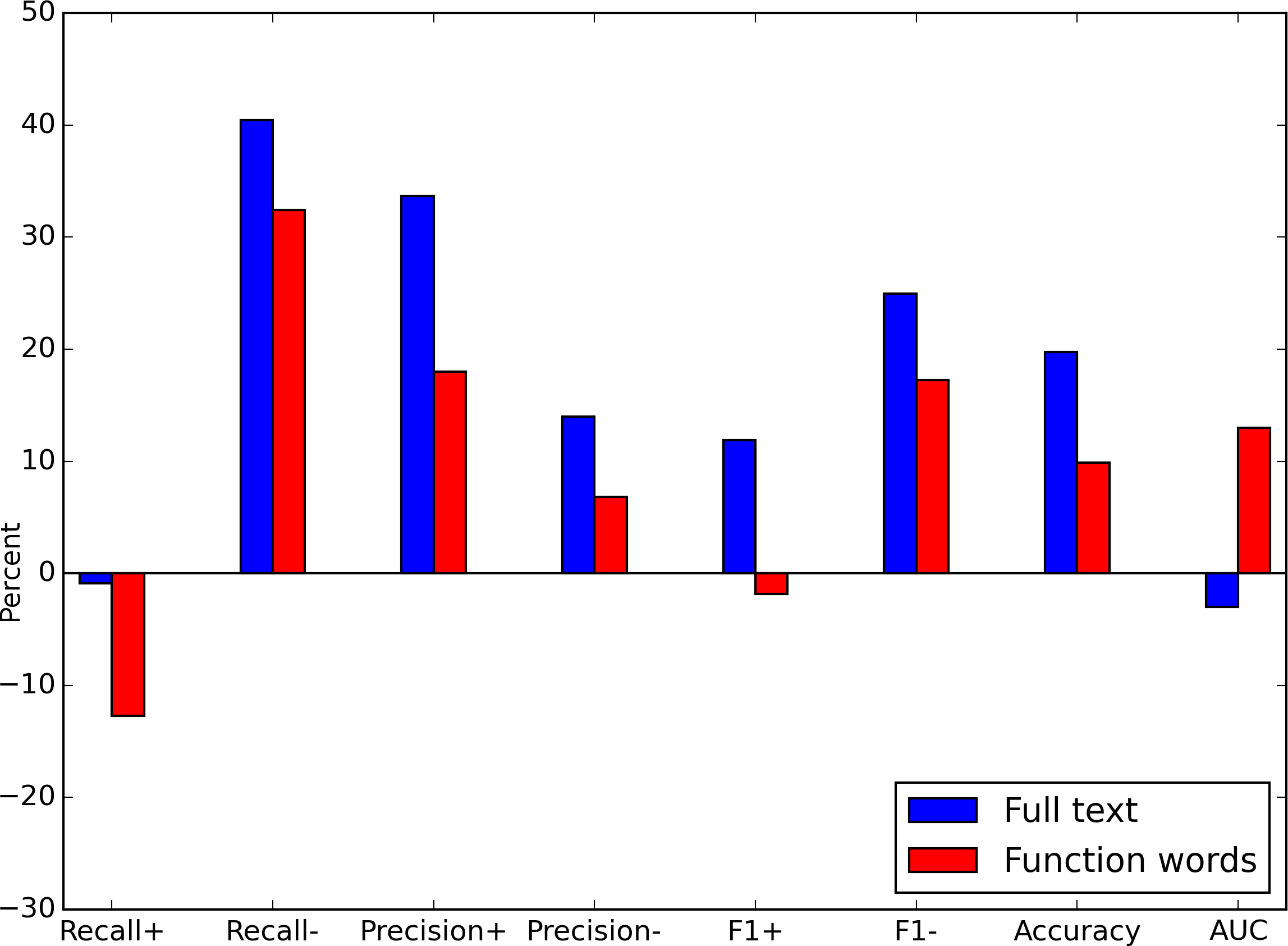}
  \caption{This bar chart shows the performance of the full text NB classifier in contrast to it only factoring in function words when using the independent posts approach on the oldest \glspl{discussion}.
  It is set into relation to the performance of a random classifier with $0$\% expressing equal performance.
  The percentages refer to the overall performance, meaning that $50$\% equates to a perfect result.
  \autoref{t:val_1d_ip} contains the visualised values.}
\end{figure}

\newpage
\section{Sliding Window Approach Using Stratified Sampling}
\begin{table}[h] 
  \begin{tabular}{c  c  c  c  c  c  c  c  c }
\bottomrule
\textbf{Classifier} & \textbf{Recall$_{+}$} & \textbf{Recall$_{-}$} & \textbf{Precision$_{+}$} & \textbf{Precision$_{-}$} & \textbf{F1$_{+}$} & \textbf{F1$_{-}$} & \textbf{Accuracy} & \textbf{AUC}
\\\toprule
SVM & 88.96 & 84.60 & 85.25 & 88.46 & 87.07 & 86.49 & 86.78 & 0.950\\
SVM (FW) & 78.73 & 50.75 & 61.52 & 70.47 & 69.07 & 59.01 & 64.74 & 0.710\\
NB & 69.57 & 93.43 & 91.37 & 75.43 & 78.99 & 83.47 & 81.50 & 0.670\\
NB (FW) & 42.27 & 79.33 & 67.16 & 57.88 & 51.88 & 66.93 & 60.80 & 0.670\\
\end{tabular}

  \caption{This table shows the performance of the SVM and NB classifier using the sliding window approach with stratified sampling.
  Function words classifications are marked with \enquote{FW}.
  All others were full text classifications.
  The plus and minus symbols indicate whether the performance metric was calculated for disruptive ($+$) or constructive contributions ($-$).}
\label{t:val_1d_sw_strat}
\end{table}

\glsaddall%
\printglossary%

\clearpage
\printbibheading%
\printbibliography[notkeyword={web},title={Literature Sources},heading=subbibliography]
\printbibliography[keyword={web},title={Wikipedia Sources},prefixnumbers={W},heading=subbibliography,resetnumbers]

\end{document}